\documentclass[11pt,a4paper]{article}
\usepackage[hyperref]{emnlp2020}
\usepackage{times}
\usepackage{latexsym}

\usepackage{microtype}

\aclfinalcopy % Uncomment this line for the final submission
\def\aclpaperid{2740} %  Enter the acl Paper ID here

\usepackage[export]{adjustbox}

\usepackage{times}  % DO NOT CHANGE THIS
\usepackage{helvet} % DO NOT CHANGE THIS
\usepackage{courier}  % DO NOT CHANGE THIS
\usepackage{url}  % DO NOT CHANGE THIS
\usepackage{graphicx} % DO NOT CHANGE THIS
\usepackage{booktabs}       % professional-quality tables
\usepackage{amsfonts}       % blackboard math symbols
\usepackage{nicefrac} 
\usepackage{latexsym}
\usepackage{multirow}
\usepackage{booktabs}
\usepackage{amsmath}
\usepackage{booktabs}
\usepackage{courier}
\usepackage{colortbl}
\usepackage[position=bottom]{subfig}
\usepackage{microtype}
\usepackage{array}
\usepackage{wrapfig}
\usepackage{color,xcolor}

\usepackage{alphalph}
\usepackage{caption}
\usepackage{makecell}

\title{\textit{CDEvalSumm}: An Empirical Study of \textit{C}ross-\textit{D}ataset \textit{Eval}uation \\for Neural \textit{Summ}arization Systems}

\author{Yiran Chen\thanks{\hspace{1mm} These two authors contributed equally.}, Pengfei Liu$\sharp$\footnotemark[1], Ming Zhong, Zi-Yi Dou$\sharp$, Danqing Wang,\\  {\bf Xipeng Qiu}\thanks{\ \  Corresponding author.} ,  {\bf Xuanjing Huang} \\
  Shanghai Key Laboratory of Intelligent Information Processing, Fudan University \\
  School of Computer Science, Fudan University \\
  2005 Songhu Road, Shanghai, China \\
  $\sharp$Carnegie Mellon University \\
  \texttt{\{yrchen19,mzhong18,dqwang18,xpqiu,xjhuang\}@fudan.edu.cn} \\ \texttt{\{zdou,pliu3\}@cs.cmu.edu}}
\date{}

\begin{document}
\maketitle
\begin{abstract}
Neural network-based models augmented with unsupervised pre-trained knowledge have achieved impressive performance on text summarization. However, most existing evaluation methods are limited to an \emph{in-domain} setting, where summarizers are trained and evaluated on the same dataset. We argue that this approach can narrow our understanding of the generalization ability for different summarization systems. In this paper, we perform an in-depth analysis of characteristics of different datasets and investigate the performance of different summarization models under a cross-dataset setting, in which a summarizer trained on one corpus will be evaluated on a range of out-of-domain corpora. A comprehensive study of 11 representative summarization systems on 5 datasets from different domains reveals the effect of model architectures and generation ways (i.e. abstractive and extractive) on model generalization ability. Further, experimental results shed light on the limitations of existing summarizers. Brief introduction and supplementary code can be found in \url{https://github.com/zide05/CDEvalSumm}.

\end{abstract}

\begin{figure}[t]
  \centering
  \includegraphics[width=0.77\linewidth]{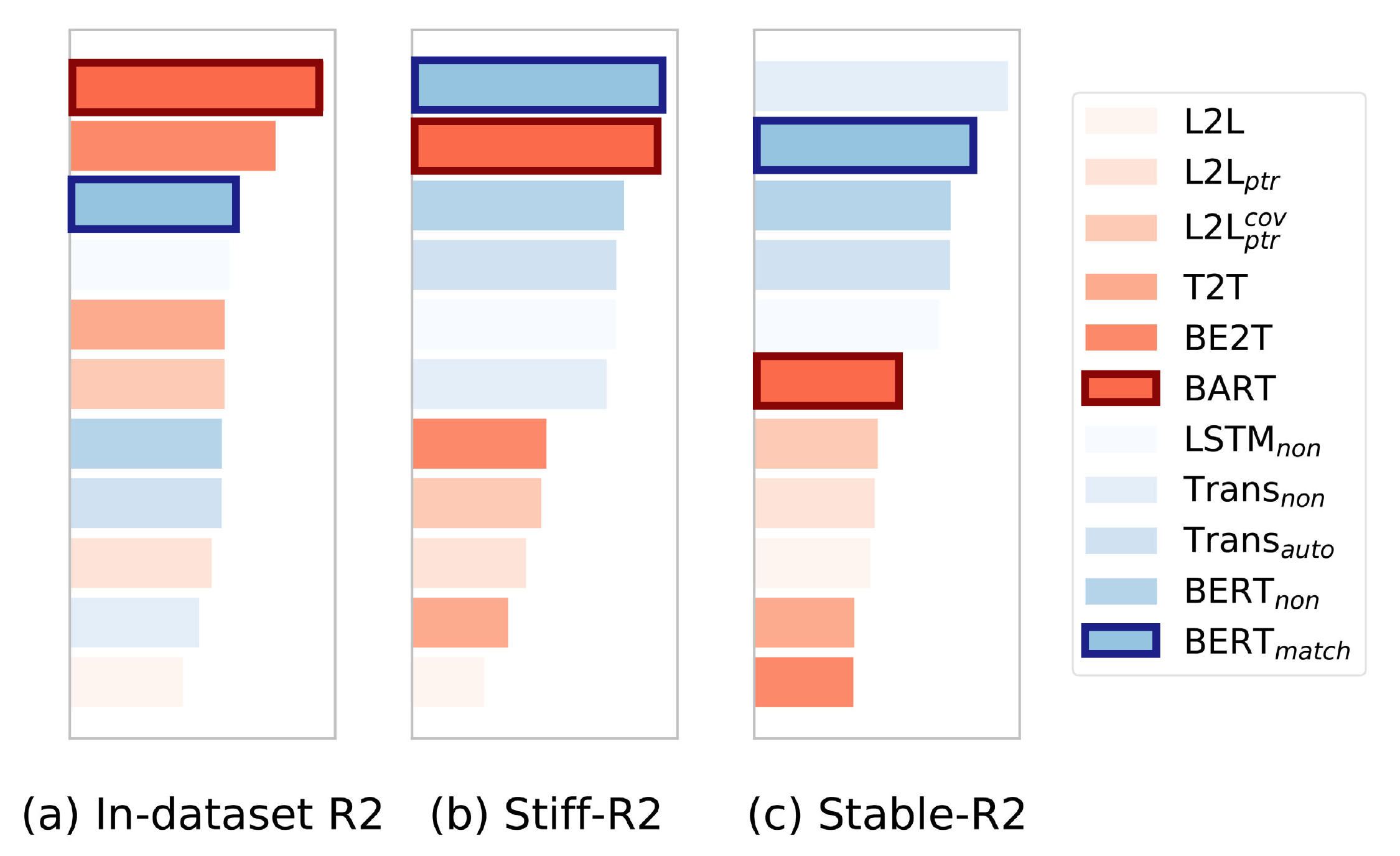}
 \caption{Ranking (descending order) of current 11 top-scoring summarization systems (Abstractive models are red while extractive ones are blue). Each system is evaluated based on three diverse evaluation methods: (a) averaging each system's in-dataset ROUGE-2 F1 scores (R2) over five datasets; (b-c) evaluating systems using our designed cross-dataset measures: \emph{stiff-R2}, \emph{stable-R2} (Sec.~\ref{measures}). Notably, \textit{BERT$_{match}$} and \textit{BART} are two state-of-the-art models for extractive and abstractive summarization respectively (highlighted by blue and red boxes).}
 \label{fig:ranking}
\end{figure}

\section{Introduction}
\label{intro}

Neural summarizers have achieved impressive performance when evaluated by ROUGE ~\citep{lin2004rouge} on in-domain setting, and the recent success of pre-trained models drives the state-of-the-art results on benchmarks to a new level ~\citep{liu2019text,liu2019fine,zhong2019searching,zhang2019pegasus,lewis2019BART,zhong2020extractive}.
However, the superior performance is not a guarantee of a perfect system since exsiting models tend to show defects when evaluated from other aspects.
For example, \citet{zhang-etal-2018-abstractiveness} observes that many abstractive systems tend to be near-extractive in practice.
\citet{cao2018faithful,wang2020asking,kryscinski2019evaluating,maynez2020faithfulness,durmus2020feqa} reveal that most generated summaries are factually incorrect.
These non-mainstream evaluation methods make it easier to identify the model's weaknesses.

Orthogonal to above two evaluation aspects, we aim to diagnose the limitation of existing systems under \emph{cross-dataset evaluation}, in which a summarization system trained on  one corpus would be evaluated on a range of out-of-dataset corpora. Instead of evaluating the quality of summarizers solely based on one dataset or multiple datasets individually, cross-dataset evaluation enables us to evaluate model performance from a  different angle.
For example, Fig.~\ref{fig:ranking} shows the ranking of $11$ summarization systems studied in this paper under different  evaluation metrics, in which the ranking list ``(a) in-dataset R2'' is obtained by traditional ranking criteria while other two are based on our designed cross-dataset measures.
Intuitively, we observe that 1) there are different definitions of a ``good'' system in various evaluation aspects;
2) abstractive and extractive systems exhibit diverse behaviors when evaluated under the cross-dataset setting.

The above example recaps the general motivation of this work, encouraging us to rethink the generalization ability of current top-scoring summarization systems from the perspective of cross-dataset evaluation.
Specifically, we ask two questions as follows:

\textbf{Q1}: {How do different neural architectures of summarizers influence the cross-dataset generalization performances?}
When designing summarization systems, a plethora of neural components can be adopted ~\citep{zhou2018neural,chen2018fast,gehrmann2018bottom,cheng2016neural,nallapati2017summarunner}.
For example, will \emph{copy} \cite{gu2016incorporating} and \emph{coverage} \cite{see2017get}  mechanisms improve the cross-dataset generalization ability of summarizers? Is there a risk that \emph{BERT-based} summarizers will perform worse when adapted to new areas compared with the ones \textit{without BERT}?
So far, the generalization ability of current summarization systems when transferring to new datasets still remains unclear, which poses a significant challenge to design a reliable system in realistic scenarios.
Thus, in this work, we take a closer look at the effect of model architectures on cross-dataset generalization setting.

\textbf{Q2}: {Do different generation ways (\textit{extractive} and \textit{abstractive}) of summarizers influence the cross-dataset generalization ability?}
Extractive and abstractive models, as two typical ways to summarize texts, usually follow diverse learning frameworks and favor different datasets. 
It would be absorbing to know their discrepancy from the perspective of cross-dataset generalization. 
(e.g., whether abstractive summarizers are better at generating informative or faithful summaries on a new test set?)

To answer the questions above, we have conducted a comprehensive experimental analysis, which involves \textit{eleven} summarization systems (including the state-of-the-art models), \textit{five} benchmark datasets from different domains, and two evaluation aspects.
Tab.~\ref{tab:structure} illustrates the overall analysis framework. We explore the effect of different architectures and generation ways on model generalization ability in order to answer \emph{Q1} and \emph{Q2}. Semantic equivalency (e.g., ROUGE) and factuality are adopted to characterize the different aspects of cross-dataset generalization ability.
Additionally, we strengthen our analysis by presenting two views of evaluation: \emph{holistic} and \emph{fine-grained} views (Sec.~\ref{measures}).

\begin{table}[t]
  \centering
  \renewcommand{\arraystretch}{2}
  \resizebox{0.47\textwidth}{12mm}{
    \begin{tabular}{lcc}
    \toprule
    \textbf{Framework} & \makecell{\textbf{Semantic equivalency} \\ (e.g., \texttt{ROUGE})} & \makecell{\textbf{Factuality} \\ (e.g., \texttt{Factcc})} \\
    \midrule
    \makecell[l]{\textbf{Q1: Architecture} \\ (e.g., \texttt{Transformer} v.s. \texttt{LSTM})} & Sec.~\ref{R1 architecture}  & Sec.~\ref{factcc analysis}\\
    \makecell[l]{\textbf{Q2: Generation way} \\ (e.g., \texttt{BERT} v.s. \texttt{BART})} & Sec.~\ref{R1 generation way}  & Sec.~\ref{factcc analysis}\\
    \bottomrule
    \end{tabular}}%
    \caption{Overall analysis framework.}
  \label{tab:structure}%
\end{table}%

Our contributions can be summarized as:
1) Cross-dataset evaluation is orthogonal to other evaluation aspects (e.g., semantic equivalence, factuality), which can be used to re-evaluate current summarization systems, accelerating the creation of more robust summarization systems.
2) We have design two measures \textit{Stiffness} and \textit{Stableness}, which could help us to characterize generalization ability in different views, encouraging us to diagnose the weaknesses of state-of-the-art systems. 
3) We conduct dataset bias-aided analysis (Sec.~\ref{dataset bias}) and suggest that a better understanding of datasets will be helpful for us to interpret systems'  behaviours.

\section{Representative Systems}

Although it's intractable to cover all neural summarization systems, we try to include more representative models to make a comprehensive evaluation.
Our selection strategy follows: 1) the source codes of systems are publicly available; 2) systems with state-of-the-art performance or the top performace on benchmark datasets (e.g., \texttt{CNNDM}~\citep{nallapati2016abstractive}) 3) systems equipped with typical neural components (e.g., Transformer, LSTM) or mechanism (e.g., copy).

\subsection{Extractive Summarizers}
Extractive summarizers directly choose and output the salient sentences (or phrases) in the original document.
Generally,  most of the existing extractive summarization systems follow a framework consisting of three major modules: \textit{sentence encoder}, \textit{document encoder} and \textit{decoder}.
In this paper, we investigate extractive summarizers with different choices of encoders and decoders.

\noindent \textbf{LSTM$_{non}$} ~\citep{kedzie2018content} 
This summarizer adopts convolutional neural network as sentence encoder and LSTM to model the cross-sentence relation. Finally, each sentence will be selected in a non-autoregressive way.

\noindent \textbf{Trans$_{non}$} ~\citep{liu2019text}
The TransformerExt model in ~\citet{liu2019text}, similar to above setting except that the document encoder is replaced with the Transformer layer.

\noindent \textbf{Trans$_{auto}$}  ~\citep{zhong2019searching}
The decoder is replaced with a pointer network to avoid the repetition (autoregressive).

\noindent \textbf{BERT$_{non}$} ~\citep{liu2019text}
The BertSumExt model in ~\citet{liu2019text}, this model is an extension of Trans$_{non}$ by introducing a BERT ~\citep{devlin2018BERT} layer.

\noindent \textbf{BERT$_{match}$} \cite{zhong2020extractive}
This is the existing state-of-the-art extractive summarization system, which introduce a matching layer using siamese BERT.

\subsection{Abstractive Summarizers}
The abstractive approach involves paraphrasing the inputs using novel words.
The current abstractive summarization systems mainly focus on the \textit{encoder-decoder} paradigm.

\noindent \textbf{L2L$_{ptr}^{cov}$} ~\citep{see2017get}
The model is a LSTM based sequence to sequence summarizer with copy and coverage mechanism.

\noindent \textbf{L2L$_{ptr}$}
We remove the coverage module and keep other parts unchanged. 

\noindent \textbf{L2L}
This model is implemented by removing the pointer network of the above summarizer.

\noindent \textbf{T2T} ~\citep{liu2019text}
A sequence to sequence model with Transformer as the encoder and decoder.

\noindent \textbf{BE2T} ~\citep{liu2019text}
A sequence to sequence model with BERT as  encoder and Transformer as decoder.

\noindent \textbf{BART} ~\cite{lewis2019BART}
A fully pre-trained sequence to sequence model. It is the existing state-of-the-art abstractive summarization system.

\section{Datasets}

We explore five typical summarization datasets: \texttt{CNNDM}, \texttt{Xsum},  \texttt{PubMed}, \texttt{Bigpatent B} and \texttt{Reddit TIFU}. \texttt{CNNDM}~\citep{nallapati2016abstractive} and \texttt{Xsum} ~\citep{narayan2018don} are news domain summarization datasets which are various in their publications and abstractiveness. \texttt{PubMed} ~\citep{cohan2018discourse} is a scientific paper dataset, which can be used to investigate the generalization ability of models on scientific domain. 
\texttt{Bigpatent B} ~\citep{sharma2019bigpatent} is the B category of \texttt{Bigpatent} (a dataset consisting of patent documents from Google Patents Public Datasets). \texttt{Reddit TIFU} ~\citep{kim2019abstractive} is a dataset with less formal posts collected from the online discussion forum Reddit.
Detailed statistics and introduction of datasets are presented in the appendix section.

\section{Evaluation for Summarization}
Existing summarization systems are usually evaluated on different datasets individually based on an automatic metric: $r = \mathrm{eval}(D, S, m)$, 
where $D$, $S$ represents a dataset (e.g., \texttt{CNNDM}) and system (e.g., \texttt{L2L}) respectively. $m$ denotes an evaluation metric (e.g., ROUGE).

\begin{figure}[ht]%
\centering
\includegraphics[width=0.42\linewidth]{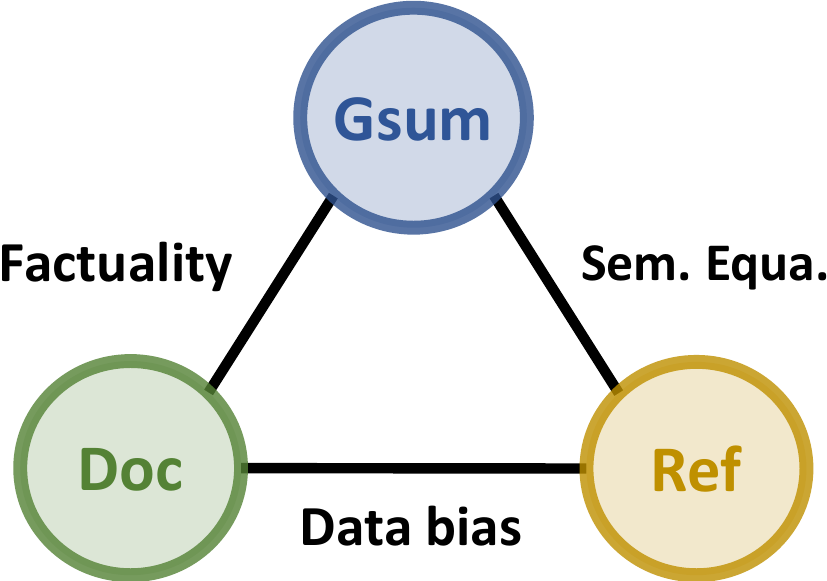}
\caption{Different metrics characterized by a relation chart among generated summaries (\texttt{Gsum}), references (\texttt{Ref}) and input documents (\texttt{Doc}).}
\label{fig:threerelation}
\end{figure}

To evaluate the quality of generated summaries, metrics can be designed from diverse perspectives, which can be abstractly characterized in Fig.~\ref{fig:threerelation}. Specifically, \textit{semantic equivalence} is used to quantify the relation between generated summaries (\texttt{Gsum}) and references (\texttt{Ref}) while \textit{factuality} aims to characterize the relation between  generated summaries (\texttt{Gsum}) and input documents (\texttt{Doc}).

Besides evaluation metrics, in this paper, we also introduce some measures that quantify the relation between input documents (\texttt{Doc}) and references (\texttt{Ref}). We claim that a better understanding of dataset biases can help us interpret models' discrepancies.

\subsection{Semantic Equivalence}

ROUGE ~\citep{lin2004rouge} is a classic metric to evaluate the quality of model generated summaries by counting the number of overlapped $n$-grams between the evaluated summaries and the ideal references.

\begin{figure*}[ht]
  \centering
  \subfloat[CNN.]{
    \label{fig:cnndm}
    \includegraphics[width=0.19\linewidth]{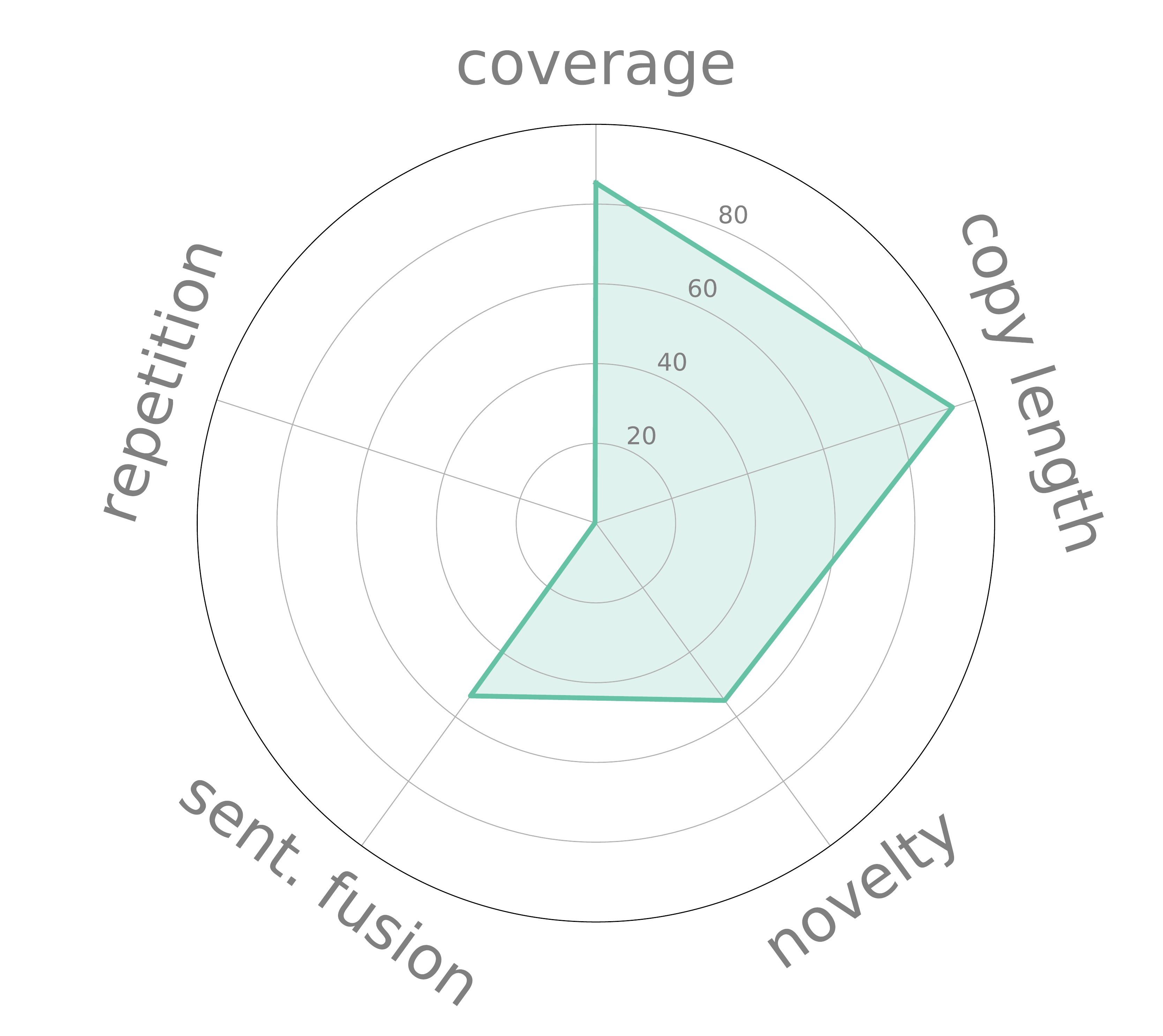}
  }
  \subfloat[Xsum]{
    \label{fig:xsum}
    \includegraphics[width=0.19\linewidth]{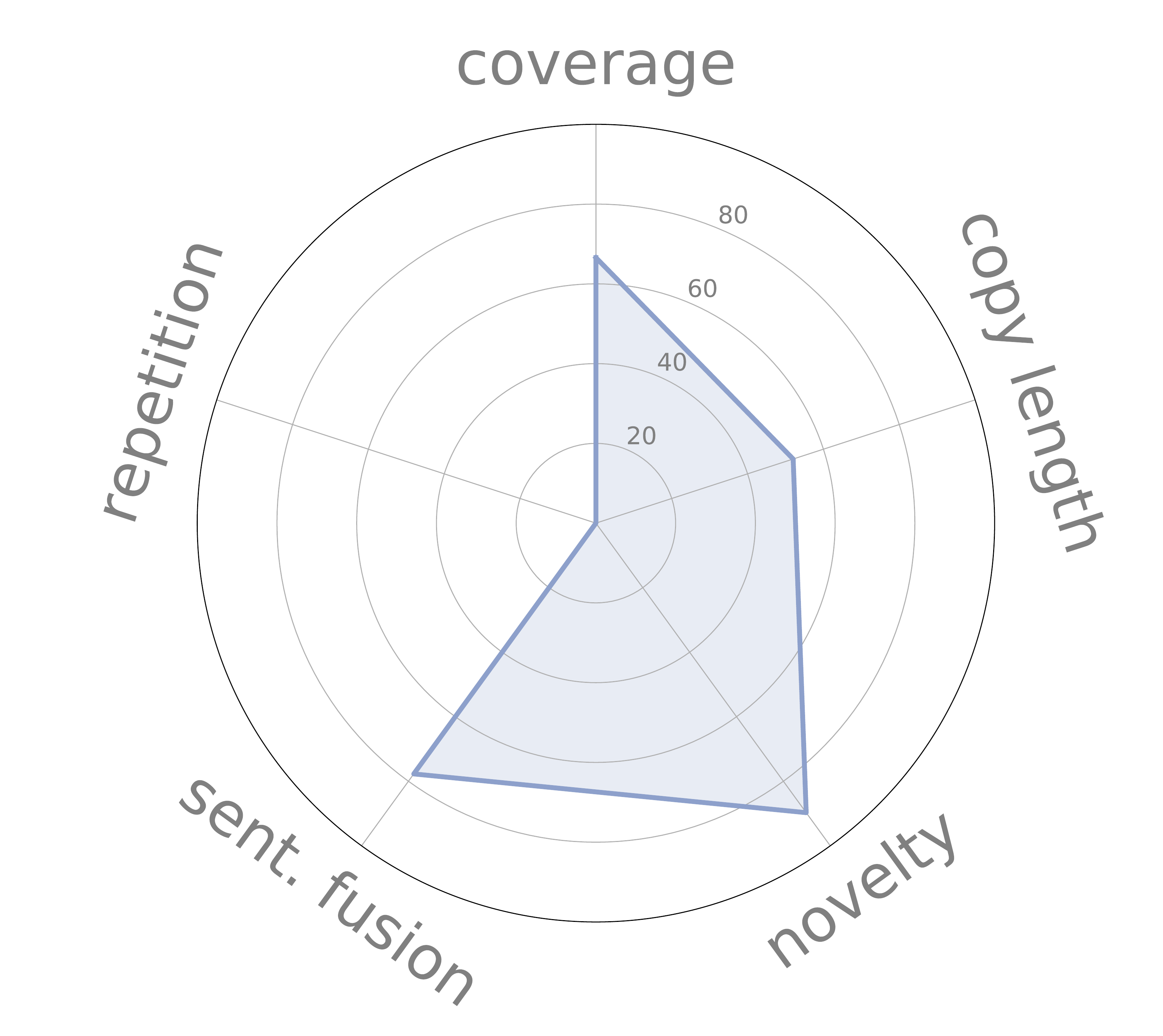}
  }
  \subfloat[PubMed]{
    \label{fig:PubMed}
    \includegraphics[width=0.19\linewidth]{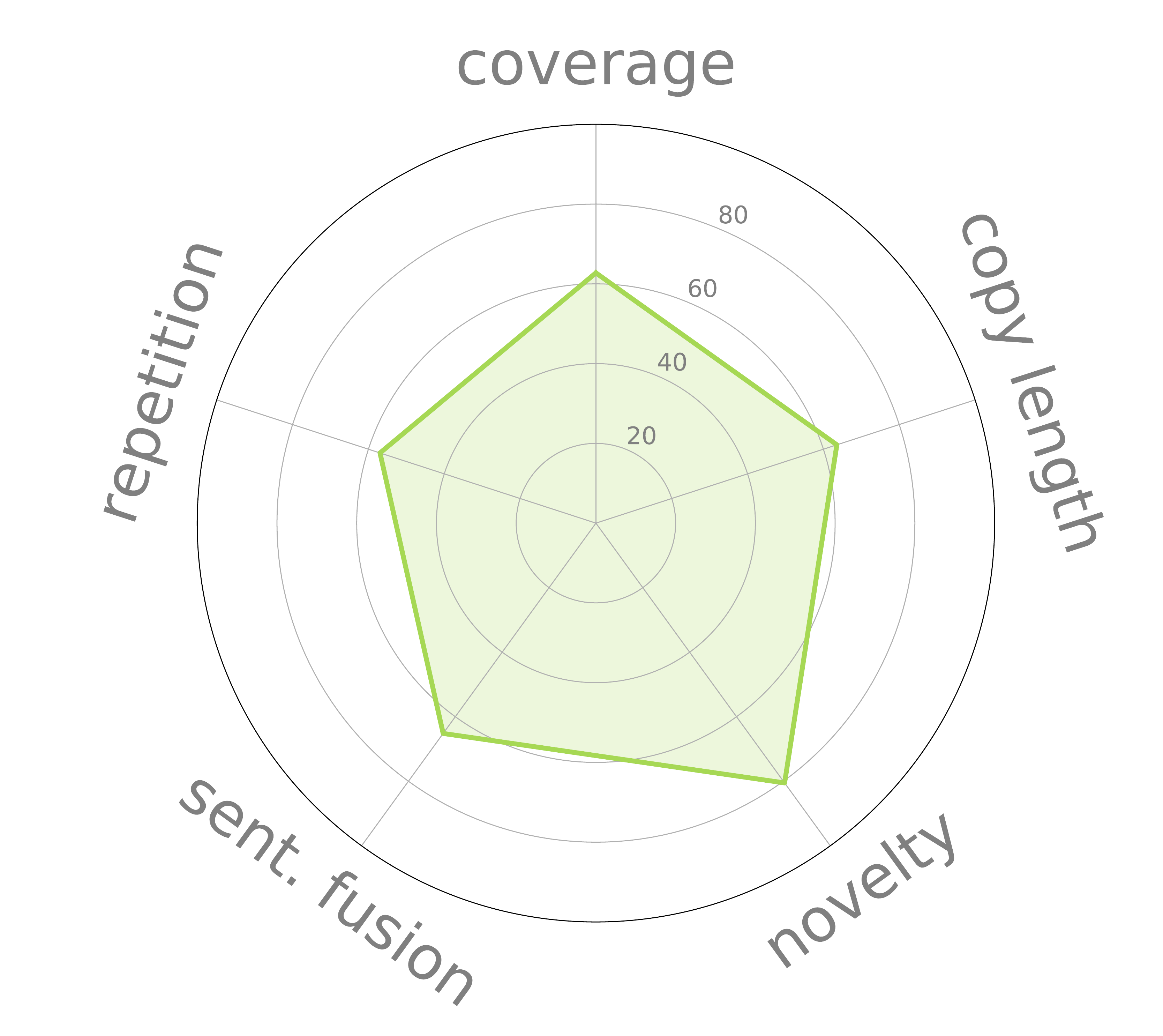}
  }
  \subfloat[Bigatent b]{
    \label{fig:bigpatent_b}
    \includegraphics[width=0.19\linewidth]{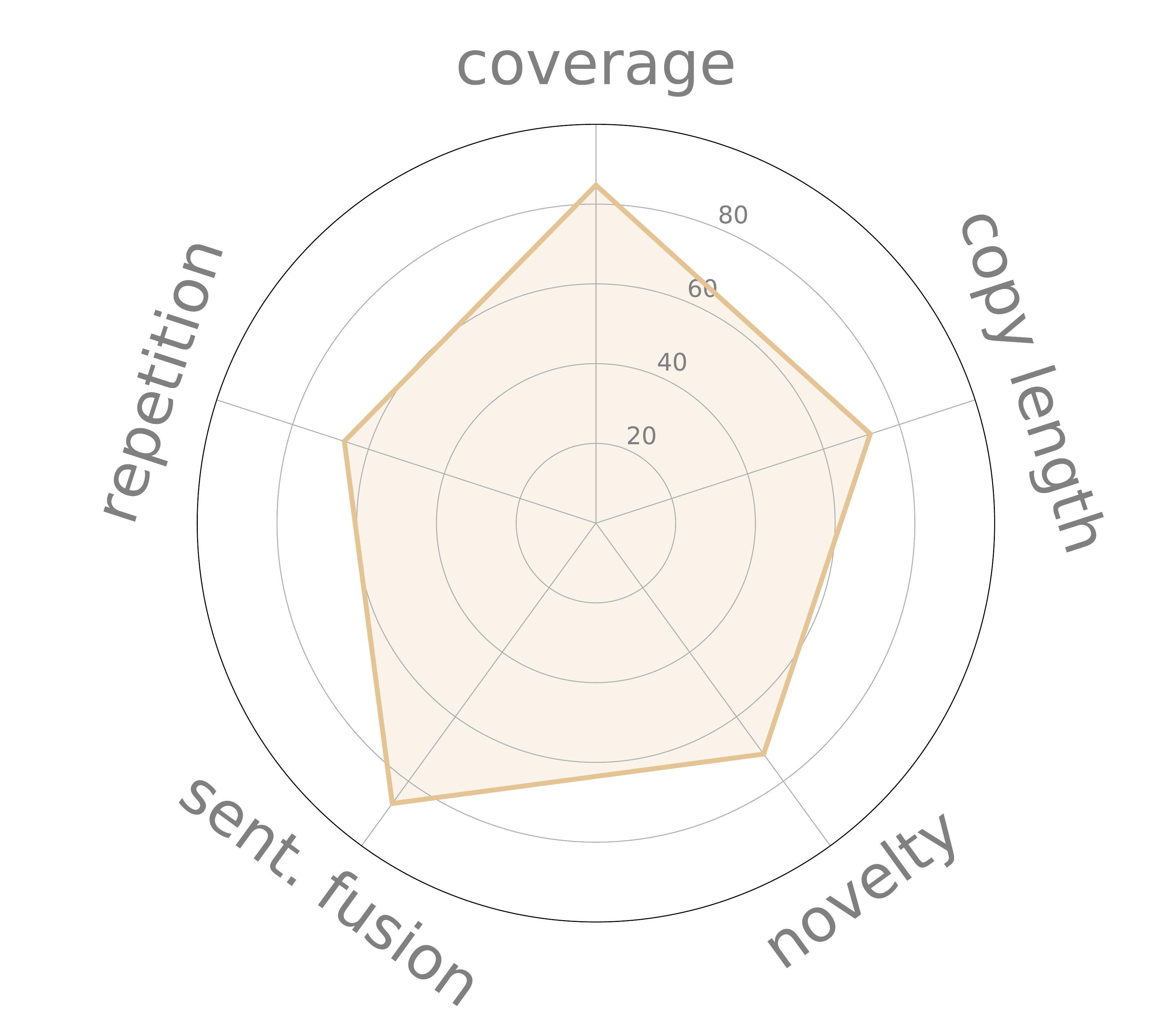}
  }
  \subfloat[Reddit]{
    \label{fig:reddit}
    \includegraphics[width=0.19\linewidth]{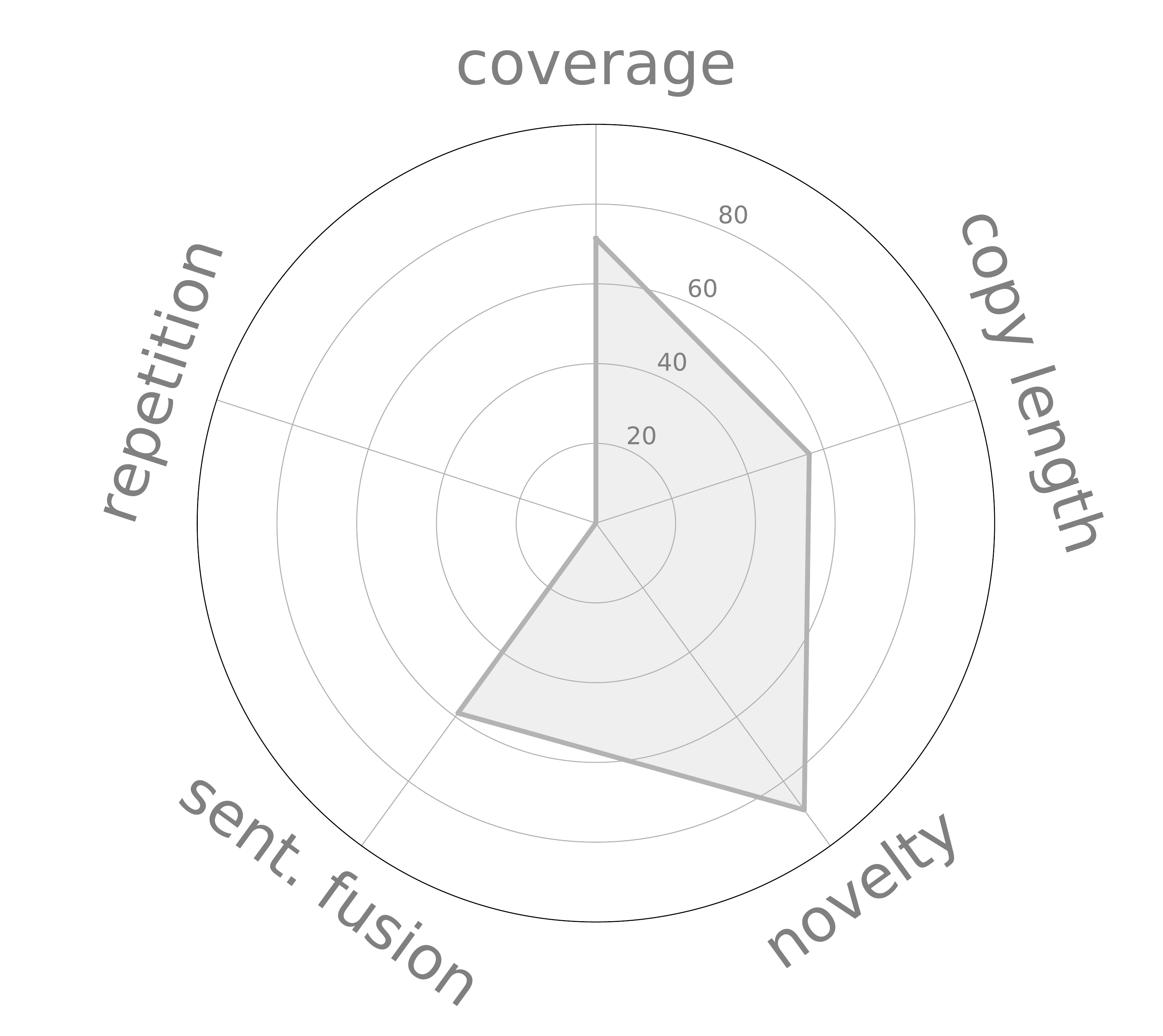}
  }

 \caption{Characteristics of test set for each dataset (the train set possesses almost the same property thus is not displayed here): \emph{coverage},\emph{copy length}, \emph{novelty}, \emph{sentence fusion score}, \emph{repetition}. Here we choose 2-gram to calculate the novelty and 3-gram for the repetition. }
 \label{fig:datasets}
\end{figure*}

\subsection{Factuality}
Apart from evaluating the semantic equivalence between generated summaries and the references, 
another evaluation aspect of recent interest is \textit{factuality}. 
In order to analyze the generalization performance of models in different perspectives, in this work, we also take the factuality evaluation into consideration.

\paragraph{Factcc} \textit{Factcc} ~\citep{kryscinski2019evaluating} is introduced to measure the fact consistency between the generated summaries and source documents. It is a model based metric which is weakly-supervised. 
We use the proportion of summary sentences that factcc predicts as factually consistent as the factuality score in this paper.

\subsection{Dataset Bias}
\label{dataset bias}
We detail several measures that could quantify the characteristics of datasets,
which are helpful for us to understand the differences among models.

\noindent \textbf{Coverage} ~\citep{grusky2018newsroom} illustrates the overlap rate between document and summary, it is defined as the proportion of the copied segments in summary. 

\noindent \textbf{Copy Length} measures the average length of segments in summary copied from source document. 

\noindent \textbf{Novelty}~\citep{see2017get} is defined as the proportion of segments in the summaries that haven't appeared in source documents. 
The segments can be instantiated as n-grams.

\noindent \textbf{Repetition}~\citep{see2017get} measures the rate of repeated segments in summaries. Similar to the above measure, we choose n-gram (n ranges from one to four) as segment unit.

\noindent \textbf{Sentence fusion score} is calculated using the result of the algorithm proposed by ~\citep{lebanoff2019scoring}, which is to find whether summary sentence is compressed from one sentence or fused from several sentences. 
Then, sentence fusion score is calculated as the proportion of \emph{fused sentences} (sentences that are fused from two or three document sentences) to all summary sentences.

A high value of coverage and copy length suggests the dataset is more extractive, while novelty represents the rate of novel units in summary and sentence fusion score represents the proportion of sentences that is fused from more than two document sentences. \citet{zhong2019closer} also explores dataset bias to aid the analysis of model performance, but they only focus on metrics for extractive summarizers.

\subsection{Dataset Bias Analysis}
\label{databias analysis}
According to the coverage and copy length results in Fig.~\ref{fig:datasets}, \texttt{CNNDM} is the most extractive dataset. \texttt{Bigpatent B} also exhibits relatively higher copy rate in summary but the copy segments is shorter than \texttt{CNNDM}.
On the other hand, \texttt{Bigaptent b}, \texttt{Xsum} obtain higher sentence fusion score, which suggests that the proportion of \emph{fused sentences} in these two datasets are high.
\texttt{Xsum} and \texttt{Reddit} obtain more 3-gram novel units in summary, reflecting these two datasets are more abstractive.
In terms of repetition in Fig.~\ref{fig:datasets}, only \texttt{PubMed} and \texttt{Bigpatent B} contain more 2-gram repeated phrases in summary.

\begin{table}[htbp]
  \setlength{\tabcolsep}{1.6pt}
\renewcommand{\arraystretch}{1.03}
  \centering
  \scriptsize
    \begin{tabular}{clcccccc}
    \toprule
    \multicolumn{2}{c}{\multirow{2}[4]{*}{\textbf{Models}}} & \multicolumn{6}{c}{\textbf{ROUGE 1}} \\
\cmidrule{3-8}    \multicolumn{2}{c}{} & \textbf{CNN.*} & \textbf{CNN.} & \textbf{Xsum} & \textbf{Pubm.} & \textbf{Patent b} & \textbf{Red.} \\
    \midrule
    \multirow{5}[2]{*}{Ext.} & LSTM$_{non}$ & 41.22  & 41.36  & 19.51  & 42.98  & 39.29  & 20.46  \\
          & Trans$_{non}$ & 40.90  & 40.84  & 15.74  & 38.45  & 34.41  & 16.25  \\
          & Trans$_{auto}$ & 41.36  & 41.35  & 19.29  & 42.74  & 38.76  & 18.55  \\
          & BERT$_{non}$ & 43.25  & 42.69  & 21.76  & 38.74  & 35.85  & 21.84  \\
          & BERT$_{match}$ & 44.22  & 44.26  & 24.97  & 41.19  & 38.89  & 25.32  \\
    \midrule
    \multirow{6}[2]{*}{Abs.} & L2L   & 31.33  & 32.80  & 28.31  & 27.84  & 30.46  & 16.89  \\
          & L2L$_{ptr}$ & 36.44  & 37.06  & 29.67  & 32.04  & 31.03  & 21.32  \\
          & L2L$_{ptr}^{cov}$ & 39.53  & 39.95  & 28.83  & 35.27  & 35.90  & 21.28  \\
          & T2T   & 40.21  & 39.90  & 29.01  & 30.71  & 42.94  & 19.96  \\
          & BE2T  & 41.72  & 41.34  & 38.99  & 37.11  & 43.10  & 26.66  \\
          & BART  & 44.16  & 44.75  & 44.73  & 45.02  & 45.78  & 34.00  \\
    \bottomrule
    \end{tabular}%
    \caption{Representative summarizers studied in this paper and their corresponding performance (ROUGE-1 F1 score) on different datasets (\texttt{CNNDM}, \texttt{Xsum}, \texttt{PubMed}, \texttt{Bigpatent B}, \texttt{Reddit}). We re-implement all 11 systems on five datasets by ourselves. All implemented results can outperform or slightly lower than the performances reported in original papers (the column of CNN.*).}
  \label{tab:in domain}%
\end{table}%

\begin{table}[t]
  \centering
      \setlength{\tabcolsep}{4pt}
    \renewcommand{\arraystretch}{1.3}  
    \begin{tabular}{cccrcccrccccc}
    & \multicolumn{2}{c}{$\mathbf{U}_{A}$} & & &  & \multicolumn{2}{c}{$\mathbf{U}_B$} &  & &   & \multicolumn{2}{c}{Measures} \\
    \multicolumn{1}{r}{} & a     & \multicolumn{1}{c}{b} &    &    & \multicolumn{1}{c}{} & a     & \multicolumn{1}{c}{b} &   &     & \multicolumn{1}{c}{} & $\mathbf{U}_A$    & \multicolumn{1}{c}{$\mathbf{U}_B$} \\
\cline{2-3}\cline{7-8}\cline{12-13}    a     & \multicolumn{1}{|c}{\cellcolor[HTML]{DAE8FC}48}    & \multicolumn{1}{c|}{40}  &   &       & \multicolumn{1}{c}{a}     & \multicolumn{1}{|c}{\cellcolor[HTML]{DAE8FC}61}    & \multicolumn{1}{c|}{43}  &   &       & \small{\textit{Stiff.}}     & \multicolumn{1}{|c}{44}   & \multicolumn{1}{c|}{\cellcolor[HTML]{F8A102}55} \\
    b     & \multicolumn{1}{|c}{41}    & \multicolumn{1}{c|}{\cellcolor[HTML]{DAE8FC}45}  &   &       & b     & \multicolumn{1}{|c}{46}    & \multicolumn{1}{c|}{\cellcolor[HTML]{DAE8FC}69} &    &       & \small{\textit{Stable.}} & \multicolumn{1}{|c}{94}   & \multicolumn{1}{c|}{\cellcolor[HTML]{F8A102}84} \\
\cline{2-3}\cline{7-8}\cline{12-13}    \end{tabular}%
    \caption{Illustration of two views (\textit{Stiffness}: $r^u$ and \textit{Stableness}: ${r}^{\sigma}$) to characterize the cross-dataset (a and b) generalization based on model $A$ and $B$. $\mathbf{U_A}$ and $\mathbf{U_B}$ represent two cross-dataset matrix of two models.  $r^{\mu}(\mathbf{U_A}) < r^{\mu}(\mathbf{U_B})$ means the model $B$ gains a better cross-dataset absolute performance  while $r^{\sigma}(\mathbf{U_A}) > r^{\sigma}(\mathbf{U_B})$ suggests the model $A$ is more robust.} 
      \label{tab:twoviews}%
\end{table}%

\section{Cross-dataset Evaluation}
\label{measures}
Despite recent impressive results on diverse summarization datasets, modern summarization systems mainly focus on extensive in-dataset architecture engineering while ignore the generalization ability which is 
indispensable when systems are required to process samples from new datasets or domains. 
Therefore, instead of evaluating the quality of summarization system solely based on one dataset, we introduce cross-dataset evaluation (a summarizer (e.g., \textit{L2L}) trained on one dataset (e.g., \texttt{CNNDM}) will be evaluated on a range of other datasets (e.g., \texttt{XSUM})).
Methodologically, we perform cross-dataset evaluation from two views: fine-grained and holistic and we will detail them below.

\subsection{Methodology}
Given a summarization system $S$, a set of datasets $\mathcal{D} = D_1, \cdots, D_N$, and evaluation metric $m$,  we can design different evaluation function to quantify the system's quality: $\mathbf{r} = \mathrm{eval}(\mathcal{D}, S, m)$. 
Depending on different forms of function $\mathrm{eval}(\cdot)$, $\mathbf{r}$ could be instantiated as either a scalar or a vector (or matrix).

\subsubsection{Fine-grained Measures}
Once $\mathbf{r}$, the cross-dataset evaluation result, is instantiated as a matrix, we can characterize the given system in a fine-grained way.
Specifically, we define $\mathbf{r}$ as: $\mathbf{r} = \mathbf{U} \in \mathbb{R}^{N\times N}$
where each cell $\mathbf{U}_{i,j}$ refers to the metric result (e.g., ROUGE) when a summarizer is trained in dataset ${D}_i$ and tested in dataset ${D}_j$ (N refers to the number of datasets).

Additionally, we can normalize each cell by the diagonal value, $\mathbf{r} =  \mathbf{U}_{ij}/\mathbf{U}_{jj} \times 100\% 
           =\mathbf{\hat{U}}$,  
$\mathbf{U}_{ij}/\mathbf{U}_{jj}$ measures how close the out-of-dataset performance (trained in ${D}_i$ and tested in ${D}_j$) of a system is to its in-dataset performance (trained in ${D}_j$ and tested in ${D}_j$).

\subsubsection{Holistic Measures}
\label{sec:holistic measures}
Instead of using a matrix, holistically, we can quantify the cross-dataset generalization ability of each summarization system using a scalar.
Specifically, we propose two views to characterize the cross-dataset generalization.

\paragraph{Stiffness}
This measure reflects the absolute performance of a system under cross-dataset setting. Given a system, its \emph{stiffness} can be calculated as: ${r}^{\mu} = \frac{1}{N\times N}\sum_{i,j} {\mathbf{U}}_{ij}$

Intuitively, a higher value of \textit{stiffness} suggests the system obtains better performance when transferred to new datasets.

\paragraph{Stableness}
It characterizes the relative performance gap between in-dataset and cross-dataset test. ${r}^{\sigma} = \frac{1}{N\times N}\sum_{i,j} \mathbf{U}_{ij}/\mathbf{U}_{jj} \times 100\%$

Generally, a higher value of \textit{stableness} suggests that the variance between in-dataset and cross-dataset results is smaller. 

Tab.~\ref{tab:twoviews} gives an example to characterize generalization ability in two views. It shows that stiffness and stableness are not always unanimous, a model with higher stiffness may obtains lower stableness.

\begin{table*}[htbp]
  \centering
  \bf
  \Huge
  \captionsetup[subfloat]{font=Huge} 
  \setlength{\tabcolsep}{3pt}
  \renewcommand{\arraystretch}{1.23}
  \resizebox{1.005\textwidth}{28mm}{
    \begin{tabular}{c|c|rrllllllrllllllrllllllrllllllrllllllrllllll}
    \toprule
    \multicolumn{3}{c}{analysis aspect} &       & \multicolumn{28}{c}{Architecture}                                                                                                                                                                                             & \multicolumn{13}{c}{Generation way} \\
\cmidrule{1-3}\cmidrule{5-45}    \multicolumn{3}{c}{model type} &       & \multicolumn{13}{c}{EXT}                                                                              &       & \multicolumn{13}{c}{ABS}                                                                              &       & \multicolumn{6}{c}{LSTM}                      &       & \multicolumn{6}{c}{BERTSUM} \\
\cmidrule{1-3}\cmidrule{5-17}\cmidrule{19-31}\cmidrule{33-38}\cmidrule{40-45}    \multicolumn{3}{c}{compare models} &       & \multicolumn{6}{c}{BERT$_{match}$ vs. BERT$_{non}$}   &       & \multicolumn{6}{c}{BERT$_{non}$ vs. Trans$_{non}$}      &       & \multicolumn{6}{c}{L2L$_{ptr}$ vs. L2L}            &       & \multicolumn{6}{c}{L2L$_{ptr}^{cov}$ vs. L2L$_{ptr}$}      &       & \multicolumn{6}{c}{LSTM$_{non}$  vs.  L2L}         &       & \multicolumn{6}{c}{BERT$_{non}$ vs. BE2T} \\
\cmidrule{1-10}\cmidrule{12-17}\cmidrule{19-24}\cmidrule{26-31}\cmidrule{33-38}\cmidrule{40-45}    \multicolumn{3}{c}{\multirow{2}[2]{*}{holistic analysis}} &       & \multicolumn{6}{c}{stiff. : 32.27 vs. 28.98}   &       & \multicolumn{6}{c}{stiff. : 28.98 vs. 28.02}   &       & \multicolumn{6}{c}{stiff. : 20.74 vs. 18.03}   &       & \multicolumn{6}{c}{stiff. : 22.81 vs. 20.74}   &       & \multicolumn{6}{c}{stiff. : 28.51 vs. 18.03}   &       & \multicolumn{6}{c}{stiff. : 28.98 vs. 23.49} \\
    \multicolumn{3}{c}{}  &       & \multicolumn{6}{c}{stable. : 91.98 vs. 88.93} &       & \multicolumn{6}{c}{stable. : 88.93 vs. 99.05} &       & \multicolumn{6}{c}{stable. : 68.63 vs. 66.93} &       & \multicolumn{6}{c}{stable. : 70.71 vs. 68.63}  &       & \multicolumn{6}{c}{stable. : 87.00 vs. 66.93} &       & \multicolumn{6}{c}{stable. : 88.93 vs. 62.93} \\
\cmidrule{1-3}\cmidrule{5-10}\cmidrule{12-17}\cmidrule{19-24}\cmidrule{26-31}\cmidrule{33-38}\cmidrule{40-45}    \multicolumn{3}{c}{fine-grain analysis} &       & \multicolumn{1}{c}{CNN.} & \multicolumn{1}{c}{Xsum} & \multicolumn{1}{c}{Pubm.} & \multicolumn{1}{c}{Patent b} & \multicolumn{1}{c}{Red.} & \multicolumn{1}{c}{\textcolor[rgb]{ 1,  0,  0}{avg}} &       & \multicolumn{1}{c}{CNN.} & \multicolumn{1}{c}{Xsum} & \multicolumn{1}{c}{Pubm.} & \multicolumn{1}{c}{Patent b} & \multicolumn{1}{c}{Red.} & \multicolumn{1}{c}{\textcolor[rgb]{ 1,  0,  0}{avg}} &       & \multicolumn{1}{c}{CNN.} & \multicolumn{1}{c}{Xsum} & \multicolumn{1}{c}{Pubm.} & \multicolumn{1}{c}{Patent b} & \multicolumn{1}{c}{Red.} & \multicolumn{1}{c}{\textcolor[rgb]{ 1,  0,  0}{avg}} &       & \multicolumn{1}{c}{CNN.} & \multicolumn{1}{c}{Xsum} & \multicolumn{1}{c}{Pubm.} & \multicolumn{1}{c}{Patent b} & \multicolumn{1}{c}{Red.} & \multicolumn{1}{c}{\textcolor[rgb]{ 1,  0,  0}{avg}} &       & \multicolumn{1}{c}{CNN.} & \multicolumn{1}{c}{Xsum} & \multicolumn{1}{c}{Pubm.} & \multicolumn{1}{c}{Patent b} & \multicolumn{1}{c}{Red.} & \multicolumn{1}{c}{\textcolor[rgb]{ 1,  0,  0}{avg}} &       & \multicolumn{1}{c}{CNN.} & \multicolumn{1}{c}{Xsum} & \multicolumn{1}{c}{Pubm.} & \multicolumn{1}{c}{Patent b} & \multicolumn{1}{c}{Red.} & \multicolumn{1}{c}{\textcolor[rgb]{ 1,  0,  0}{avg}} \\
\cmidrule{1-10}\cmidrule{12-17}\cmidrule{19-24}\cmidrule{26-31}\cmidrule{33-38}\cmidrule{40-45}    \multirow{14}[4]{*}{\rotatebox{90}{ROUGE}} & \multirow{7}[2]{*}{\rotatebox{90}{origin}} & CNN.  &       & \multicolumn{6}{l}{\multirow{7}[2]{*}[8pt]{\subfloat[ ]{         \includegraphics[height=200pt,width=420pt]{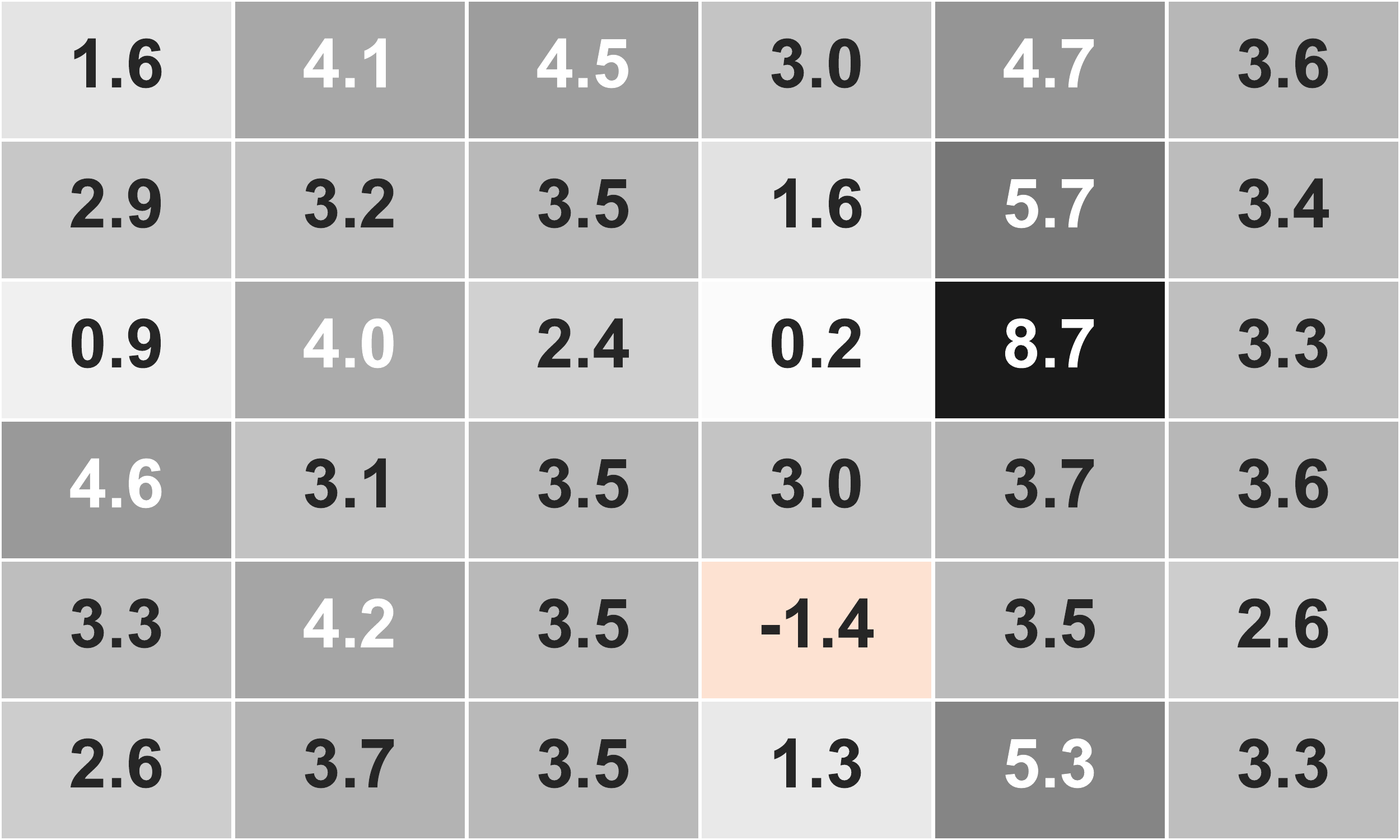} \label{fig:rouge origin matchsum_vs_ext_bert} }}}  &       & \multicolumn{6}{l}{\multirow{7}[2]{*}[8pt]{\subfloat[ ]{         \includegraphics[height=200pt,width=420pt]{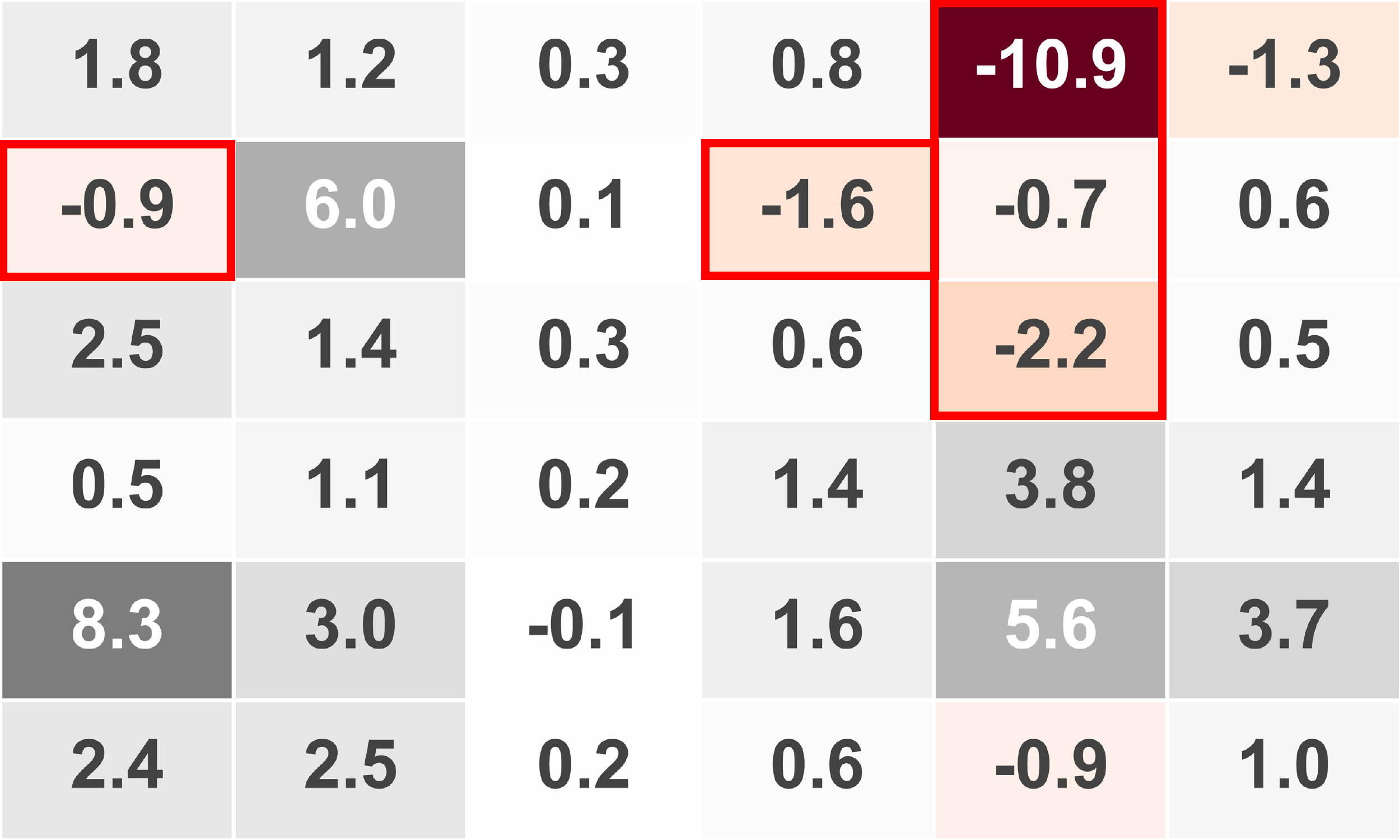} \label{fig:rouge origin ext_BERT_vs_baseline_ext_BERT} }}}  &       & \multicolumn{6}{l}{\multirow{7}[2]{*}[8pt]{\subfloat[ ]{         \includegraphics[height=200pt,width=420pt]{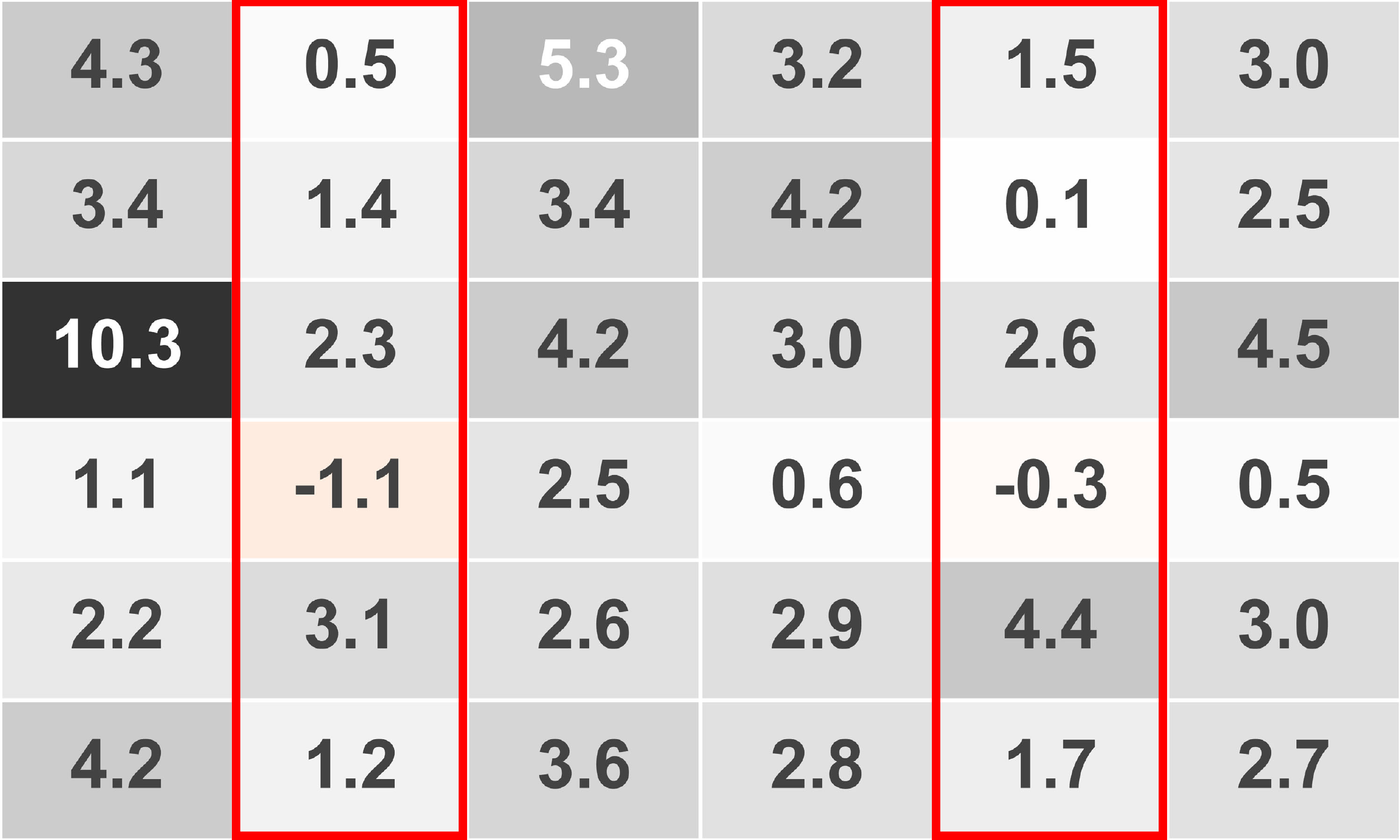} \label{fig:rouge origin LSTM_pointer_gen_vs_LSTM} }}}  &       & \multicolumn{6}{l}{\multirow{7}[2]{*}[8pt]{\subfloat[ ]{         \includegraphics[height=200pt,width=420pt]{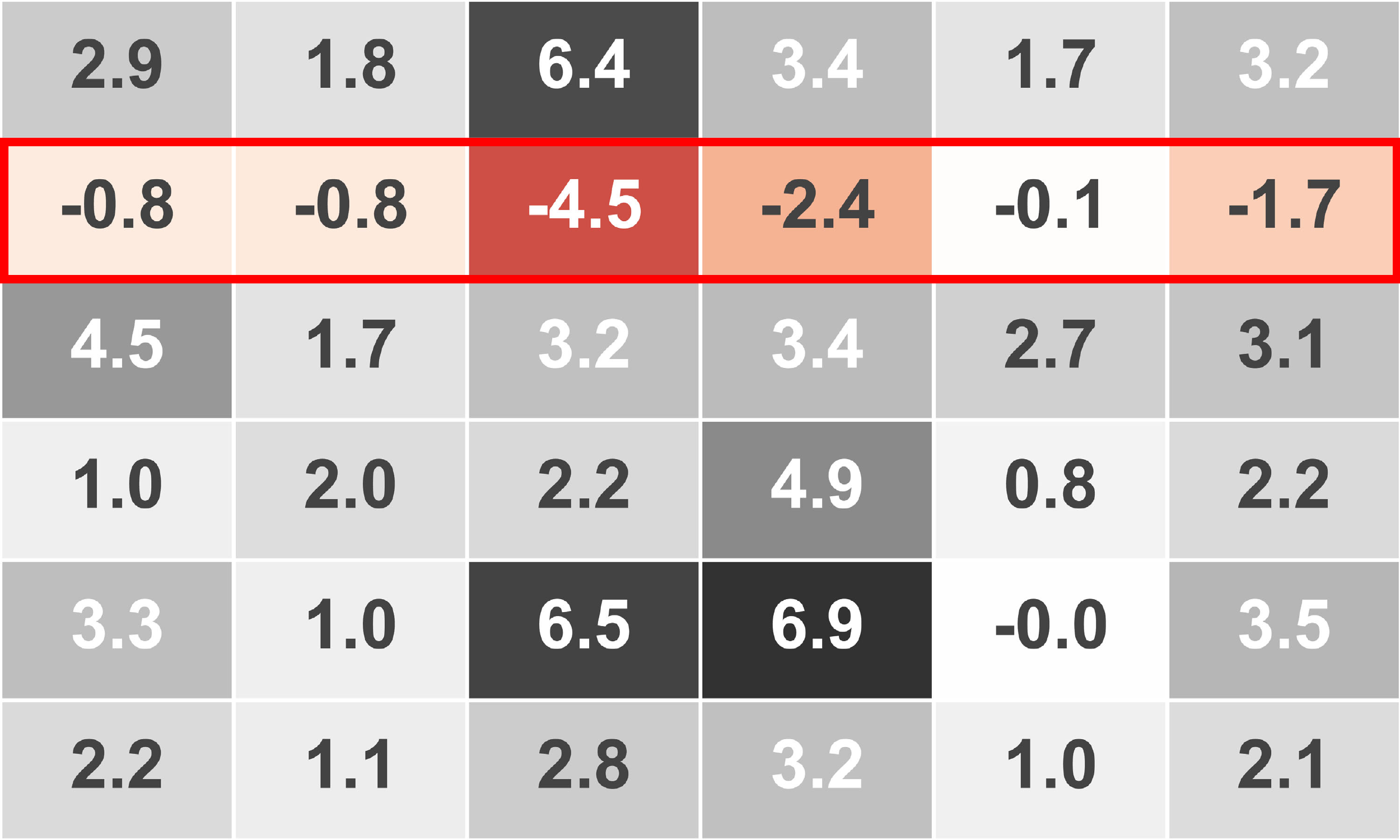} \label{fig:rouge origin LSTM_pointer_gen_coverage_vs_LSTM_pointer_gen} }}}  &       & \multicolumn{6}{l}{\multirow{7}[2]{*}[8pt]{\subfloat[ ]{         \includegraphics[height=200pt,width=420pt]{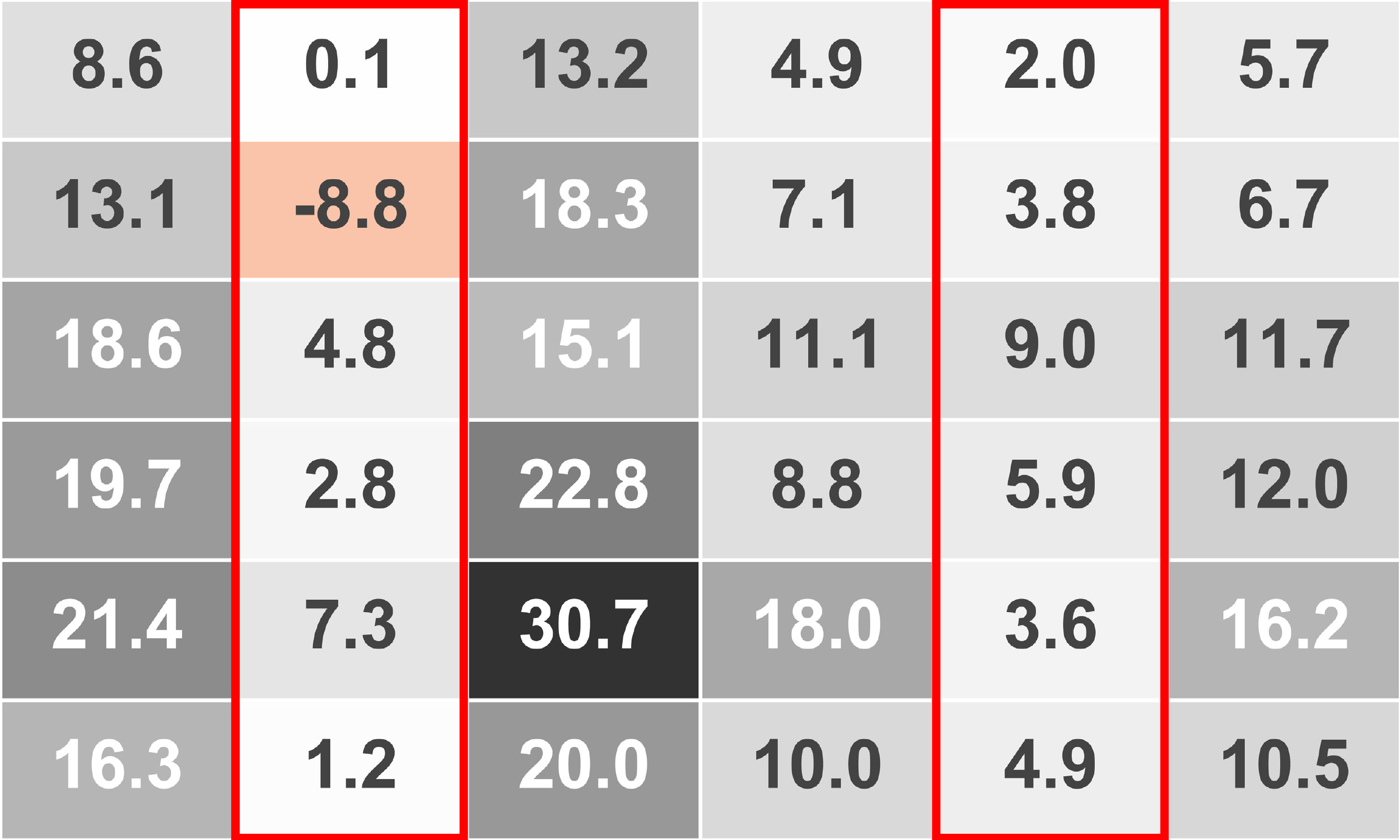} \label{fig:rouge origin BiLSTM_SequenceLabelling_vs_LSTM} }}}  &       & \multicolumn{6}{l}{\multirow{7}[2]{*}[8pt]{\subfloat[ ]{         \includegraphics[height=200pt,width=420pt]{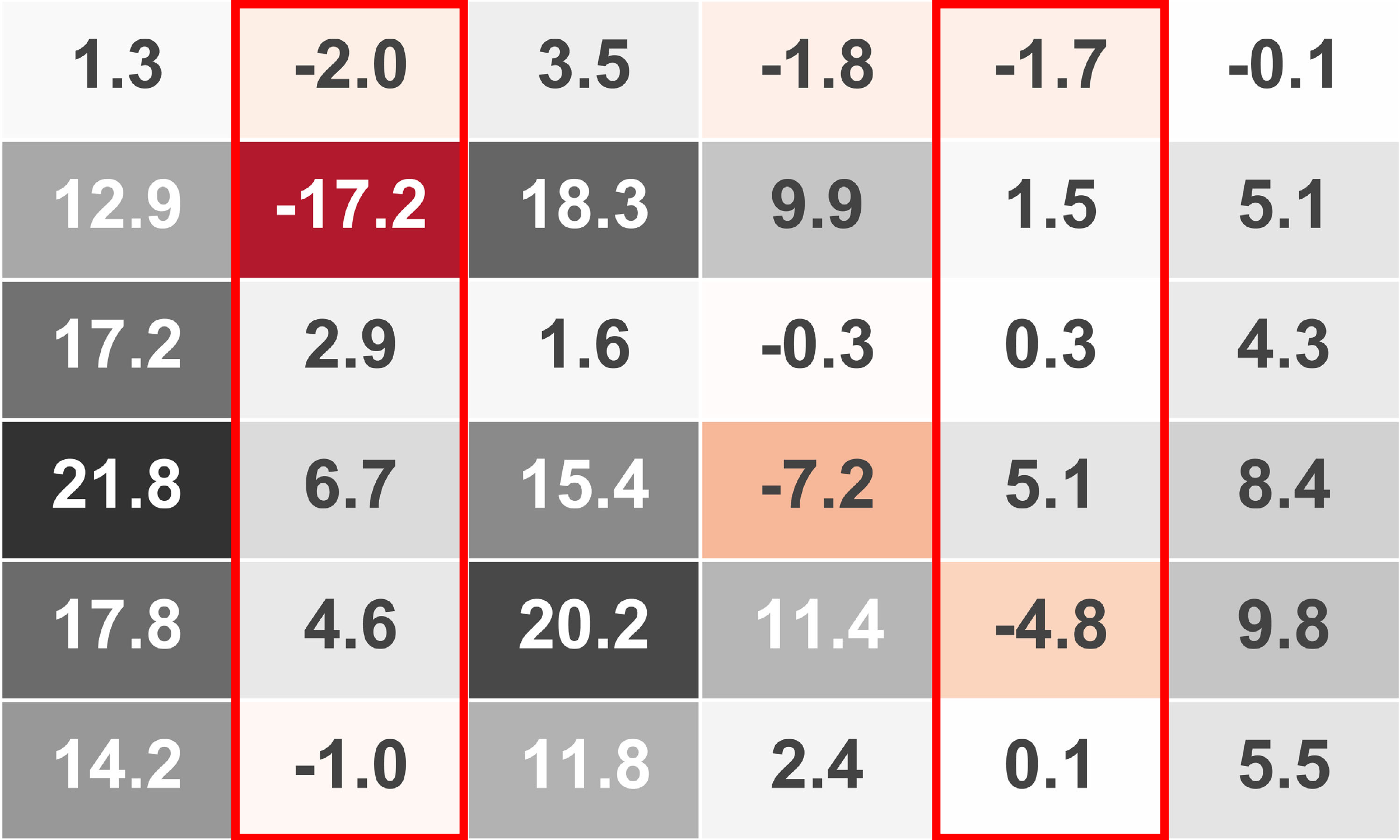} \label{fig:rouge origin ext_bert_vs_abs_bert} }}} \\
          &       & Xsum  &       & \multicolumn{6}{l}{}                          &       & \multicolumn{6}{l}{}                          &       & \multicolumn{6}{l}{}                          &       & \multicolumn{6}{l}{}                          &       & \multicolumn{6}{l}{}                          &       & \multicolumn{6}{l}{} \\
          &       & Pubm. &       & \multicolumn{6}{l}{}                          &       & \multicolumn{6}{l}{}                          &       & \multicolumn{6}{l}{}                          &       & \multicolumn{6}{l}{}                          &       & \multicolumn{6}{l}{}                          &       & \multicolumn{6}{l}{} \\
          &       & Patent b &       & \multicolumn{6}{l}{}                          &       & \multicolumn{6}{l}{}                          &       & \multicolumn{6}{l}{}                          &       & \multicolumn{6}{l}{}                          &       & \multicolumn{6}{l}{}                          &       & \multicolumn{6}{l}{} \\
          &       & Red.  &       & \multicolumn{6}{l}{}                          &       & \multicolumn{6}{l}{}                          &       & \multicolumn{6}{l}{}                          &       & \multicolumn{6}{l}{}                          &       & \multicolumn{6}{l}{}                          &       & \multicolumn{6}{l}{} \\
          &       & \textcolor[rgb]{ 1,  0,  0}{avg} &       & \multicolumn{6}{l}{}                          &       & \multicolumn{6}{l}{}                          &       & \multicolumn{6}{l}{}                          &       & \multicolumn{6}{l}{}                          &       & \multicolumn{6}{l}{}                          &       & \multicolumn{6}{l}{} \\
          &       &       &       & \multicolumn{6}{l}{}                          &       & \multicolumn{6}{l}{}                          &       & \multicolumn{6}{l}{}                          &       & \multicolumn{6}{l}{}                          &       & \multicolumn{6}{l}{}                          &       & \multicolumn{6}{l}{} \\
\cmidrule{5-10}\cmidrule{12-24}\cmidrule{26-31}\cmidrule{33-38}\cmidrule{40-45}          & \multirow{7}[2]{*}{\rotatebox{90}{normali.}} & CNN.  &       & \multicolumn{6}{l}{\multirow{7}[2]{*}[8pt]{\subfloat[ ]{         \includegraphics[height=200pt,width=420pt]{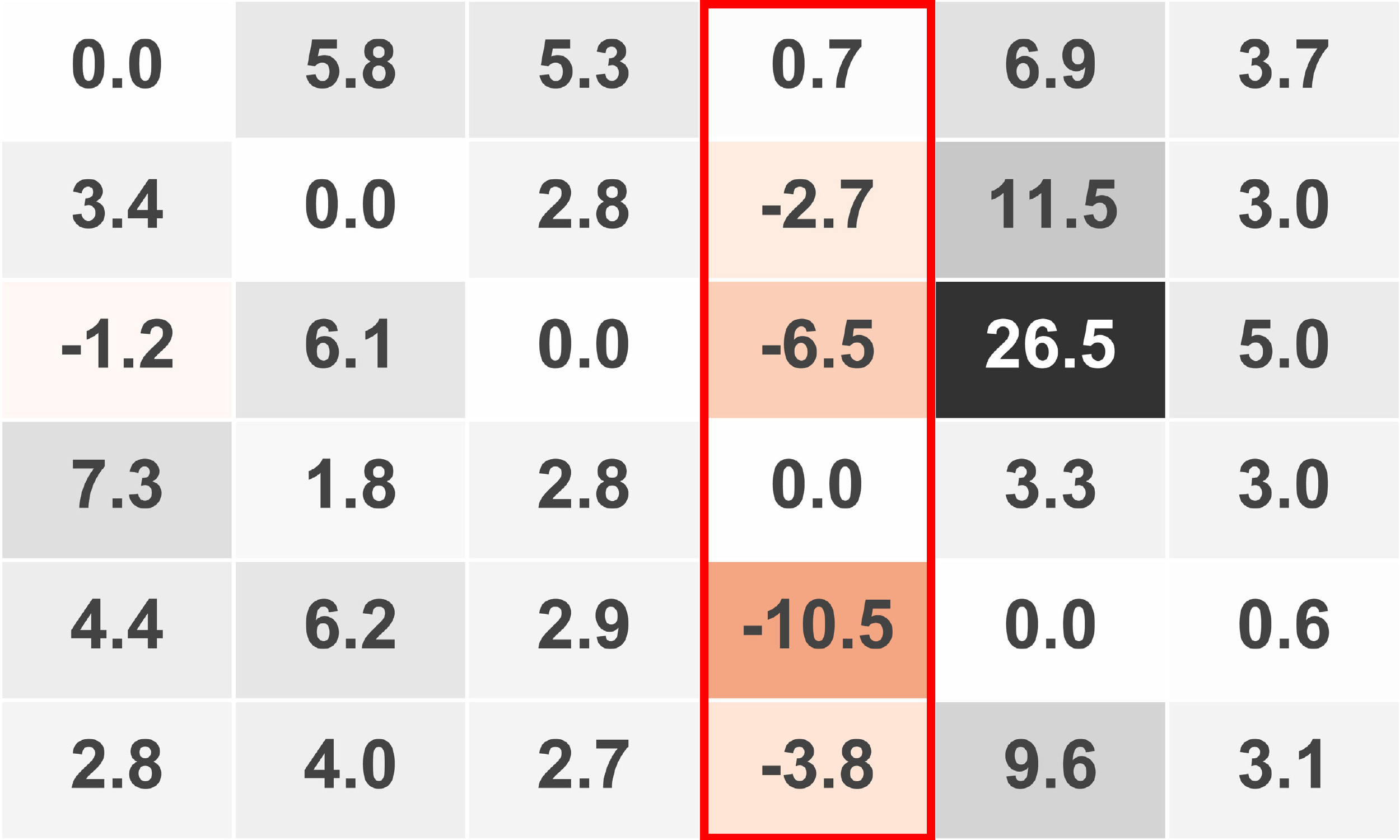} \label{fig:rouge normalized matchsum_vs_ext_bert} }}}  &       & \multicolumn{6}{l}{\multirow{7}[2]{*}[8pt]{\subfloat[ ]{         \includegraphics[height=200pt,width=420pt]{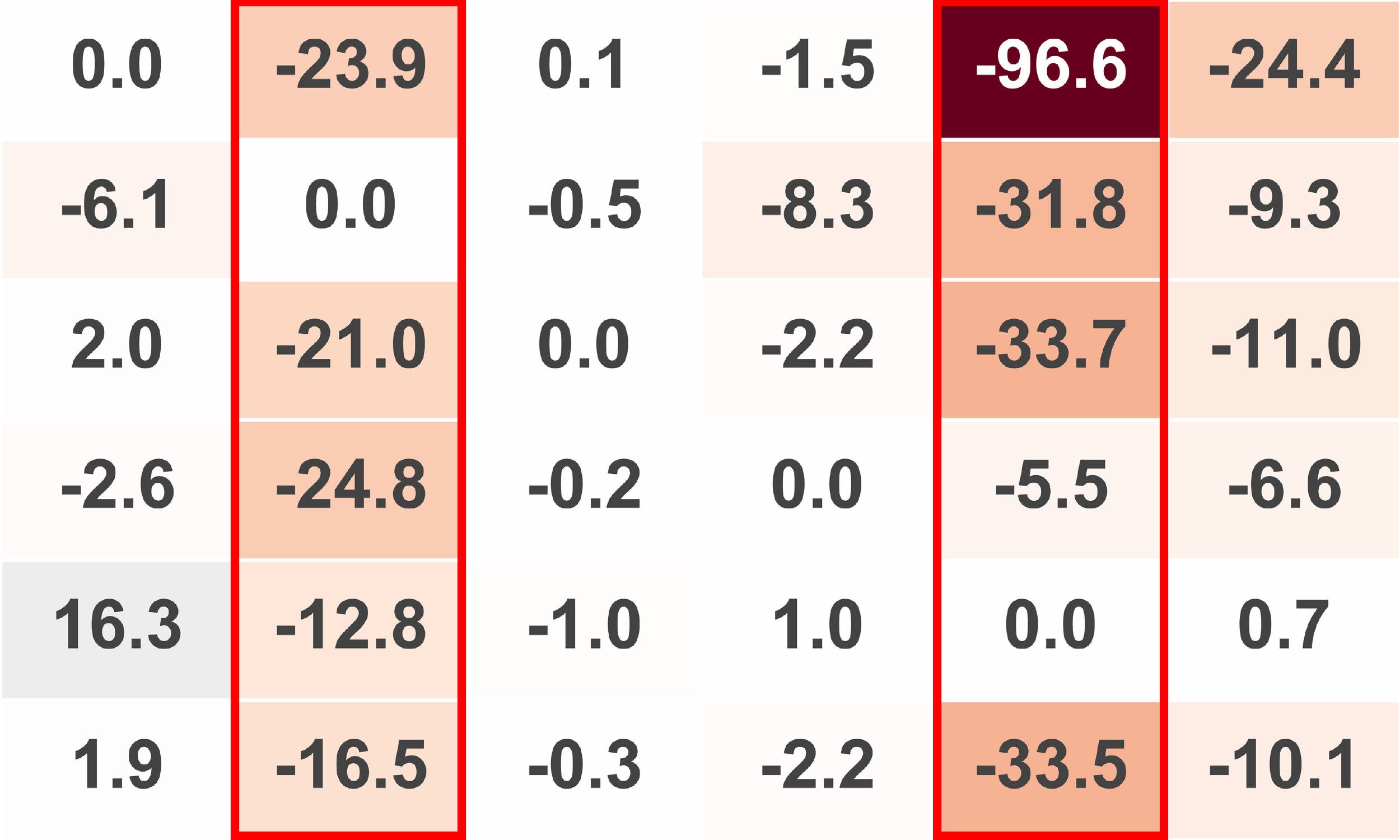} \label{fig:rouge normalized ext_BERT_vs_baseline_ext_BERT} }}}  &       & \multicolumn{6}{l}{\multirow{7}[2]{*}[8pt]{\subfloat[ ]{         \includegraphics[height=200pt,width=420pt]{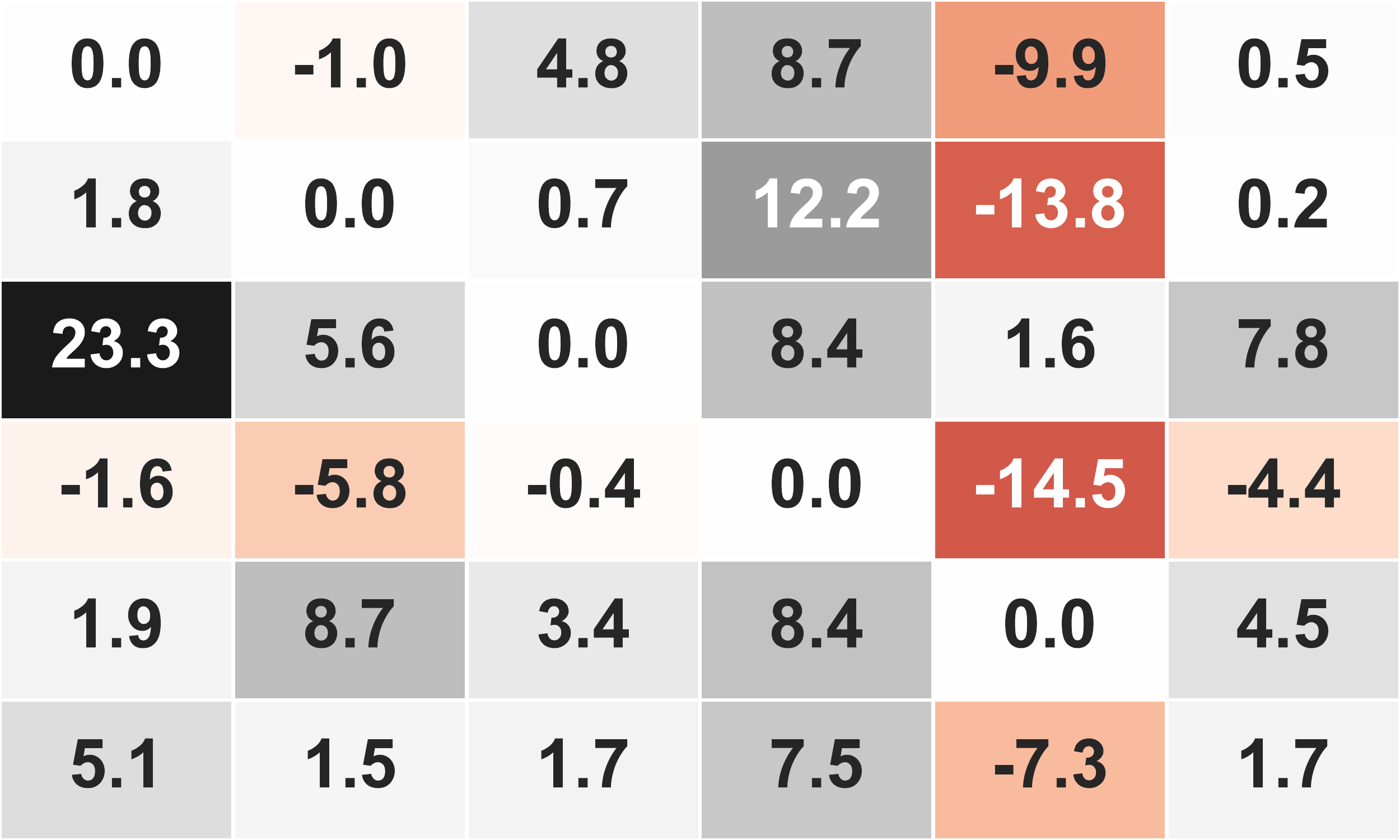} \label{fig:rouge normalized LSTM_pointer_gen_vs_LSTM} }}}  &       & \multicolumn{6}{l}{\multirow{7}[2]{*}[8pt]{\subfloat[ ]{         \includegraphics[height=200pt,width=420pt]{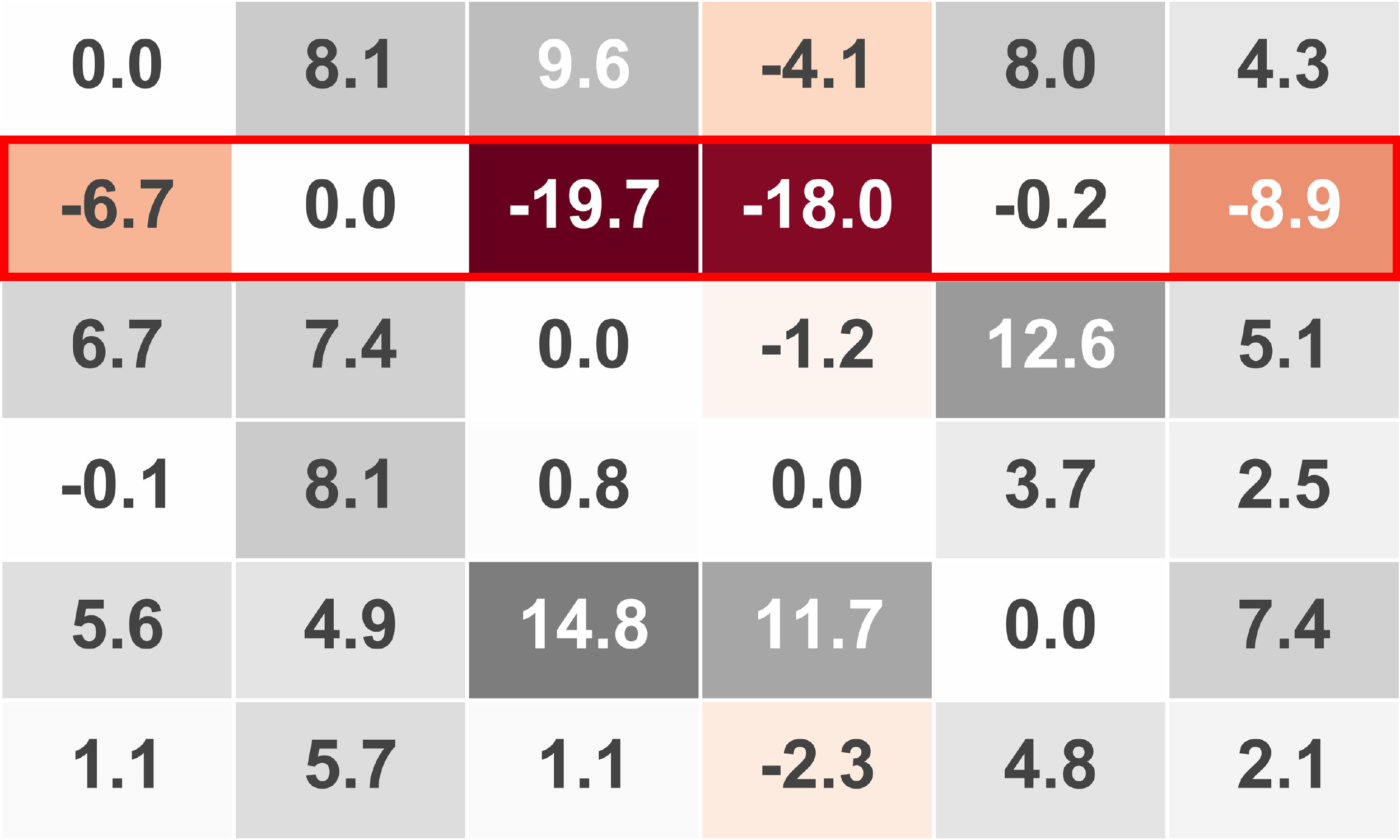} \label{fig:rouge normalized LSTM_pointer_gen_coverage_vs_LSTM_pointer_gen} }}}  &       & \multicolumn{6}{l}{\multirow{7}[2]{*}[8pt]{\subfloat[ ]{         \includegraphics[height=200pt,width=420pt]{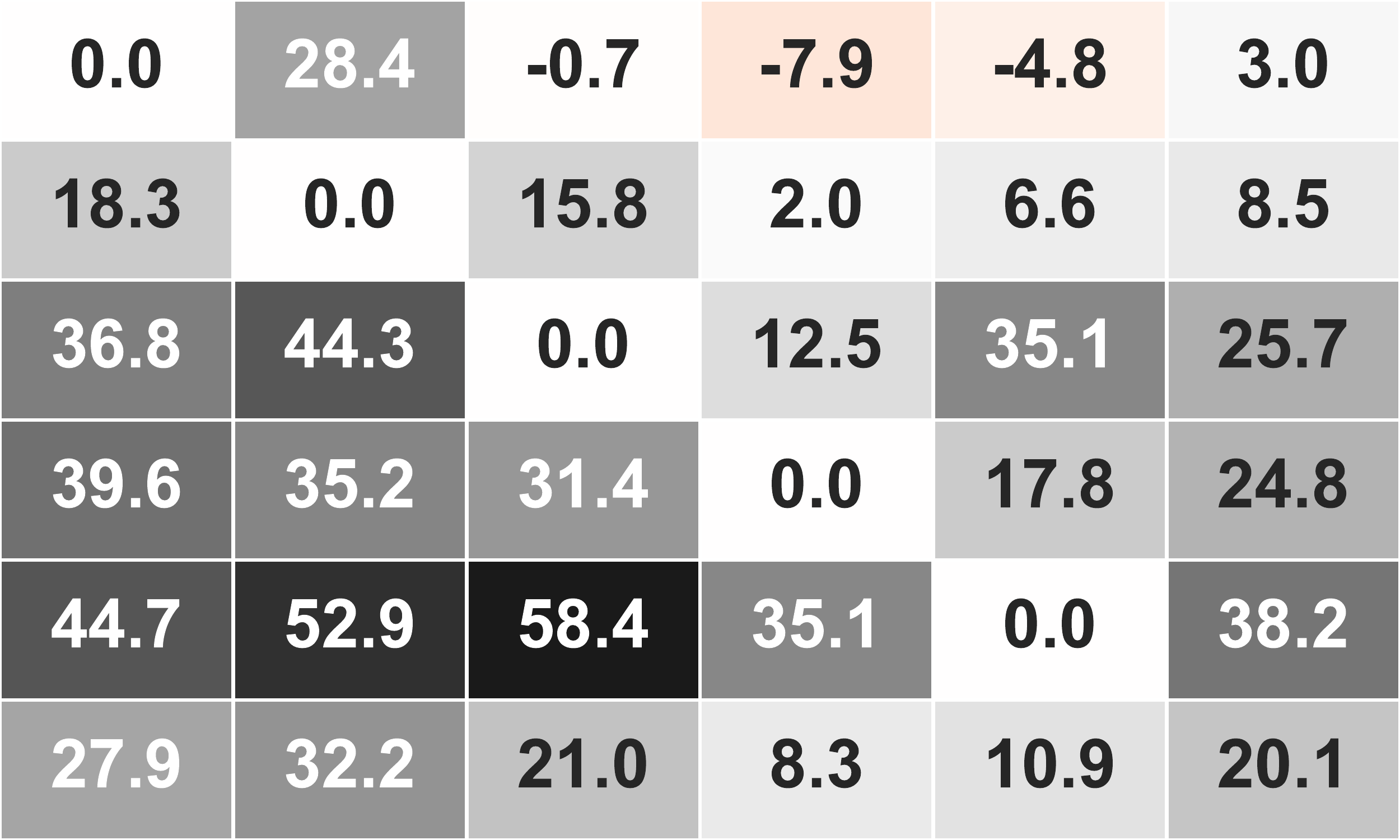} \label{fig:rouge normalized BiLSTM_SequenceLabelling_vs_LSTM} }}}  &       & \multicolumn{6}{l}{\multirow{7}[2]{*}[8pt]{\subfloat[ ]{         \includegraphics[height=200pt,width=420pt]{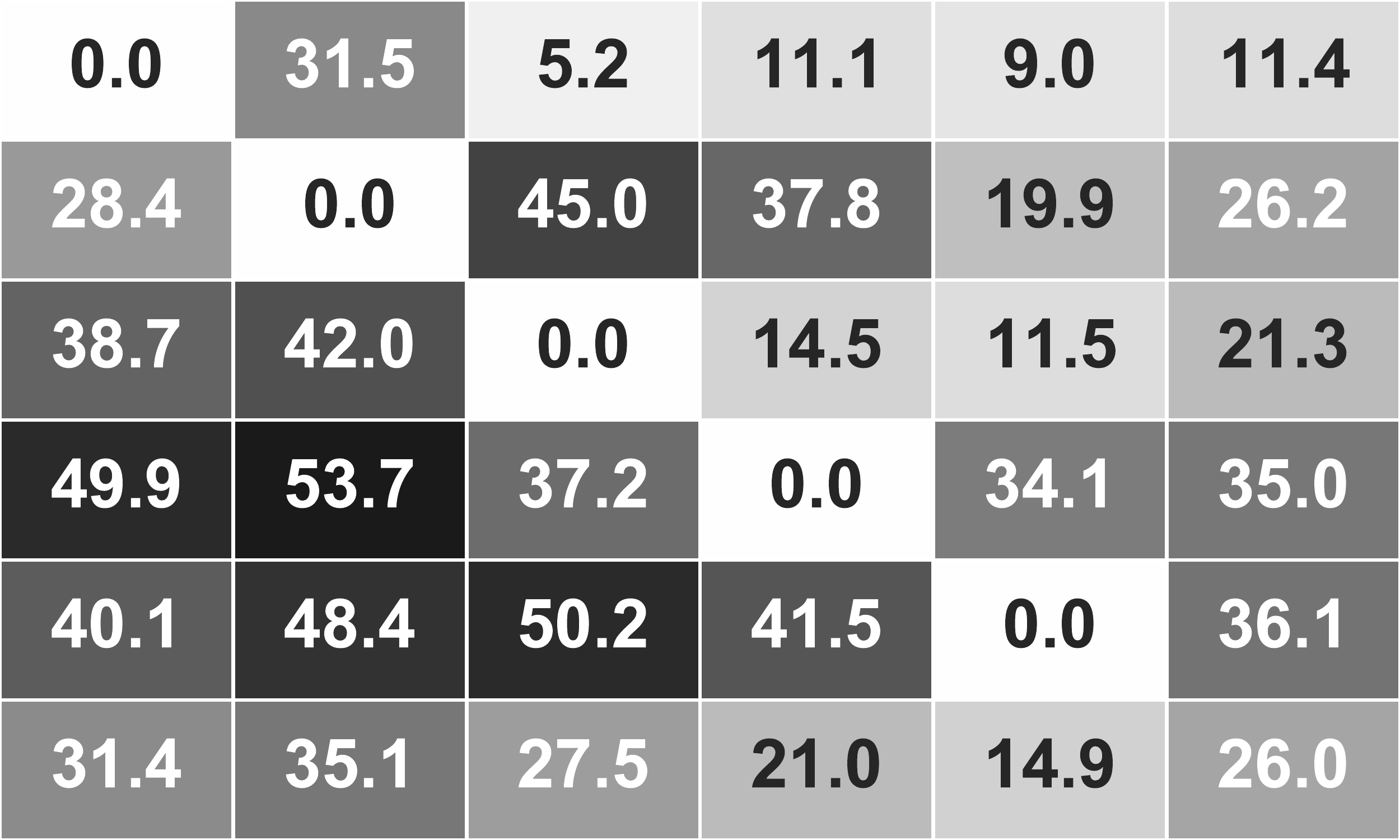} \label{fig:rouge normalized ext_bert_vs_abs_bert} }}} \\
          &       & Xsum  &       & \multicolumn{6}{l}{}                          &       & \multicolumn{6}{l}{}                          &       & \multicolumn{6}{l}{}                          &       & \multicolumn{6}{l}{}                          &       & \multicolumn{6}{l}{}                          &       & \multicolumn{6}{l}{} \\
          &       & Pubm. &       & \multicolumn{6}{l}{}                          &       & \multicolumn{6}{l}{}                          &       & \multicolumn{6}{l}{}                          &       & \multicolumn{6}{l}{}                          &       & \multicolumn{6}{l}{}                          &       & \multicolumn{6}{l}{} \\
          &       & Patent b &       & \multicolumn{6}{l}{}                          &       & \multicolumn{6}{l}{}                          &       & \multicolumn{6}{l}{}                          &       & \multicolumn{6}{l}{}                          &       & \multicolumn{6}{l}{}                          &       & \multicolumn{6}{l}{} \\
          &       & Red.  &       & \multicolumn{6}{l}{}                          &       & \multicolumn{6}{l}{}                          &       & \multicolumn{6}{l}{}                          &       & \multicolumn{6}{l}{}                          &       & \multicolumn{6}{l}{}                          &       & \multicolumn{6}{l}{} \\
          &       & \textcolor[rgb]{ 1,  0,  0}{avg} &       & \multicolumn{6}{l}{}                          &       & \multicolumn{6}{l}{}                          &       & \multicolumn{6}{l}{}                          &       & \multicolumn{6}{l}{}                          &       & \multicolumn{6}{l}{}                          &       & \multicolumn{6}{l}{} \\
          &       &       &       & \multicolumn{6}{l}{}                          &       & \multicolumn{6}{l}{}                          &       & \multicolumn{6}{l}{}                          &       & \multicolumn{6}{l}{}                          &       & \multicolumn{6}{l}{}                          &       & \multicolumn{6}{l}{} \\
    \bottomrule
    \end{tabular}}%
    \caption{The difference of ROUGE-1 F1 scores between different model pairs. Every column of the table represents the compared results of one pair of models. The line of holistic analysis displays the overall stiffness and stableness of compared models.
    The rest of the table is fine-grained results, the first line of which is the origin compared results ($\mathbf{U_A} - \mathbf{U_B}$ for model pairs $A$ and $B$) and the second line is the normalized compared results ($\mathbf{\hat{U}_A} - \mathbf{\hat{U}_B}$ for model pairs $A$ and $B$). For all heatmap, `grey' and `red' represent positive and negative respectively.
    Here we only display compared results for limited pairs of models, all other results are displayed in appendix.}
  \label{tab:pair compare}%
\end{table*}%

\begin{figure}[ht]
  \centering
  \vspace{-5pt}
  \setlength{\belowcaptionskip}{-0.3cm}
  \subfloat[stiffness (r$^{\mu}$)]{
    \label{fig:R1 stiff}
    \includegraphics[width=0.75\linewidth]{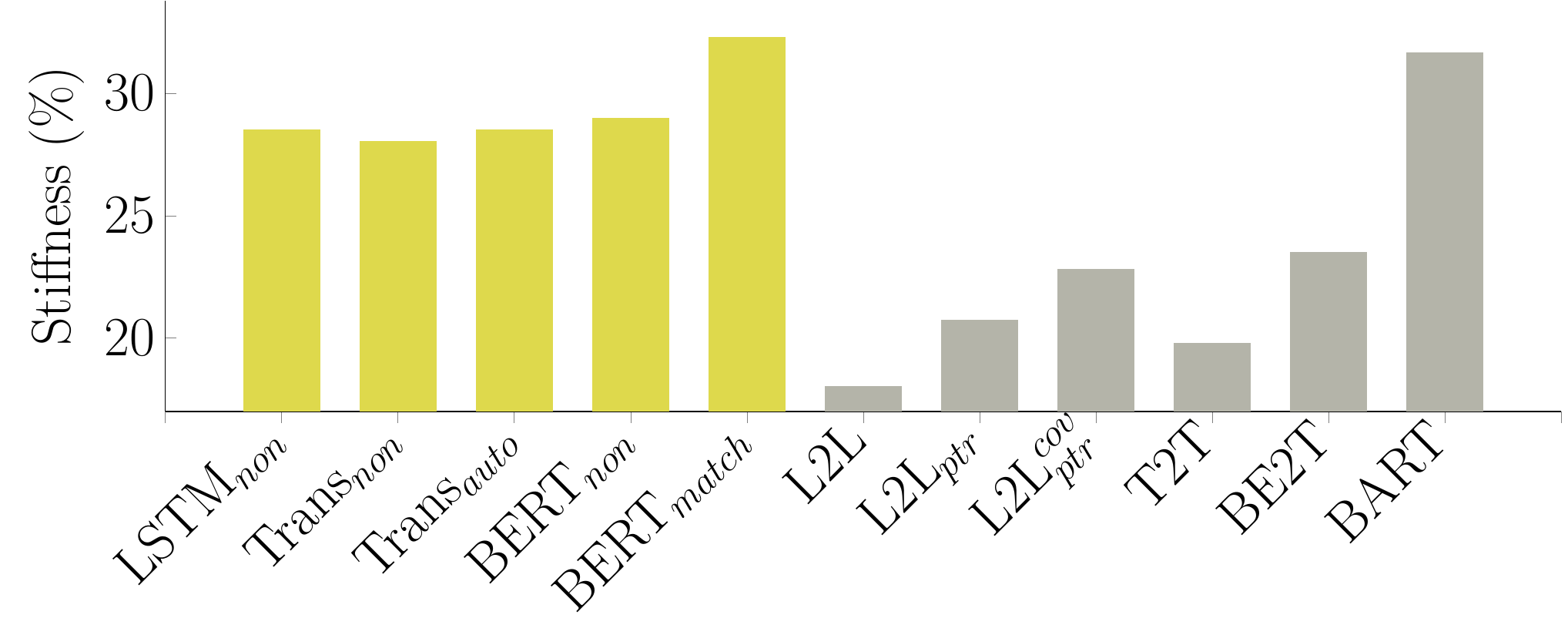}
  } 
  
  \subfloat[stableness (r$^{\sigma}$)]{
    \label{fig:R1 stable}
    \includegraphics[width=0.75\linewidth]{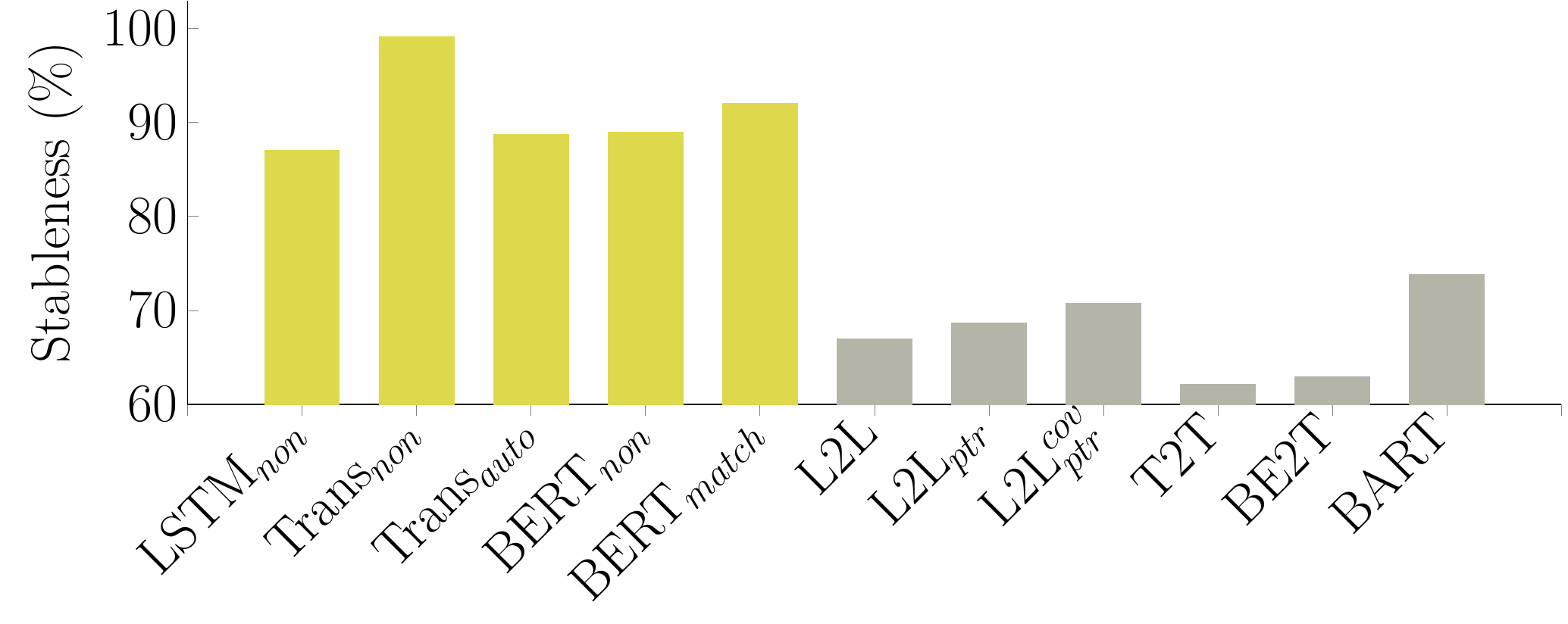}
  }
 \caption{Illustration of stiffness and stableness of ROUGE-1 F1 scores for  various models. Yellow bars stand for extractive models and grey bars stand for abstractive models.}
 \label{fig:all R1}
\end{figure}

\section{Experiment}
In what follows, we analyze different summarization systems in terms of semantic equivalence and factuality. Moreover, the results are studied in holistic and fine-grained views based on the measures defined above. Holistic results are showed in Fig.~\ref{fig:all R1} and Fig.~\ref{fig:all factcc}. On the other hand, Tab.~\ref{tab:pair compare} and Tab.~\ref{tab:fine grain factcc} display the fine-grained observations. Tab.~\ref{tab:in domain} dispalys the in-dataset results of all models on five benchmark datasets.

\subsection{Semantic Equivalence Analysis}
We conduct pair-wise Wilcoxon Signed-Rank significant test with $\alpha =$ 0.05. The null hypothesis is that the  expected performances (stiffness and stableness) of a pair of summarization models are identical. We report the observations that are statistically significant.
\subsubsection{Architecture}
\label{R1 architecture}

\paragraph{Match based reranking improves stiffness significantly}
\textit{BERT$_{match}$}, which using semantic match scores to rerank candidate summaries enhances the stiffness of model significantly in Fig.~\ref{fig:R1 stiff} while obtaining comparable stableness with other extractive models in Fig.~\ref{fig:R1 stable}. This indicates that \textit{BERT$_{match}$} not only increases the absolute performance but also retaining robustness.

\paragraph{\textit{BERT$_{match}$} is not stable when transferred from other datasets to \texttt{Bigpatent B}}
As Tab.~\ref{fig:rouge normalized matchsum_vs_ext_bert} shows, when compared to \textit{BERT$_{non}$}, \textit{BERT$_{match}$} obtains larger in-dataset and cross-dataset performance gap when tested in \texttt{Bigpatent B}. This is because \texttt{Bigpatent B} possesses higher sentence fusion score and higher repetition compared with other datasets as Sec.~\ref{databias analysis} demonstrates. When served as test set, such dataset brings great challenge for \textit{BERT$_{match}$} to correctly rank the candidate summaries while it provides more training signals when served as training set. Thus the in-dataset (\texttt{Bigpatent b}) trained model obtain much higher score compared with cross-dataset models which trained from other datasets and cause lower stableness.

\paragraph{Non-autoregressive decoder is more robust than autoregressive for extractive models.}
Regarding the decoder of extractive systems, as shown in Fig.~\ref{fig:R1 stiff} and Fig.~\ref{fig:R1 stable},
 the non-autoregressive extractive decoder (\textit{Trans$_{non}$}) is more stable while it possesses lower stiffness than its autoregressive counterpart (\textit{Trans$_{auto}$}).

\paragraph{Pointer network and coverage mechanism are instrumental in improving stiffness and stableness of abstractive systems.}
The pointer network and coverage mechanism do enhance the absolute performance of abstractive system as Fig.~\ref{fig:R1 stiff} demonstrates (r$^{\mu}$(\textit{L2L$_{ptr}^{cov}$})\ $ > $\ r$^{\mu}$(\textit{L2L$_{ptr}$})\ $ > $\ r$^{\mu}$(\textit{L2L})).
Also, the stableness results of \textit{L2L$_{ptr}$} and \textit{L2L} in Fig.~\ref{fig:R1 stable} reveals that once removing the pointer mechanism, the value of $r^{\sigma}$ for \textit{L2L$_{ptr}$} decreases, which suggests that \textit{the system will be more stable if it's augmented the ability to directly extract text spans from the the source document}.

\paragraph{However, pointer network brings trivial improvement when tested in \texttt{Xsum} and \texttt{Reddit}}
The absolute model performance improvement of pointer network is trival when tested in \texttt{xsum} and \texttt{Reddit} as showed in Tab.~\ref{fig:rouge origin LSTM_pointer_gen_vs_LSTM}, which is in line with expectations because these two datasets are more abstractive as analyzed in Sec.~\ref{databias analysis}.

\paragraph{On the other hand, coverage is not that helpful when tested in \texttt{Reddit} and \texttt{Xsum} and even harmful when trained in \texttt{Xsum}.}
The heatmap of \textit{L2L$_{ptr}^{cov}$} vs. \textit{L2L$_{ptr}$} in Tab.\ref{fig:rouge origin LSTM_pointer_gen_coverage_vs_LSTM_pointer_gen}) shows that when tested in \texttt{Reddit} and \texttt{Xsum}, the improvement of coverage mechanism is trivial. These two datasets possess less repetition, thus coverage can not provide much help when transferred to these datasets.
Moreover, when trained in \texttt{Xsum}, \textit{L2L$_{ptr}^{cov}$} gets lower stiffness compared with \textit{L2L$_{ptr}$}, which is in accordance with the normalized result in Tab.~\ref{fig:rouge normalized LSTM_pointer_gen_coverage_vs_LSTM_pointer_gen}. This is because the gold summaries of \texttt{Xsum} exhibit lower repetition score (as analyzed in Sec.~\ref{databias analysis}), thus can't provide enough learning signals for coverage mechanism.

\paragraph{BERT sometimes brings unstableness.}
As shown in Fig.~\ref{fig:R1 stiff}, there is no doubt that once summarizers (extractive or abstractive) are equipped with pre-trained encoder, the stiffness will increase significantly (e.g., $r^{\mu}$(\textit{BE2T} $>>$ $r^{\mu}$(\textit{T2T}), suggesting that the overall cross-dataset performance has been improved.
\textit{However, we are surprised to find (from Fig.~\ref{fig:R1 stable}) that BERT sometimes leads to unstableness} (i.e. $r^{\sigma}$(\textit{Trans$_{non}$}) $>$ $r^{\sigma}$(\textit{BERT$_{non}$})). 
This result enlightens us to search for other architectures or learning schemas to offset the unstableness brought by BERT.

As the heatmap of \textit{BERT$_{non}$} vs. \textit{Trans$_{non}$} in Tab.~\ref{fig:rouge normalized ext_BERT_vs_baseline_ext_BERT} shows, BERT brings unstableness especially when tested in \texttt{Reddit} and \texttt{Xsum}.

\paragraph{BERT sometimes can even harm the absolute cross-dataset performance.}
\textit{BERT$_{non}$} performs worse than \textit{Trans$_{non}$} in some cells (e.g., trained in \texttt{Xsum} and tested in \texttt{CNNDM}) in Tab.~\ref{fig:rouge origin ext_BERT_vs_baseline_ext_BERT}

\paragraph{BART shows superior performance in terms of stiffness and stableness.}
As Fig.~\ref{fig:R1 stiff} shows, \textit{BART} obtains the highest stiffness among all abstractive models, and is even comparable with 
\textit{BERT$_{match}$}.
In addition, \textit{BART} is also outstanding in terms of stableness when compared with other abstractive models (Fig.~\ref{fig:R1 stable}). The performance gap between \textit{BART} and \textit{BE2T} proves that for abstractive models, pre-training the whole sequence to sequence model works better than using the pre-trained model in either side of encoder or decoder.

\subsubsection{Generation ways}
\label{R1 generation way}

\paragraph{Extractive models are superior to abstractive models in terms of stiffness and robustness.}
Extractive models show superior advantage of absolute performance as shown in Fig.~\ref{fig:R1 stiff}.
Moreover, comparing the stableness of abstractive and extractive models in Fig.~\ref{fig:R1 stable},
we surprisingly find that \textit{abstractive approaches except for \textit{BART} are extremely brittle} since their $r^{\sigma}$ value is much lower than any extractive approaches with a maximum margin of 37$\%$, and the gap can be reduced by introducing pointer network. 
This observation poses a great challenge to the development of the abstractive systems, encouraging research to pay more attention to improve the generalization ability.
Also, we have provided hints for the solution, such as enabling the model to extract granular information from the source document or using the well pre-trained sequence to sequence model (e.g., \textit{BART}).

\paragraph{When tested in \texttt{Xsum} and \texttt{Reddit}, abstractive systems possess comparable or even better performance.}

The supremacy of extractive models is not retained in all datasets (Tab.~\ref{fig:rouge origin ext_bert_vs_abs_bert} and Tab.~\ref{fig:rouge origin BiLSTM_SequenceLabelling_vs_LSTM})
Though extractive models obtain higher stiffness scores when tested in \texttt{CNNDM} and \texttt{PubMed},  abstractive approaches (\textit{BE2T}, \textit{L2L}) obtained higher or comparable stiffness scores when tested at \texttt{XSUM} and \texttt{Reddit}. 
This is because \texttt{Xsum} and \texttt{Reddit} are more abstractive as analyzed in Sec.~\ref{databias analysis}.

\begin{figure}[t]
  \centering
  \vspace{-10pt}
  \setlength{\belowcaptionskip}{-0.3cm}
  \subfloat[stiffness (r$^{\mu}$)]{
    \label{fig:factcc stiff}
    \includegraphics[width=0.75\linewidth]{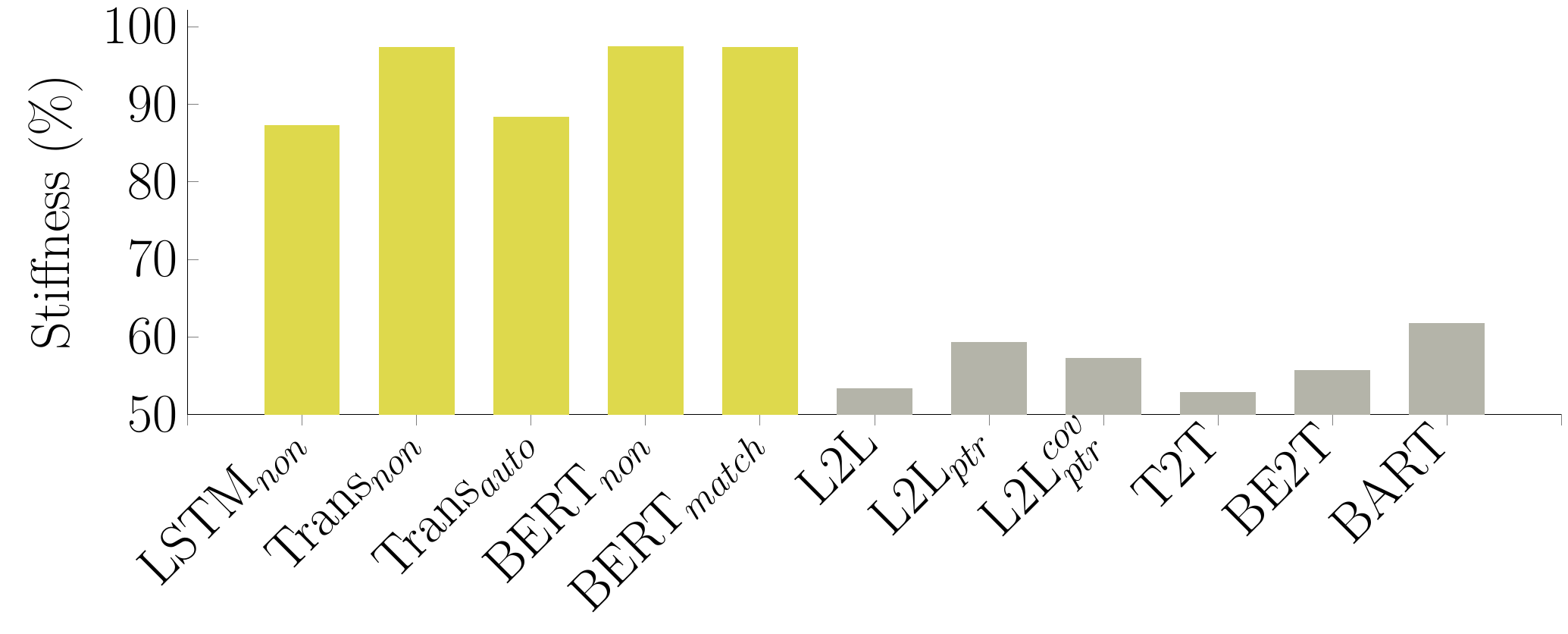}
  } 
  
  \subfloat[stableness (r$^{\sigma}$)]{
    \label{fig:factcc stable}
    \includegraphics[width=0.75\linewidth]{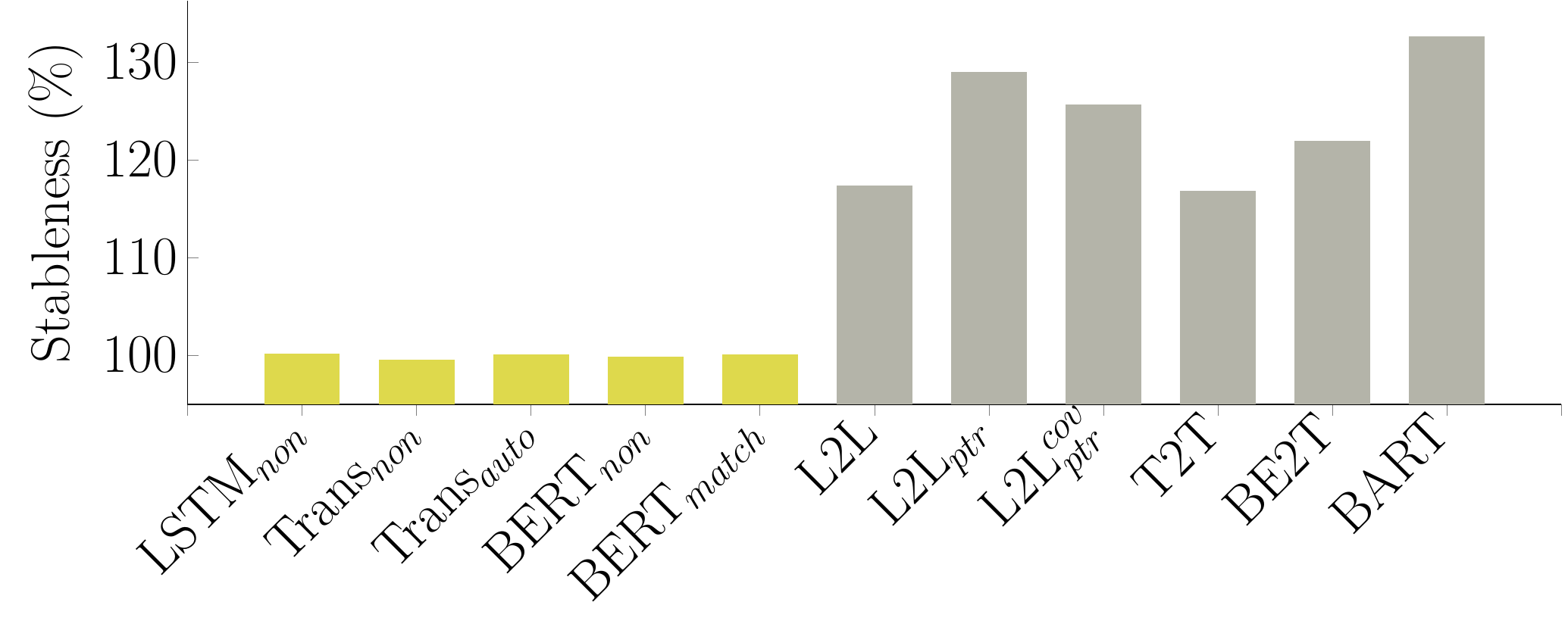}
  }
 \caption{Illustration of stiffness and stableness of factuality scores for various models. Yellow bars stand for extractive systems and grey bars stand for abstractive systems.}
 \label{fig:all factcc}
\end{figure}

\subsection{Factuality Analysis}
\label{factcc analysis}

1) All extractive models can achieve higher factuality scores while all abstractive models obtain quite lower ones (Fig.~\ref{fig:factcc stiff}).
One interesting observation is,  for extractive models, not all factuality scores under the in-dataset setting are $100\%$ in Tab.~\ref{tab:fine grain factcc} (on-diagonal values), which reveals the limitation of existing factuality checker.

\noindent 2) \textit{BART} can significantly improve the ability to generate factual summaries compared with other abstractive models as showed in Fig.~\ref{fig:factcc stiff}, even compared with \textit{L2L$_{ptr}$} which equipped with pointer network and tend to copy from source document.

\noindent 3) Abstractive models obtain higher stableness of factuality scores in Fig.~\ref{fig:factcc stable} which surpass 100\%. 
This is because when tested in abstractive datasets (e.g., \texttt{Xsum} as Sec. \ref{databias analysis} shows), abstractive summarizers trained in-dataset tend to be more abstractive and obtain lower factuality score while it gets higher factuality score when trained on other datasets which are more extractive (e.g., \texttt{CNNDM}). The superiority of cross-dataset results over in-dataset results thus leads to higher stableness.

\begin{table}[tbp]
  \centering
    \setlength{\tabcolsep}{2.6pt}
      \large
      \renewcommand{\arraystretch}{1.1}
        \resizebox{0.48\textwidth}{23mm}{
    \begin{tabular}{ccccccccccccc}
    \toprule
    \multirow{2}[4]{*}{\textbf{EXT models}} & \multicolumn{5}{c}{Trans$_{non}$}          &       & \multicolumn{5}{c}{BERT$_{match}$}         &  \\
\cmidrule{2-6}\cmidrule{8-12}          & CNN.  & XSUM  & Pubm. & Patent B & Red.  & \textcolor[rgb]{ .753,  .314,  .302}{avg} & CNN.  & XSUM  & Pubm. & Patent B & Red.  & \textcolor[rgb]{ .753,  .314,  .302}{avg} \\
    \midrule
    CNN   & 100.0  & 100.0  & 98.0  & 99.1  & 100.0  & \textcolor[rgb]{ .753,  .314,  .302}{99.4 } & 99.8  & 99.4  & 92.9  & 95.7  & 99.1  & \textcolor[rgb]{ .753,  .314,  .302}{97.4 } \\
    XSUM  & 99.8  & 100.0  & 97.4  & 98.2  & 100.0  & \textcolor[rgb]{ .753,  .314,  .302}{99.1 } & 99.7  & 99.5  & 93.2  & 95.1  & 98.8  & \textcolor[rgb]{ .753,  .314,  .302}{97.3 } \\
    Pubm. & 97.7  & 98.8  & 95.1  & 94.7  & 100.0  & \textcolor[rgb]{ .753,  .314,  .302}{97.3 } & 99.7  & 99.2  & 93.1  & 95.2  & 99.3  & \textcolor[rgb]{ .753,  .314,  .302}{97.3 } \\
    Patent B & 98.3  & 99.8  & 96.3  & 97.4  & 99.5  & \textcolor[rgb]{ .753,  .314,  .302}{98.3 } & 99.7  & 99.0  & 93.0  & 94.5  & 98.4  & \textcolor[rgb]{ .753,  .314,  .302}{96.9 } \\
    Reddit & 90.3  & 94.1  & 94.1  & 86.7  & 96.3  & \textcolor[rgb]{ .753,  .314,  .302}{92.3 } & 99.7  & 99.3  & 93.1  & 96.1  & 99.3  & \textcolor[rgb]{ .753,  .314,  .302}{97.5 } \\
    \textcolor[rgb]{ .753,  .314,  .302}{avg} & \textcolor[rgb]{ .753,  .314,  .302}{97.2 } & \textcolor[rgb]{ .753,  .314,  .302}{98.6 } & \textcolor[rgb]{ .753,  .314,  .302}{96.2 } & \textcolor[rgb]{ .753,  .314,  .302}{95.2 } & \textcolor[rgb]{ .753,  .314,  .302}{99.2 } & \textcolor[rgb]{ .753,  .314,  .302}{97.3 } & \textcolor[rgb]{ .753,  .314,  .302}{99.7 } & \textcolor[rgb]{ .753,  .314,  .302}{99.3 } & \textcolor[rgb]{ .753,  .314,  .302}{93.0 } & \textcolor[rgb]{ .753,  .314,  .302}{95.3 } & \textcolor[rgb]{ .753,  .314,  .302}{99.0 } & \textcolor[rgb]{ .753,  .314,  .302}{97.3 } \\
    \midrule
    \multirow{2}[4]{*}{\textbf{ABS models}} & \multicolumn{5}{c}{T2T}               &       & \multicolumn{5}{c}{BART}              &  \\
\cmidrule{2-6}\cmidrule{8-12}          & CNN.  & XSUM  & Pubm. & Patent B & Red.  & \textcolor[rgb]{ .753,  .314,  .302}{avg} & CNN.  & XSUM  & Pubm. & Patent B & Red.  & \textcolor[rgb]{ .753,  .314,  .302}{avg} \\
    \midrule
    CNN   & 72.4  & 75.7  & 71.5  & 71.8  & 70.5  & \textcolor[rgb]{ .753,  .314,  .302}{72.4 } & 69.9  & 77.9  & 87.4  & 84.1  & 90.2  & \textcolor[rgb]{ .753,  .314,  .302}{81.9 } \\
    XSUM  & 9.7   & 22.6  & 10.8  & 9.9   & 19.1  & \textcolor[rgb]{ .753,  .314,  .302}{14.4 } & 35.5  & 24.7  & 36.1  & 50.1  & 50.7  & \textcolor[rgb]{ .753,  .314,  .302}{39.4 } \\
    Pubm. & 58.5  & 59.3  & 56.2  & 72.3  & 34.9  & \textcolor[rgb]{ .753,  .314,  .302}{56.2 } & 69.5  & 61.5  & 58.4  & 61.3  & 94.1  & \textcolor[rgb]{ .753,  .314,  .302}{69.0 } \\
    Patent B & 79.2  & 81.2  & 84.4  & 68.7  & 73.9  & \textcolor[rgb]{ .753,  .314,  .302}{77.5 } & 52.1  & 53.8  & 69.0  & 67.4  & 76.8  & \textcolor[rgb]{ .753,  .314,  .302}{63.8 } \\
    Reddit & 34.8  & 35.7  & 50.6  & 44.6  & 52.5  & \textcolor[rgb]{ .753,  .314,  .302}{43.6 } & 59.6  & 50.3  & 69.1  & 49.3  & 44.2  & \textcolor[rgb]{ .753,  .314,  .302}{54.5 } \\
    \textcolor[rgb]{ .753,  .314,  .302}{avg} & \textcolor[rgb]{ .753,  .314,  .302}{50.9 } & \textcolor[rgb]{ .753,  .314,  .302}{54.9 } & \textcolor[rgb]{ .753,  .314,  .302}{54.7 } & \textcolor[rgb]{ .753,  .314,  .302}{53.5 } & \textcolor[rgb]{ .753,  .314,  .302}{50.2 } & \textcolor[rgb]{ .753,  .314,  .302}{52.8 } & \textcolor[rgb]{ .753,  .314,  .302}{57.3 } & \textcolor[rgb]{ .753,  .314,  .302}{53.6 } & \textcolor[rgb]{ .753,  .314,  .302}{64.0 } & \textcolor[rgb]{ .753,  .314,  .302}{62.4 } & \textcolor[rgb]{ .753,  .314,  .302}{71.2 } & \textcolor[rgb]{ .753,  .314,  .302}{61.7 } \\
    \bottomrule
    \end{tabular}}%
    \caption{Cross-dataset factuality scores for extractive and abstractive models.}
  \label{tab:fine grain factcc}%
\end{table}%

\section{Related Work}
Our work is connected to the following threads of topics of NLP research.

\paragraph{Cross-Dataset Generalization in NLP}
Recently, more researchers shift their focus from individual dataset to cross-dataset evaluation, aiming to get a comprehensive understanding of system's generalization ability. 
\citet{fried-etal-2019-cross} explores the generalization ability of different constituency parsers. 
\citet{talmor-berant-2019-multiqa}, on the other hand, shows the generalization ability of reading comprehension models can be improved by pre-training on one or two other reading comprehension datasets.
\citet{fu2020rethinking} studies the model generalization in the field of NER. They
point out the bottleneck of the existing NER systems through in-depth analyses and provide suggestions for further improvement.
Different from the above works, we attempt to explore generalization ability for summarization systems.

\paragraph{Diagnosing Limitations of Existing Summarization Systems}

Beyond ROUGE, some recent works try to explore the weaknesses of existing systems from divese aspects.
\citet{zhang-etal-2018-abstractiveness} tries to figure out to what extent the neural abstractive summarization systems are abstractive and discovers many of abstractive systems tend to perform near-extractive.
On the other hand, \citet{cao2018faithful} and \citet{kryscinski2019evaluating} study the factuality problem in modern neural summarization systems. The former puts forward one model that combining source document and preliminary extracted fact description and prove the effectiveness of this model in terms of factuality correctness. While the latter contributes to design a model-based automatic factuality evaluation metric. Abstractiveness and factuality error the above works studied are 
orthogonal to this work and can be easily combined with cross-dataset evaluation framework in this paper as Sec.~\ref{factcc analysis} shows. 
Moreover, \citet{wang2019exploring,hua2017pilot} attempt to investigate the domain shift problem on text summarization while they focus on a single generation way (either abstractive or extractive).
We also investigate the generalization of summarizers when transferring to different datasets, but include more datasets and models.

\section{Conclusion}
By performing a comprehensive evaluation on eleven summarization systems and five mainstream datasets, we summarize our observations below:

1) Abstractive summarizers are extremely brittle compared with extractive approaches, and the maximum gap between them reaches 37$\%$ in terms of the measure \textit{stableness} (ROUGE) defined in this paper. 
2) \textit{BART} (SOTA system) is superior over other abstractive models and even comparable with extractive models in terms of stiffness (ROUGE). On the other hand, it is robust when transferring between datasets as it possesses high stableness (ROUGE).
3) \textit{BERT$_{match}$} (SOTA system) performs excellently in terms of stiffness, while still lacks stableness when transferred to \texttt{Bigpatent B} from other datasets.  
4) The robustness of models can be improved through either equipped the model with ability to copy span from source document (i.e., ~\citet{lebanoff2019scoring}) or make use of well trained sequence to sequence pre-trained model (\textit{BART}).
5) Simply adding BERT on encoder could improve the stiffness (ROUGE) of model but will cause larger cross-dataset and in-dataset performance gap, a better way should be found to merge BERT into abstractive model, or a better training strategy should be applied to offset the negative influence it brings.
6) Existing factuality checker (Factcc) is limited in predictive power of positive samples  (Sec.\ref{factcc analysis}).
7) Out-of-domain systems can even surpass in-domain systems in terms of factuality. (Sec.\ref{factcc analysis})

\section*{Acknowledgements}
We would like to thank the anonymous
reviewers for their detailed comments and constructive suggestions. This work was supported by the National Natural Science Foundation of China (No. 62022027 and 61976056), Science and Technology on Parallel and Distributed Processing Laboratory (PDL).

\bibliography{nlp,nlp1}
\bibliographystyle{acl_natbib}

\appendix

\section{Appendices}
\label{sec:appendix}

\subsection{Detailed Dataset introduction}
\paragraph{\texttt{CNN/DailyMail}}
The CNN/DailyMail question answering dataset \citep{hermann2015teaching} modified by \citet{nallapati2016abstractive} is commonly used for summarization. The dataset consists of online news articles with paired human-generated summaries.
For the data preprocessing, we use the non-anonymized data as \citet{see2017get}, which doesn't replace named entities.

\paragraph{\texttt{XSUM}}
XSUM \citep{narayan2018don} is a dataset consists of the articles and the single-sentence answers of the question ``What is the article about?" as summary. It is more abstractive compared with CNN/DailyMail.

\paragraph{\textsc{Pubmed}}
\textsc{Pubmed} \citep{cohan2018discourse} is drawn from scientific papers specifically medical journal articles from the PubMed Open Access Subset. We use the introduction as source document and the abstract as summary here.

\paragraph{\textsc{Bigpatent}}
\textsc{Bigpatent} \citep{sharma2019bigpatent} consists of 1.3 million records of U.S. patent documents and the corresponding summaries are created by human. According to Cooperative Patent Classification (CPC), the dataset is divided to nine categories. One of the nine categories is chosen as a dataset in difference domain in our experiment (Category B: Performing Operations; Transporting).

\paragraph{\textsc{Reddit TIFU}}
\textsc{Reddit TIFU} \citep{kim2019abstractive} is a dataset with less formal posts compared with datasets mentioned above which mostly use formal documents as source. It is collected from the online discussion forum Reddit.
They regard the body text as source, the title as short summary, and the TL;DR summary as long summary, thus making two sets of datasets: TIFU-short and TIFU-long. TIFU-long is used in this paper.

\subsection{Dataset statistics}
The detailed dataset statistics are presented in Tab. \ref{tab:datasetstat}
\setlength{\tabcolsep}{3pt}
\renewcommand{\arraystretch}{1.2}
\begin{table}[htbp]
  \centering
  \setlength{\belowcaptionskip}{-5pt}
  \small
    \begin{tabular}{lccccc}
    \toprule
    \textbf{Datasets} & \textbf{Statistics} &  \textbf{Topics} & \multicolumn{1}{l}{\textbf{Oracle}} & \multicolumn{1}{l}{\textbf{Lead-k}} \\
    \midrule
    CNNDM & 2,764/123/107M & News  & 55.21 & 40.32 \\
    Xsum  & 1126/60/59M & News  & 30.41 & 16.38 \\
    Pubmed & 644/36/38M & Scientific  & 46.21 & 37.52 \\
    BigPatent B & 4,812/265/262M & Patents  & 51.53 & 31.85 \\
    Reddit & 206/3.3/3.6M & Posts  & 36.47 & 11.09 \\
    \bottomrule
    \end{tabular}%
    \caption{Detailed statistics of five datasets.  Lead-$k$ indicates ROUGE-1 F1 score of the first $k$ sentences in the document and Oracle indicates the globally optimal combination of sentences in terms of ROUGE-1 F1 scores with ground truth, the latter represents the upper bound of extractive models.}
  \label{tab:datasetstat}%
\end{table}%

\subsection{Experimental setup}
\subsubsection{Extractive Summarizers}
We use the same training setup in \citep{zhong2019searching}. We use cross entropy as loss function to train \textit{LSTM$_{non}$} and \textit{Trans$_{auto}$}. The hidden state dimension of LSTM in \textit{LSTM$_{non}$} is set to 512 and the hidden state dimension of Transformer in \textit{Trans$_{auto}$} is 2048. We use Transformer with 8 heads.

\textit{BERT$_{non}$} and \textit{Trans$_{non}$} is constructed according to \citet{liu2019text}. 
All documents and summaries are truncated to 512 tokens when training. \textit{BERT$_{non}$} and \textit{Trans$_{non}$} are trained for 50000 steps, the gradient is accumulated every two steps. We use Adam as optimizer and the learning rate is set to 2e-3.

\textit{BERT$_{match}$} is trained as in  \citet{zhong2020extractive}. It uses the base version of BERT as base model. We use Adam optimizer with warming up. The learning rate schedule follows \citet{vaswani2017attention}.

\subsubsection{Abstractive Summarizers}
\textit{L2L}, \textit{L2L$_{ptr}$} and \textit{L2L$_{ptr}^{cov}$} are trained using the pytorch reproduced version code of \citet{see2017get}. We use the same size of vocabulary(50k), hidden state dimension (256) and word embedding dimension (128) as in the paper. All of three models are trained with 650000 maximum training steps, We use Adagrad to train the models with learning rate of 0.15.

\textit{BE2T} and \textit{T2T} is constructed according to \citet{liu2019text}.
We use two separate optimizers for the decoder and encoder regarding \textit{BE2T} to offset the mismatch of encoder and decoder, since the former is pre-trained while the latter is not. Learning rates for the optimizers of encoder and decoder are 0.002 and 0.2 respectively.
On the other hand, \textit{BE2T} and \textit{T2T} are trained with gradient accumulation every five steps, training step for which is 200000. 

\textit{BART} uses the large pre-trained sequence to sequence model in  \citet{lewis2019BART}. The total learning step when fine-tuning is set to 20000 with 500 steps warming up. We use Adam as optimizer and learning rate is 3e-05.

\begin{table*}[htbp]
  \setlength{\tabcolsep}{1.9pt}
\renewcommand{\arraystretch}{1.1}
  \centering
  \scriptsize
    \begin{tabular}{clcccrcccrcccrcccrccc}
    \toprule
    \multicolumn{2}{c}{\multirow{2}[4]{*}{\textbf{Models}}} & \multicolumn{3}{c}{\textbf{CNNDM}} &       & \multicolumn{3}{c}{\textbf{XSUM}} &       & \multicolumn{3}{c}{\textbf{PubMed}} &       & \multicolumn{3}{c}{\textbf{Bigpatent b}} &       & \multicolumn{3}{c}{\textbf{Reddit}} \\
\cmidrule{3-5}\cmidrule{7-9}\cmidrule{11-13}\cmidrule{15-17}\cmidrule{19-21}    \multicolumn{2}{c}{} & \textbf{R1} & \textbf{R2} & \textbf{RL} &       & \textbf{R1} & \textbf{R2} & \textbf{RL} &       & \textbf{R1} & \textbf{R2} & \textbf{RL} &       & \textbf{R1} & \textbf{R2} & \textbf{RL} &       & \textbf{R1} & \textbf{R2} & \textbf{RL} \\
\cmidrule{1-5}\cmidrule{7-9}\cmidrule{11-13}\cmidrule{15-17}\cmidrule{19-21}    \multirow{5}[2]{*}{Ext.} & LSTM$_{non}$ & 41.36  & 18.81  & 37.73  &       & 19.51  & 3.10  & 14.50  &       & 42.98  & 16.59  & 38.28  &       & 39.29  & 13.07  & 32.61  &       & 20.46  & 5.05  & 16.33  \\
          & Trans$_{non}$ & 40.84  & 18.23  & 37.09  &       & 15.74  & 1.67  & 11.58  &       & 38.45  & 13.28  & 34.16  &       & 34.41  & 10.05  & 28.75  &       & 16.25  & 2.60  & 12.57  \\
          & Trans$_{auto}$ & 41.35  & 18.77  & 37.75  &       & 19.29  & 2.80  & 14.21  &       & 42.74  & 16.34  & 38.05  &       & 38.76  & 12.60  & 32.17  &       & 18.55  & 3.44  & 14.62  \\
          & BERT$_{non}$ & 42.69  & 19.88  & 38.99  &       & 21.76  & 4.24  & 16.00  &       & 38.74  & 13.62  & 34.48  &       & 35.85  & 11.05  & 29.97  &       & 21.84  & 5.21  & 17.15  \\
          & BERT$_{match}$ & 44.26  & 20.58  & 40.40  &       & 24.97  & 4.76  & 18.48  &       & 41.19  & 14.91  & 36.73  &       & 38.89  & 12.82  & 32.48  &       & 25.32  & 6.16  & 20.17  \\
\cmidrule{1-5}\cmidrule{7-9}\cmidrule{11-13}\cmidrule{15-17}\cmidrule{19-21}    \multirow{6}[2]{*}{Abs.} & L2L   & 32.80  & 12.84  & 30.34  &       & 28.31  & 8.71  & 22.30  &       & 27.84  & 7.45  & 25.69  &       & 30.46  & 9.76  & 27.61  &       & 16.89  & 1.24  & 13.63  \\
          & L2L$_{ptr}$ & 37.06  & 15.96  & 33.74  &       & 29.67  & 9.58  & 23.40  &       & 32.04  & 10.38  & 28.97  &       & 31.03  & 9.92  & 25.35  &       & 21.32  & 4.46  & 17.14  \\
          & L2L$_{ptr}^{cov}$ & 39.95  & 17.54  & 36.25  &       & 28.83  & 8.83  & 22.62  &       & 35.27  & 11.89  & 31.92  &       & 35.90  & 12.31  & 32.78  &       & 21.28  & 4.39  & 17.22  \\
          & T2T   & 39.90  & 17.66  & 37.08  &       & 29.01  & 9.13  & 22.77  &       & 30.71  & 8.10  & 27.97  &       & 42.94  & 16.75  & 37.06  &       & 19.96  & 3.36  & 15.60  \\
          & BE2T  & 41.34  & 18.98  & 38.41  &       & 38.99  & 16.64  & 31.23  &       & 37.11  & 13.38  & 33.72  &       & 43.10  & 17.11  & 37.34  &       & 26.66  & 7.00  & 21.21  \\
          & BART  & 44.75  & 21.69  & 41.46  &       & 44.73  & 21.99  & 37.02  &       & 45.02  & 16.94  & 41.17  &       & 45.78  & 18.31  & 38.98  &       & 34.00  & 11.88  & 26.91  \\
    \bottomrule
    \end{tabular}%
    \caption{Representative summarizers we have studied in this paper and their correspond performance (ROUGE-1 F1, ROUGE-2 F1, ROUGE-L F1) on different datasets.}
  \label{tab:all in domain R1}%
\end{table*}%

\subsection{In-dataset ROUGE results for all models}
Tab.~\ref{tab:all in domain R1} displays in-dataset ROUGE-1 F1 ,ROUGE-2 F1 ,ROUGE-L F1 scores.

\subsection{The ROUGE-1 F1 score difference of all model pairs which are meaningful to compare}
The holistic and fine-grained results of pair-wise comparison are displayed in Tab.~\ref{tab:all fine grain R1}.

\subsection{Cross-dataset factuality results of all models}
The cross-dataset factcc results for abstractive models are shown in Tab.~\ref{tab:all factcc abs} and the factcc results of extractive models are demonstrated in Tab.~\ref{tab:all factcc ext}.

\subsection{Code urls}
\subsubsection{Training code urls}
The models and their training code urls are listed below: 

\textit{LSTM$_{non}$} and \textit{Trans$_{auto}$} are trained from the code in \citet{zhong2019searching}, the code url is \url{https://github.com/maszhongming/Effective_Extractive_Summarization}.

We use the code from \citet{liu2019text} for \textit{BERT$_{non}$}, \textit{Trans$_{non}$}, \textit{BE2T} and \textit{T2T}. Code url is \url{https://github.com/nlpyang/PreSumm}.

\textit{BERT$_{match}$} uses the code from \citet{zhong2020extractive} and the code url is \url{https://github.com/maszhongming/MatchSum}.

\textit{L2L}, \textit{L2L$_{ptr}$} and \textit{L2L$_{ptr}^{cov}$} are trained from the code of \citet{see2017get}, code url is \url{https://github.com/atulkum/pointer_summarizer}.

We use code in fairseq \citep{ott2019fairseq} to fine-tune \textit{BART}, the code url is \url{https://github.com/pytorch/fairseq/tree/master/examples/bart}.

\subsubsection{Evaluation code urls}
The evaluation metrics code urls are listed below:

We use pyrouge (\url{https://github.com/bheinzerling/pyrouge}) to evaluate the ROUGE performance of models.

The url for Factcc~\citep{kryscinski2019evaluating} is \url{https://github.com/salesforce/factCC}.

The url for other metrics for dataset bias is \url{https://github.com/zide05/CDEvalSumm/tree/master/Data-bias-metrics}.

\begin{table*}[htbp]
  \centering
    \setlength{\tabcolsep}{1.5pt}
      \scriptsize
      \renewcommand{\arraystretch}{1.3}
        \resizebox{1\textwidth}{16mm}{
    \begin{tabular}{ccccccccccccccccccccccccccccccccccccc}
    \toprule
    \multirow{2}[4]{*}{\textbf{ABS models}} & \multicolumn{5}{c}{L2L}               &       & \multicolumn{5}{c}{L2L$_{ptr}$}            &       & \multicolumn{5}{c}{L2L$_{ptr}^{cov}$}         &       & \multicolumn{5}{c}{T2T}               &       & \multicolumn{5}{c}{BE2T}              &       & \multicolumn{5}{c}{BART}              &  \\
\cmidrule{2-6}\cmidrule{8-12}\cmidrule{14-18}\cmidrule{20-24}\cmidrule{26-30}\cmidrule{32-36}          & \rotatebox{90}{CNN.}  & \rotatebox{90}{XSUM}  & \rotatebox{90}{Pubm.} & \rotatebox{90}{Patent B} & \rotatebox{90}{Red.}  & \textcolor[rgb]{ .753,  .314,  .302}{\rotatebox{90}{avg}} & \rotatebox{90}{CNN.}  & \rotatebox{90}{XSUM}  & \rotatebox{90}{Pubm.} & \rotatebox{90}{Patent B} & \rotatebox{90}{Red.}  & \textcolor[rgb]{ .753,  .314,  .302}{\rotatebox{90}{avg}} & \rotatebox{90}{CNN.}  & \rotatebox{90}{XSUM}  & \rotatebox{90}{Pubm.} & \rotatebox{90}{Patent B} & \rotatebox{90}{Red.}  & \textcolor[rgb]{ .753,  .314,  .302}{\rotatebox{90}{avg}} & \rotatebox{90}{CNN.}  & \rotatebox{90}{XSUM}  & \rotatebox{90}{Pubm.} & \rotatebox{90}{Patent B} & \rotatebox{90}{Red.}  & \textcolor[rgb]{ .753,  .314,  .302}{\rotatebox{90}{avg}} & \rotatebox{90}{CNN.}  & \rotatebox{90}{XSUM}  & \rotatebox{90}{Pubm.} & \rotatebox{90}{Patent B} & \rotatebox{90}{Red.}  & \textcolor[rgb]{ .753,  .314,  .302}{\rotatebox{90}{avg}} & \rotatebox{90}{CNN.}  & \rotatebox{90}{XSUM}  & \rotatebox{90}{Pubm.} & \rotatebox{90}{Patent B} & \rotatebox{90}{Red.}  & \textcolor[rgb]{ .753,  .314,  .302}{\rotatebox{90}{avg}} \\
    \midrule
    CNN   & 68.6  & 71.1  & 73.3  & 69.9  & 53.9  & \textcolor[rgb]{ .753,  .314,  .302}{67.4 } & 89.4  & 91.3  & 92.2  & 91.7  & 83.5  & \textcolor[rgb]{ .753,  .314,  .302}{89.6 } & 95.9  & 94.5  & 90.9  & 96.9  & 94.6  & \textcolor[rgb]{ .753,  .314,  .302}{94.6 } & 72.4  & 75.7  & 71.5  & 71.8  & 70.5  & \textcolor[rgb]{ .753,  .314,  .302}{72.4 } & 78.7  & 83.9  & 87.7  & 92.1  & 78.7  & \textcolor[rgb]{ .753,  .314,  .302}{84.2 } & 69.9  & 77.9  & 87.4  & 84.1  & 90.2  & \textcolor[rgb]{ .753,  .314,  .302}{81.9 } \\
    XSUM  & 13.4  & 23.5  & 18.1  & 13.2  & 31.0  & \textcolor[rgb]{ .753,  .314,  .302}{19.8 } & 6.3   & 17.8  & 9.0   & 8.2   & 23.2  & \textcolor[rgb]{ .753,  .314,  .302}{12.9 } & 7.4   & 18.1  & 11.0  & 7.6   & 6.5   & \textcolor[rgb]{ .753,  .314,  .302}{10.1 } & 9.7   & 22.6  & 10.8  & 9.9   & 19.1  & \textcolor[rgb]{ .753,  .314,  .302}{14.4 } & 14.5  & 21.1  & 29.8  & 8.7   & 31.3  & \textcolor[rgb]{ .753,  .314,  .302}{21.1 } & 35.5  & 24.7  & 36.1  & 50.1  & 50.7  & \textcolor[rgb]{ .753,  .314,  .302}{39.4 } \\
    Pubm. & 61.0  & 70.0  & 62.8  & 78.6  & 46.6  & \textcolor[rgb]{ .753,  .314,  .302}{63.8 } & 77.6  & 80.7  & 81.5  & 75.1  & 85.9  & \textcolor[rgb]{ .753,  .314,  .302}{80.2 } & 70.7  & 75.6  & 76.6  & 67.9  & 75.4  & \textcolor[rgb]{ .753,  .314,  .302}{73.2 } & 58.5  & 59.3  & 56.2  & 72.3  & 34.9  & \textcolor[rgb]{ .753,  .314,  .302}{56.2 } & 55.4  & 58.7  & 70.8  & 71.7  & 56.4  & \textcolor[rgb]{ .753,  .314,  .302}{62.6 } & 69.5  & 61.5  & 58.4  & 61.3  & 94.1  & \textcolor[rgb]{ .753,  .314,  .302}{69.0 } \\
    Patent B & 94.4  & 94.3  & 89.0  & 71.9  & 91.0  & \textcolor[rgb]{ .753,  .314,  .302}{88.1 } & 65.2  & 60.3  & 70.9  & 62.8  & 71.0  & \textcolor[rgb]{ .753,  .314,  .302}{66.0 } & 67.0  & 63.3  & 64.6  & 61.6  & 77.4  & \textcolor[rgb]{ .753,  .314,  .302}{66.8 } & 79.2  & 81.2  & 84.4  & 68.7  & 73.9  & \textcolor[rgb]{ .753,  .314,  .302}{77.5 } & 85.4  & 88.4  & 80.3  & 66.5  & 82.0  & \textcolor[rgb]{ .753,  .314,  .302}{80.6 } & 52.1  & 53.8  & 69.0  & 67.4  & 76.8  & \textcolor[rgb]{ .753,  .314,  .302}{63.8 } \\
    Red. & 20.9  & 40.2  & 11.1  & 13.2  & 50.9  & \textcolor[rgb]{ .753,  .314,  .302}{27.3 } & 37.2  & 21.5  & 55.2  & 62.6  & 61.1  & \textcolor[rgb]{ .753,  .314,  .302}{47.5 } & 27.4  & 23.5  & 42.9  & 49.7  & 62.2  & \textcolor[rgb]{ .753,  .314,  .302}{41.1 } & 34.8  & 35.7  & 50.6  & 44.6  & 52.5  & \textcolor[rgb]{ .753,  .314,  .302}{43.6 } & 17.2  & 25.7  & 25.1  & 30.0  & 50.3  & \textcolor[rgb]{ .753,  .314,  .302}{29.6 } & 59.6  & 50.3  & 69.1  & 49.3  & 44.2  & \textcolor[rgb]{ .753,  .314,  .302}{54.5 } \\
    \textcolor[rgb]{ .753,  .314,  .302}{avg} & \textcolor[rgb]{ .753,  .314,  .302}{51.7 } & \textcolor[rgb]{ .753,  .314,  .302}{59.8 } & \textcolor[rgb]{ .753,  .314,  .302}{50.9 } & \textcolor[rgb]{ .753,  .314,  .302}{49.4 } & \textcolor[rgb]{ .753,  .314,  .302}{54.7 } & \textcolor[rgb]{ .753,  .314,  .302}{53.3 } & \textcolor[rgb]{ .753,  .314,  .302}{55.2 } & \textcolor[rgb]{ .753,  .314,  .302}{54.3 } & \textcolor[rgb]{ .753,  .314,  .302}{61.8 } & \textcolor[rgb]{ .753,  .314,  .302}{60.1 } & \textcolor[rgb]{ .753,  .314,  .302}{65.0 } & \textcolor[rgb]{ .753,  .314,  .302}{59.2 } & \textcolor[rgb]{ .753,  .314,  .302}{53.7 } & \textcolor[rgb]{ .753,  .314,  .302}{55.0 } & \textcolor[rgb]{ .753,  .314,  .302}{57.2 } & \textcolor[rgb]{ .753,  .314,  .302}{56.7 } & \textcolor[rgb]{ .753,  .314,  .302}{63.2 } & \textcolor[rgb]{ .753,  .314,  .302}{57.2 } & \textcolor[rgb]{ .753,  .314,  .302}{50.9 } & \textcolor[rgb]{ .753,  .314,  .302}{54.9 } & \textcolor[rgb]{ .753,  .314,  .302}{54.7 } & \textcolor[rgb]{ .753,  .314,  .302}{53.5 } & \textcolor[rgb]{ .753,  .314,  .302}{50.2 } & \textcolor[rgb]{ .753,  .314,  .302}{52.8 } & \textcolor[rgb]{ .753,  .314,  .302}{50.2 } & \textcolor[rgb]{ .753,  .314,  .302}{55.6 } & \textcolor[rgb]{ .753,  .314,  .302}{58.7 } & \textcolor[rgb]{ .753,  .314,  .302}{53.8 } & \textcolor[rgb]{ .753,  .314,  .302}{59.8 } & \textcolor[rgb]{ .753,  .314,  .302}{55.6 } & \textcolor[rgb]{ .753,  .314,  .302}{57.3 } & \textcolor[rgb]{ .753,  .314,  .302}{53.6 } & \textcolor[rgb]{ .753,  .314,  .302}{64.0 } & \textcolor[rgb]{ .753,  .314,  .302}{62.4 } & \textcolor[rgb]{ .753,  .314,  .302}{71.2 } & \textcolor[rgb]{ .753,  .314,  .302}{61.7 } \\
    \bottomrule
    \end{tabular}}%
    \caption{factcc result for Abstractive models}
  \label{tab:all factcc abs}%
\end{table*}%

\begin{table*}[htbp]
  \setlength{\tabcolsep}{1.5pt}
        \renewcommand{\arraystretch}{1.3}
      \centering
      \scriptsize
        \resizebox{0.95\textwidth}{16mm}{
    \begin{tabular}{ccccccccccccccccccccccccccccccc}
    \toprule
    \multirow{2}[4]{*}{\textbf{EXT models}} & \multicolumn{5}{c}{LSTM$_{non}$}           &       & \multicolumn{5}{c}{Trans$_{non}$}          &       & \multicolumn{5}{c}{Trans$_{auto}$}         &       & \multicolumn{5}{c}{BERT$_{non}$}           &       & \multicolumn{5}{c}{BERT$_{match}$}         &  \\
\cmidrule{2-6}\cmidrule{8-12}\cmidrule{14-18}\cmidrule{20-24}\cmidrule{26-30}          & \rotatebox{90}{CNN.}  & \rotatebox{90}{XSUM}  & \rotatebox{90}{Pubm.} & \rotatebox{90}{Patent B} & \rotatebox{90}{Red.}  & \textcolor[rgb]{ .753,  .314,  .302}{\rotatebox{90}{avg}} & \rotatebox{90}{CNN.}  & \rotatebox{90}{XSUM}  & \rotatebox{90}{Pubm.} & \rotatebox{90}{Patent B} & \rotatebox{90}{Red.}  & \textcolor[rgb]{ .753,  .314,  .302}{\rotatebox{90}{avg}} & \rotatebox{90}{CNN.}  & \rotatebox{90}{XSUM}  & \rotatebox{90}{Pubm.} & \rotatebox{90}{Patent B} & \rotatebox{90}{Red.}  & \textcolor[rgb]{ .753,  .314,  .302}{\rotatebox{90}{avg}} & \rotatebox{90}{CNN.}  & \rotatebox{90}{XSUM}  & \rotatebox{90}{Pubm.} & \rotatebox{90}{Patent B} & \rotatebox{90}{Red.}  & \textcolor[rgb]{ .753,  .314,  .302}{\rotatebox{90}{avg}} & \rotatebox{90}{CNN.}  & \rotatebox{90}{XSUM}  & \rotatebox{90}{Pubm.} & \rotatebox{90}{Patent B} & \rotatebox{90}{Red.}  & \textcolor[rgb]{ .753,  .314,  .302}{\rotatebox{90}{avg}} \\
    \midrule
    CNN   & 99.2  & 99.9  & 96.0  & 99.1  & 95.2  & \textcolor[rgb]{ .753,  .314,  .302}{97.9 } & 100.0  & 100.0  & 98.0  & 99.1  & 100.0  & \textcolor[rgb]{ .753,  .314,  .302}{99.4 } & 98.1  & 100.0  & 91.3  & 93.5  & 100.0  & \textcolor[rgb]{ .753,  .314,  .302}{96.6 } & 99.6  & 99.9  & 97.3  & 98.2  & 98.6  & \textcolor[rgb]{ .753,  .314,  .302}{98.7 } & 99.8  & 99.4  & 92.9  & 95.7  & 99.1  & \textcolor[rgb]{ .753,  .314,  .302}{97.4 } \\
    XSUM  & 84.1  & 94.3  & 90.3  & 81.4  & 94.1  & \textcolor[rgb]{ .753,  .314,  .302}{88.9 } & 99.8  & 100.0  & 97.4  & 98.2  & 100.0  & \textcolor[rgb]{ .753,  .314,  .302}{99.1 } & 86.8  & 99.3  & 82.9  & 69.9  & 100.0  & \textcolor[rgb]{ .753,  .314,  .302}{87.8 } & 98.4  & 99.7  & 96.6  & 95.7  & 99.9  & \textcolor[rgb]{ .753,  .314,  .302}{98.1 } & 99.7  & 99.5  & 93.2  & 95.1  & 98.8  & \textcolor[rgb]{ .753,  .314,  .302}{97.3 } \\
    Pubm. & 70.5  & 84.3  & 80.8  & 65.1  & 89.0  & \textcolor[rgb]{ .753,  .314,  .302}{77.9 } & 97.7  & 98.8  & 95.1  & 94.7  & 100.0  & \textcolor[rgb]{ .753,  .314,  .302}{97.3 } & 87.5  & 99.6  & 79.0  & 64.4  & 99.7  & \textcolor[rgb]{ .753,  .314,  .302}{86.1 } & 95.3  & 99.3  & 95.1  & 94.3  & 99.5  & \textcolor[rgb]{ .753,  .314,  .302}{96.7 } & 99.7  & 99.2  & 93.1  & 95.2  & 99.3  & \textcolor[rgb]{ .753,  .314,  .302}{97.3 } \\
    Patent B & 86.1  & 96.0  & 90.9  & 74.1  & 96.0  & \textcolor[rgb]{ .753,  .314,  .302}{88.6 } & 98.3  & 99.8  & 96.3  & 97.4  & 99.5  & \textcolor[rgb]{ .753,  .314,  .302}{98.3 } & 90.7  & 99.8  & 85.5  & 68.8  & 99.7  & \textcolor[rgb]{ .753,  .314,  .302}{88.9 } & 97.0  & 99.0  & 96.0  & 94.8  & 99.1  & \textcolor[rgb]{ .753,  .314,  .302}{97.2 } & 99.7  & 99.0  & 93.0  & 94.5  & 98.4  & \textcolor[rgb]{ .753,  .314,  .302}{96.9 } \\
    Red. & 81.0  & 92.1  & 86.9  & 64.6  & 90.2  & \textcolor[rgb]{ .753,  .314,  .302}{83.0 } & 90.3  & 94.1  & 94.1  & 86.7  & 96.3  & \textcolor[rgb]{ .753,  .314,  .302}{92.3 } & 79.4  & 98.7  & 79.6  & 56.4  & 98.1  & \textcolor[rgb]{ .753,  .314,  .302}{82.5 } & 97.0  & 98.9  & 95.3  & 91.9  & 98.8  & \textcolor[rgb]{ .753,  .314,  .302}{96.4 } & 99.7  & 99.3  & 93.1  & 96.1  & 99.3  & \textcolor[rgb]{ .753,  .314,  .302}{97.5 } \\
    \textcolor[rgb]{ .753,  .314,  .302}{avg} & \textcolor[rgb]{ .753,  .314,  .302}{84.2 } & \textcolor[rgb]{ .753,  .314,  .302}{93.3 } & \textcolor[rgb]{ .753,  .314,  .302}{89.0 } & \textcolor[rgb]{ .753,  .314,  .302}{76.8 } & \textcolor[rgb]{ .753,  .314,  .302}{92.9 } & \textcolor[rgb]{ .753,  .314,  .302}{87.2 } & \textcolor[rgb]{ .753,  .314,  .302}{97.2 } & \textcolor[rgb]{ .753,  .314,  .302}{98.6 } & \textcolor[rgb]{ .753,  .314,  .302}{96.2 } & \textcolor[rgb]{ .753,  .314,  .302}{95.2 } & \textcolor[rgb]{ .753,  .314,  .302}{99.2 } & \textcolor[rgb]{ .753,  .314,  .302}{97.3 } & \textcolor[rgb]{ .753,  .314,  .302}{88.5 } & \textcolor[rgb]{ .753,  .314,  .302}{99.5 } & \textcolor[rgb]{ .753,  .314,  .302}{83.7 } & \textcolor[rgb]{ .753,  .314,  .302}{70.6 } & \textcolor[rgb]{ .753,  .314,  .302}{99.5 } & \textcolor[rgb]{ .753,  .314,  .302}{88.4 } & \textcolor[rgb]{ .753,  .314,  .302}{97.5 } & \textcolor[rgb]{ .753,  .314,  .302}{99.4 } & \textcolor[rgb]{ .753,  .314,  .302}{96.1 } & \textcolor[rgb]{ .753,  .314,  .302}{95.0 } & \textcolor[rgb]{ .753,  .314,  .302}{99.2 } & \textcolor[rgb]{ .753,  .314,  .302}{97.4 } & \textcolor[rgb]{ .753,  .314,  .302}{99.7 } & \textcolor[rgb]{ .753,  .314,  .302}{99.3 } & \textcolor[rgb]{ .753,  .314,  .302}{93.0 } & \textcolor[rgb]{ .753,  .314,  .302}{95.3 } & \textcolor[rgb]{ .753,  .314,  .302}{99.0 } & \textcolor[rgb]{ .753,  .314,  .302}{97.3 } \\
    \bottomrule
    \end{tabular}}%
    \caption{factcc result for Extractive models}
  \label{tab:all factcc ext}%
\end{table*}%

\begin{table*}[htbp]
  \centering
  \Huge
  \setlength{\tabcolsep}{16pt}
  \resizebox{0.999\textwidth}{50mm}{
    \begin{tabular}{c|c|rrllllllrllllllrllllllrllllllrllllllrllllllrllllllr}
    \toprule
    \multicolumn{3}{c}{analysis aspect} &       & \multicolumn{49}{c}{Architecture} \\
\cmidrule{1-3}\cmidrule{5-53}    \multicolumn{3}{c}{model type} &       & \multicolumn{49}{c}{ABS} \\
\cmidrule{1-3}\cmidrule{5-53}    \multicolumn{3}{c}{compare models} &       & \multicolumn{6}{c}{L2L$_{ptr}$ vs. L2L}            &       & \multicolumn{6}{c}{L2L$_{ptr}^{cov}$ vs. L2L$_{ptr}$}      &       & \multicolumn{6}{c}{T2T vs. L2L}               &       & \multicolumn{6}{c}{BE2T vs. T2T}              &       & \multicolumn{6}{c}{BART vs. BE2T}             &       & \multicolumn{6}{c}{BART vs. L2L}              &       & \multicolumn{6}{c}{BART vs. T2T}              &  \\
\cmidrule{1-10}\cmidrule{12-17}\cmidrule{19-24}\cmidrule{26-31}\cmidrule{33-38}\cmidrule{40-45}\cmidrule{47-52}    \multicolumn{3}{c}{\multirow{2}[2]{*}{holistic analysis}} &       & \multicolumn{6}{c}{stiff. : 20.74 vs. 18.03}  &       & \multicolumn{6}{c}{stiff. : 22.81 vs. 20.74}  &       & \multicolumn{6}{c}{stiff. : 19.79 vs. 18.03 } &       & \multicolumn{6}{c}{stiff. : 23.49 vs 19.79}   &       & \multicolumn{6}{c}{stiff. : 31.66 vs. 23.49}  &       & \multicolumn{6}{c}{stiff. : 31.66 vs. 18.03 } &       & \multicolumn{6}{c}{stiff. : 31.66 vs. 19.79}  &  \\
    \multicolumn{3}{c}{}  &       & \multicolumn{6}{c}{stable. : 68.63 vs. 66.93} &       & \multicolumn{6}{c}{stable. : 70.71 vs. 68.63} &       & \multicolumn{6}{c}{stable. : 62.12 vs. 66.93} &       & \multicolumn{6}{c}{stable. : 62.93 vs. 62.12} &       & \multicolumn{6}{c}{stable. : 73.83 vs. 62.93} &       & \multicolumn{6}{c}{stable. : 73.83 vs. 66.93} &       & \multicolumn{6}{c}{stable. : 73.83 vs. 62.12} &  \\
\cmidrule{1-3}\cmidrule{5-10}\cmidrule{12-17}\cmidrule{19-24}\cmidrule{26-31}\cmidrule{33-38}\cmidrule{40-45}\cmidrule{47-52}    \multicolumn{3}{c}{fine-grain analysis} &       & \multicolumn{1}{c}{\rotatebox{90}{CNN.}} & \multicolumn{1}{c}{\rotatebox{90}{Xsum}} & \multicolumn{1}{c}{\rotatebox{90}{Pubm.}} & \multicolumn{1}{c}{\rotatebox{90}{Patent b}} & \multicolumn{1}{c}{\rotatebox{90}{Red.}} & \multicolumn{1}{c}{\textcolor[rgb]{ 1,  0,  0}{\rotatebox{90}{avg}}} &       & \multicolumn{1}{c}{\rotatebox{90}{CNN.}} & \multicolumn{1}{c}{\rotatebox{90}{Xsum}} & \multicolumn{1}{c}{\rotatebox{90}{Pubm.}} & \multicolumn{1}{c}{\rotatebox{90}{Patent b}} & \multicolumn{1}{c}{\rotatebox{90}{Red.}} & \multicolumn{1}{c}{\textcolor[rgb]{ 1,  0,  0}{\rotatebox{90}{avg}}} &       & \multicolumn{1}{c}{\rotatebox{90}{CNN.}} & \multicolumn{1}{c}{\rotatebox{90}{Xsum}} & \multicolumn{1}{c}{\rotatebox{90}{Pubm.}} & \multicolumn{1}{c}{\rotatebox{90}{Patent b}} & \multicolumn{1}{c}{\rotatebox{90}{Red.}} & \multicolumn{1}{c}{\textcolor[rgb]{ 1,  0,  0}{\rotatebox{90}{avg}}} &       & \multicolumn{1}{c}{\rotatebox{90}{CNN.}} & \multicolumn{1}{c}{\rotatebox{90}{Xsum}} & \multicolumn{1}{c}{\rotatebox{90}{Pubm.}} & \multicolumn{1}{c}{\rotatebox{90}{Patent b}} & \multicolumn{1}{c}{\rotatebox{90}{Red.}} & \multicolumn{1}{c}{\textcolor[rgb]{ 1,  0,  0}{\rotatebox{90}{avg}}} &       & \multicolumn{1}{c}{\rotatebox{90}{CNN.}} & \multicolumn{1}{c}{\rotatebox{90}{Xsum}} & \multicolumn{1}{c}{\rotatebox{90}{Pubm.}} & \multicolumn{1}{c}{\rotatebox{90}{Patent b}} & \multicolumn{1}{c}{\rotatebox{90}{Red.}} & \multicolumn{1}{c}{\textcolor[rgb]{ 1,  0,  0}{\rotatebox{90}{avg}}} &       & \multicolumn{1}{c}{\rotatebox{90}{CNN.}} & \multicolumn{1}{c}{\rotatebox{90}{Xsum}} & \multicolumn{1}{c}{\rotatebox{90}{Pubm.}} & \multicolumn{1}{c}{\rotatebox{90}{Patent b}} & \multicolumn{1}{c}{\rotatebox{90}{Red.}} & \multicolumn{1}{c}{\textcolor[rgb]{ 1,  0,  0}{\rotatebox{90}{avg}}} &       & \multicolumn{1}{c}{\rotatebox{90}{CNN.}} & \multicolumn{1}{c}{\rotatebox{90}{Xsum}} & \multicolumn{1}{c}{\rotatebox{90}{Pubm.}} & \multicolumn{1}{c}{\rotatebox{90}{Patent b}} & \multicolumn{1}{c}{\rotatebox{90}{Red.}} & \multicolumn{1}{c}{\textcolor[rgb]{ 1,  0,  0}{\rotatebox{90}{avg}}} &  \\
\cmidrule{1-10}\cmidrule{12-17}\cmidrule{19-24}\cmidrule{26-31}\cmidrule{33-38}\cmidrule{40-45}\cmidrule{47-52}    \multirow{12}[4]{*}{\rotatebox{90}{ROUGE}} & \multirow{6}[2]{*}{\rotatebox{90}{origin}} & CNN.  &       & \multicolumn{6}{l}{\multirow{6}[2]{*}{\includegraphics[height=200pt,width=280pt]{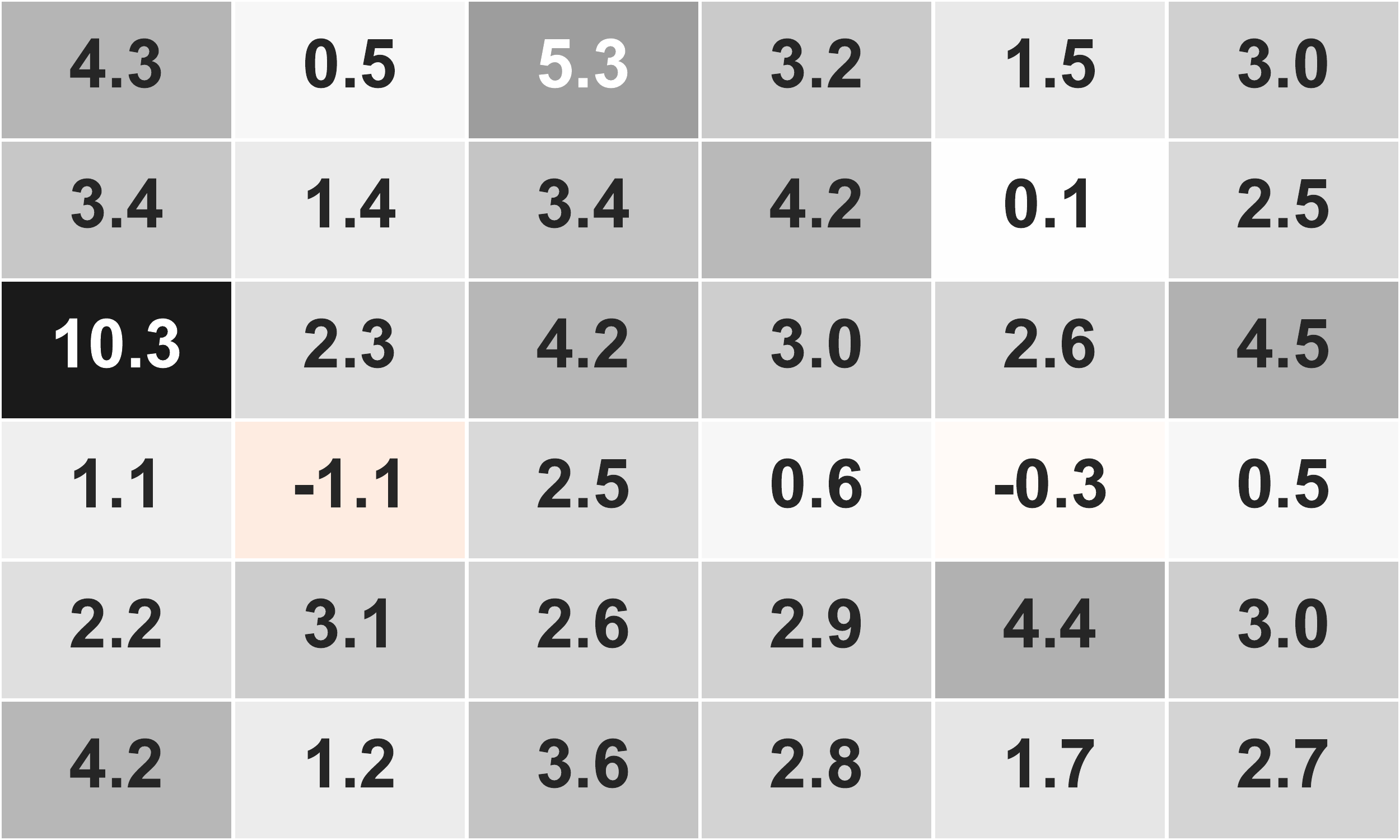} }}  &       & \multicolumn{6}{l}{\multirow{6}[2]{*}{\includegraphics[height=200pt,width=280pt]{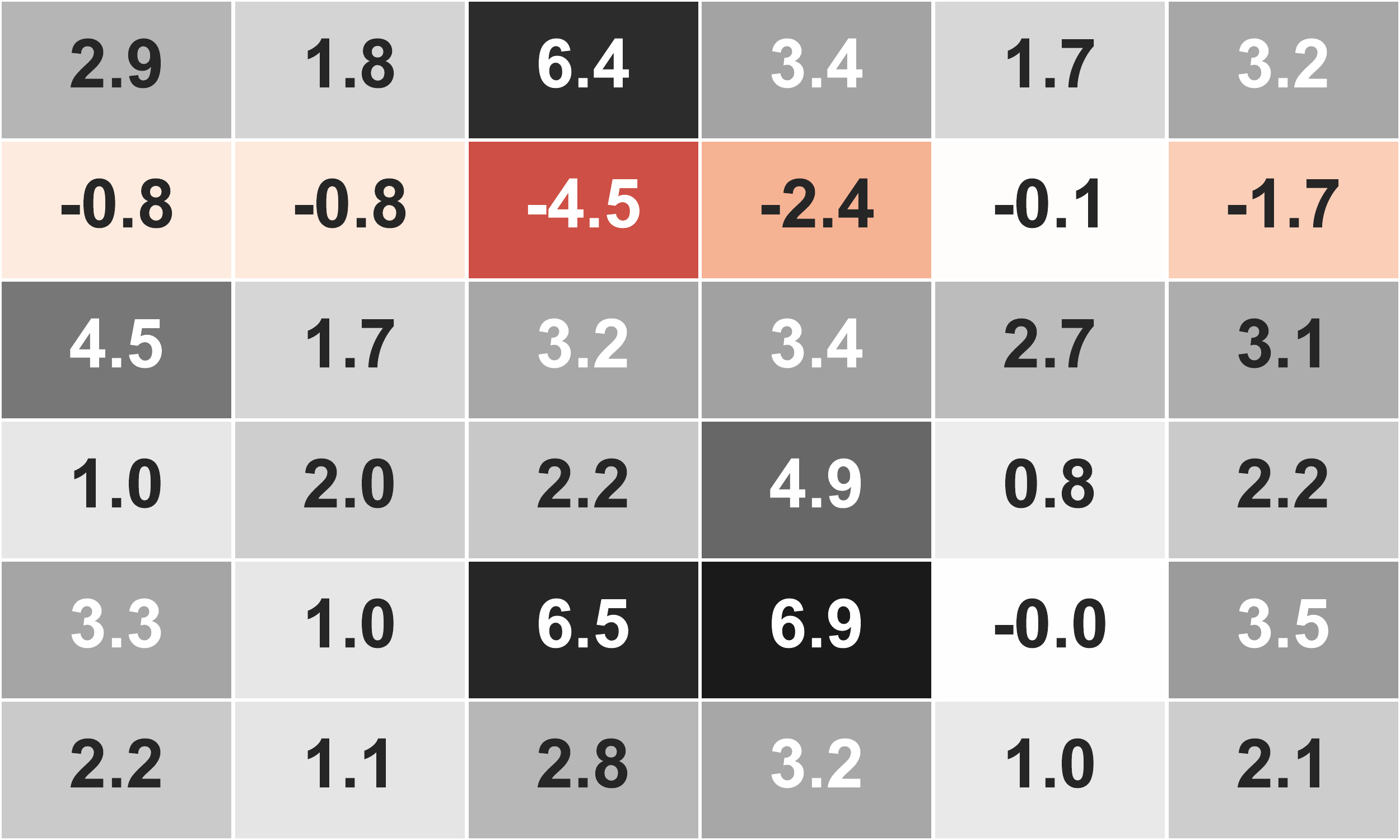} }}  &       & \multicolumn{6}{l}{\multirow{6}[2]{*}{\includegraphics[height=200pt,width=280pt]{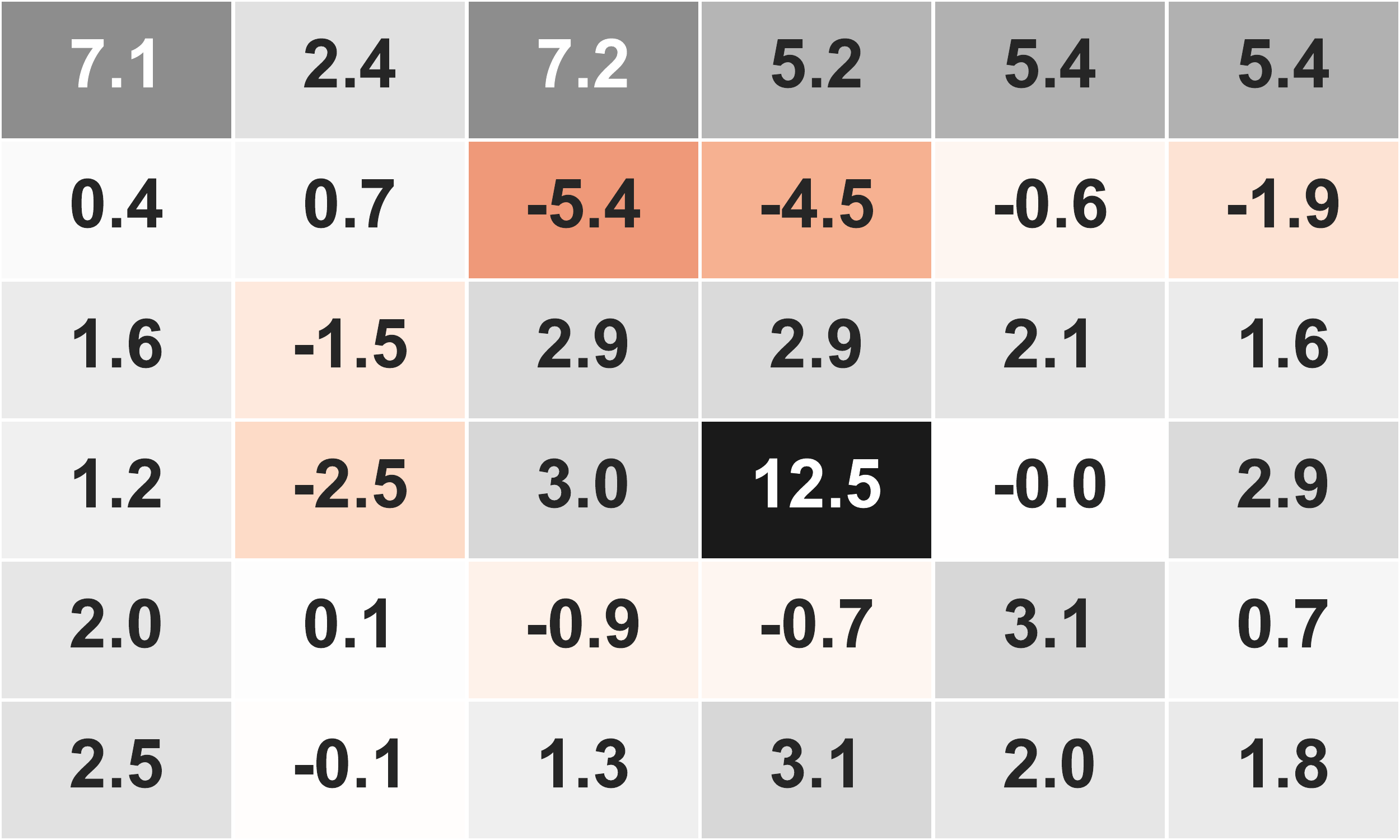} }}  &       & \multicolumn{6}{l}{\multirow{6}[2]{*}{\includegraphics[height=200pt,width=280pt]{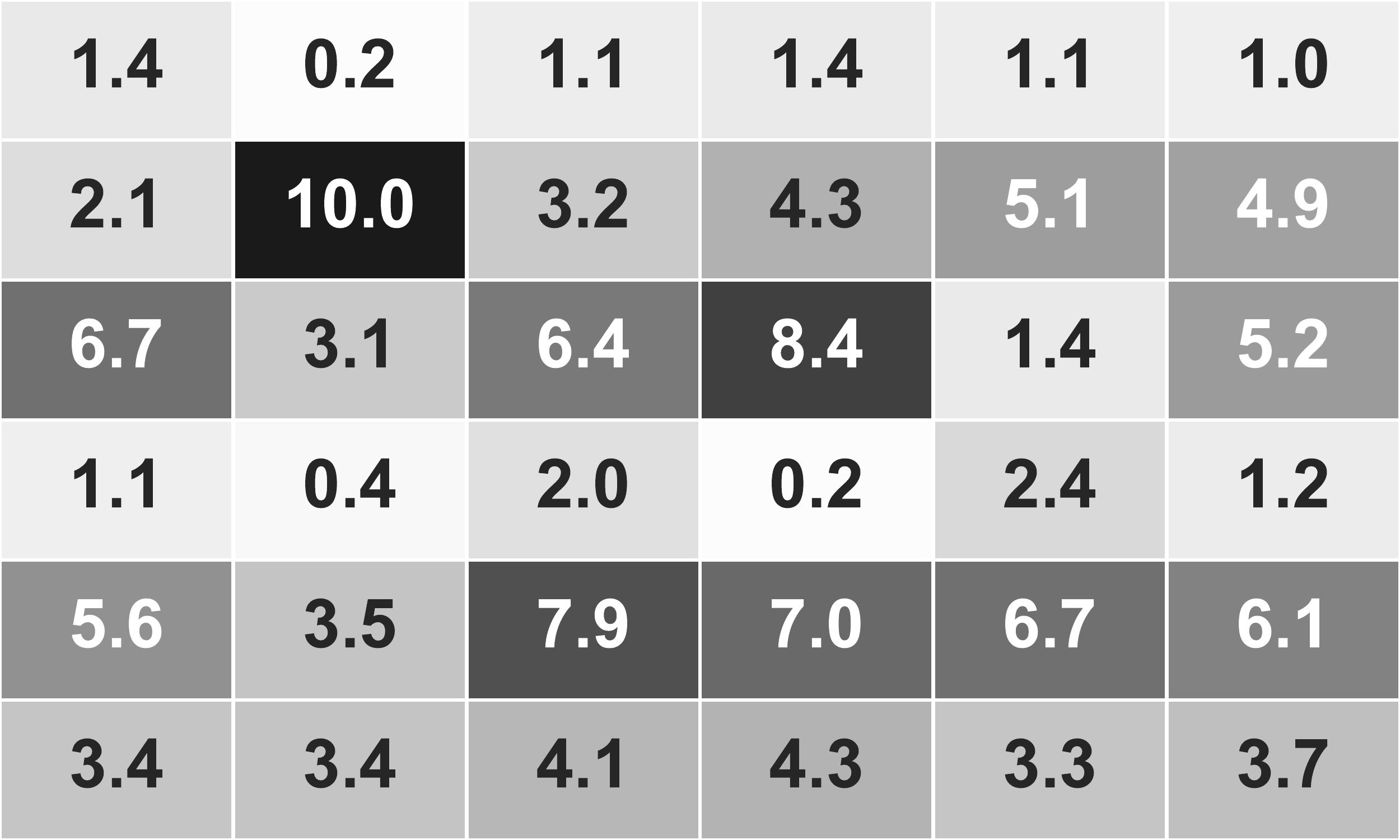} }}  &       & \multicolumn{6}{l}{\multirow{6}[2]{*}{\includegraphics[height=200pt,width=280pt]{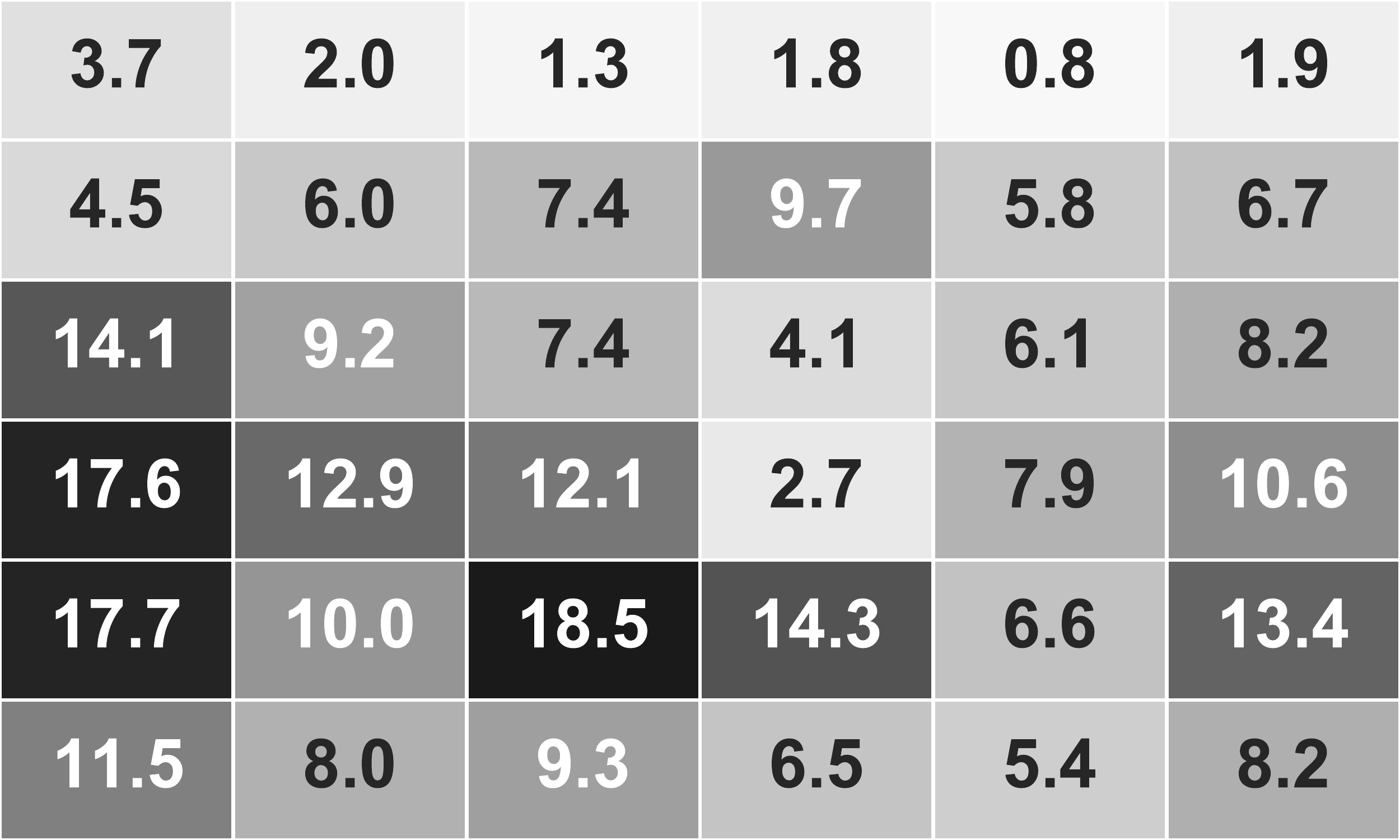} }}  &       & \multicolumn{6}{l}{\multirow{6}[2]{*}{\includegraphics[height=200pt,width=280pt]{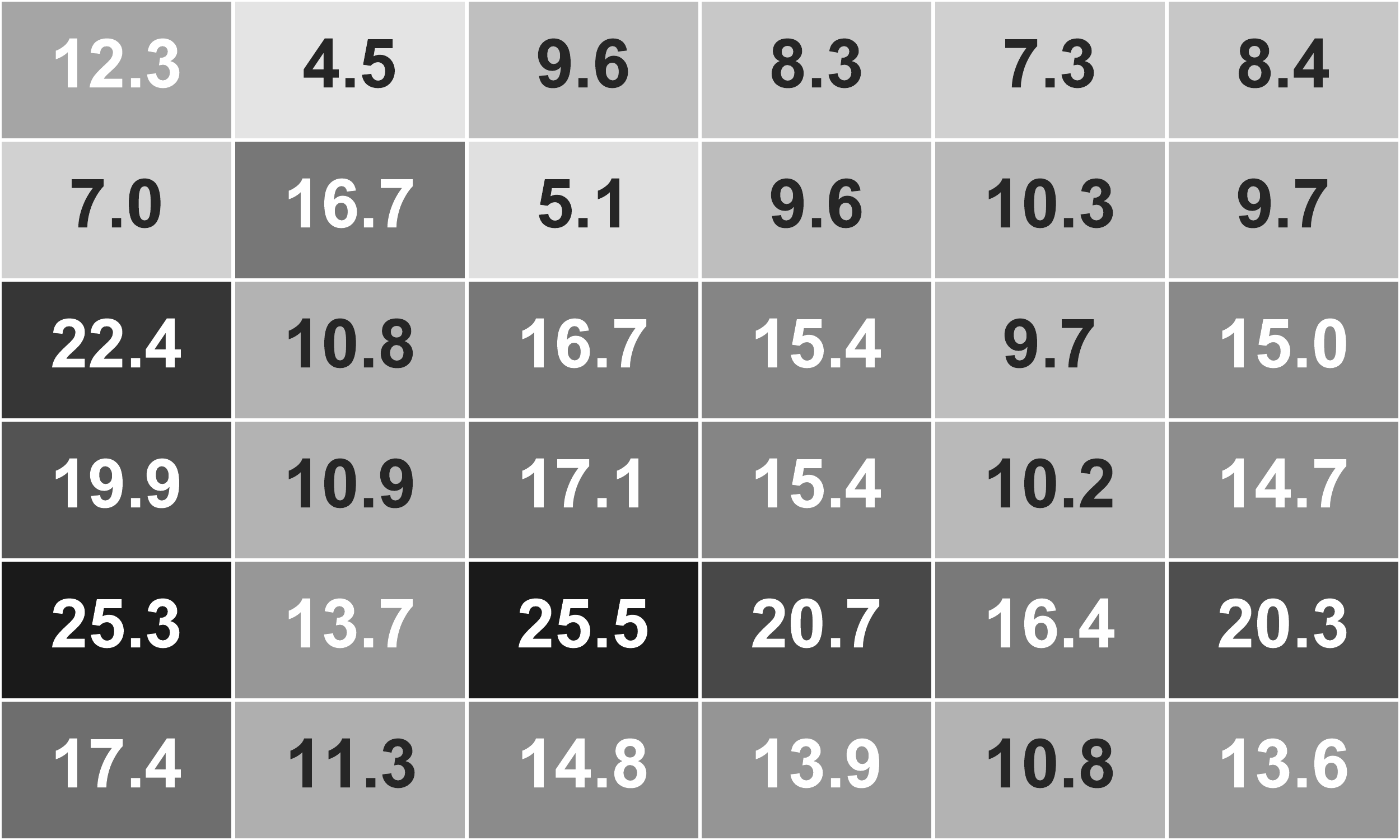} }}  &       & \multicolumn{6}{l}{\multirow{6}[2]{*}{\includegraphics[height=200pt,width=280pt]{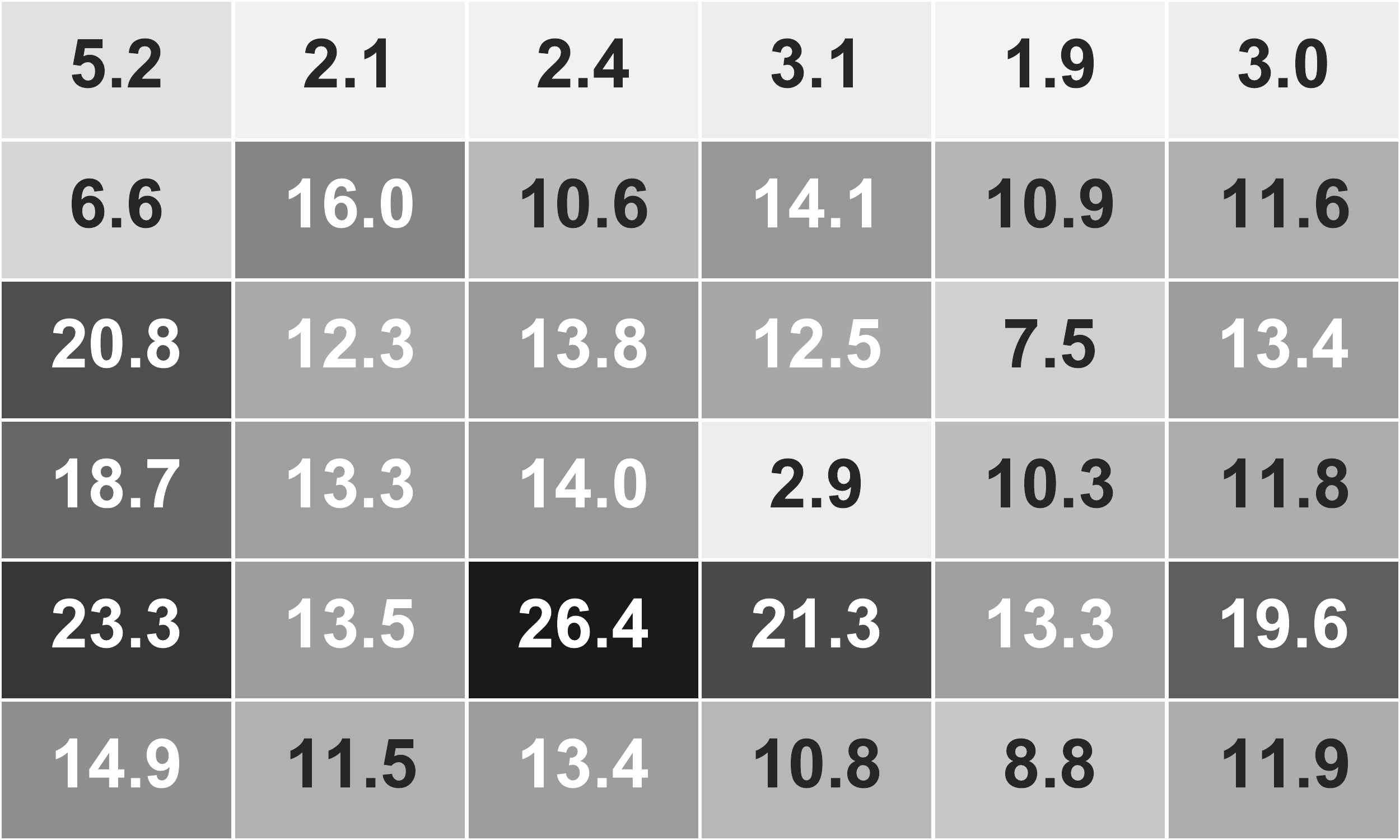} }}  &  \\
          &       & Xsum  &       & \multicolumn{6}{l}{}                          &       & \multicolumn{6}{l}{}                          &       & \multicolumn{6}{l}{}                          &       & \multicolumn{6}{l}{}                          &       & \multicolumn{6}{l}{}                          &       & \multicolumn{6}{l}{}                          &       & \multicolumn{6}{l}{}                          &  \\
          &       & Pubm. &       & \multicolumn{6}{l}{}                          &       & \multicolumn{6}{l}{}                          &       & \multicolumn{6}{l}{}                          &       & \multicolumn{6}{l}{}                          &       & \multicolumn{6}{l}{}                          &       & \multicolumn{6}{l}{}                          &       & \multicolumn{6}{l}{}                          &  \\
          &       & Patent b &       & \multicolumn{6}{l}{}                          &       & \multicolumn{6}{l}{}                          &       & \multicolumn{6}{l}{}                          &       & \multicolumn{6}{l}{}                          &       & \multicolumn{6}{l}{}                          &       & \multicolumn{6}{l}{}                          &       & \multicolumn{6}{l}{}                          &  \\
          &       & Red.  &       & \multicolumn{6}{l}{}                          &       & \multicolumn{6}{l}{}                          &       & \multicolumn{6}{l}{}                          &       & \multicolumn{6}{l}{}                          &       & \multicolumn{6}{l}{}                          &       & \multicolumn{6}{l}{}                          &       & \multicolumn{6}{l}{}                          &  \\
          &       & \textcolor[rgb]{ 1,  0,  0}{avg} &       & \multicolumn{6}{l}{}                          &       & \multicolumn{6}{l}{}                          &       & \multicolumn{6}{l}{}                          &       & \multicolumn{6}{l}{}                          &       & \multicolumn{6}{l}{}                          &       & \multicolumn{6}{l}{}                          &       & \multicolumn{6}{l}{}                          &  \\
\cmidrule{5-10}\cmidrule{12-17}\cmidrule{19-24}\cmidrule{26-31}\cmidrule{33-38}\cmidrule{40-45}\cmidrule{47-52}          & \multirow{6}[2]{*}{\rotatebox{90}{normali.}} & CNN.  &       & \multicolumn{6}{l}{\multirow{6}[2]{*}{\includegraphics[height=200pt,width=280pt]{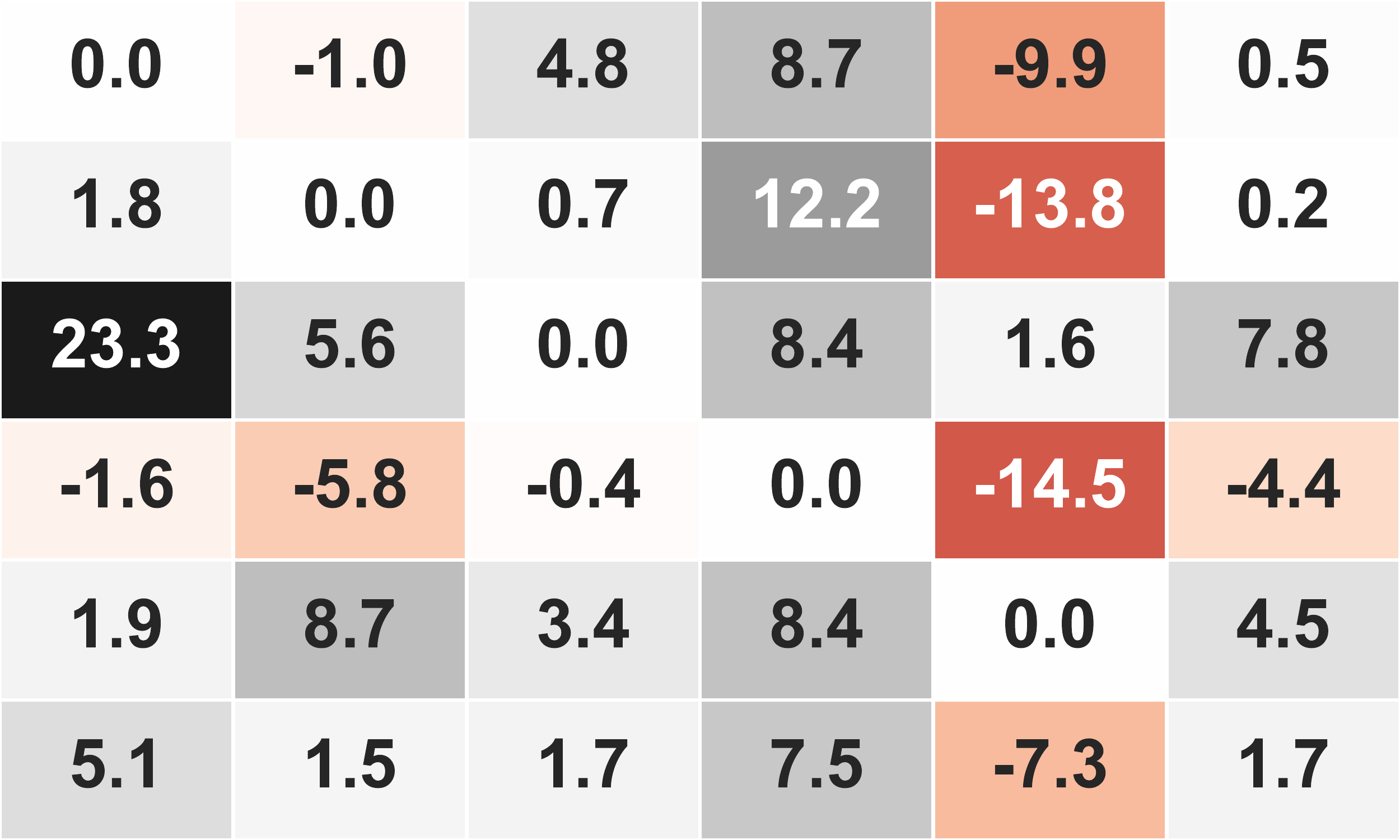} }}  &       & \multicolumn{6}{l}{\multirow{6}[2]{*}{\includegraphics[height=200pt,width=280pt]{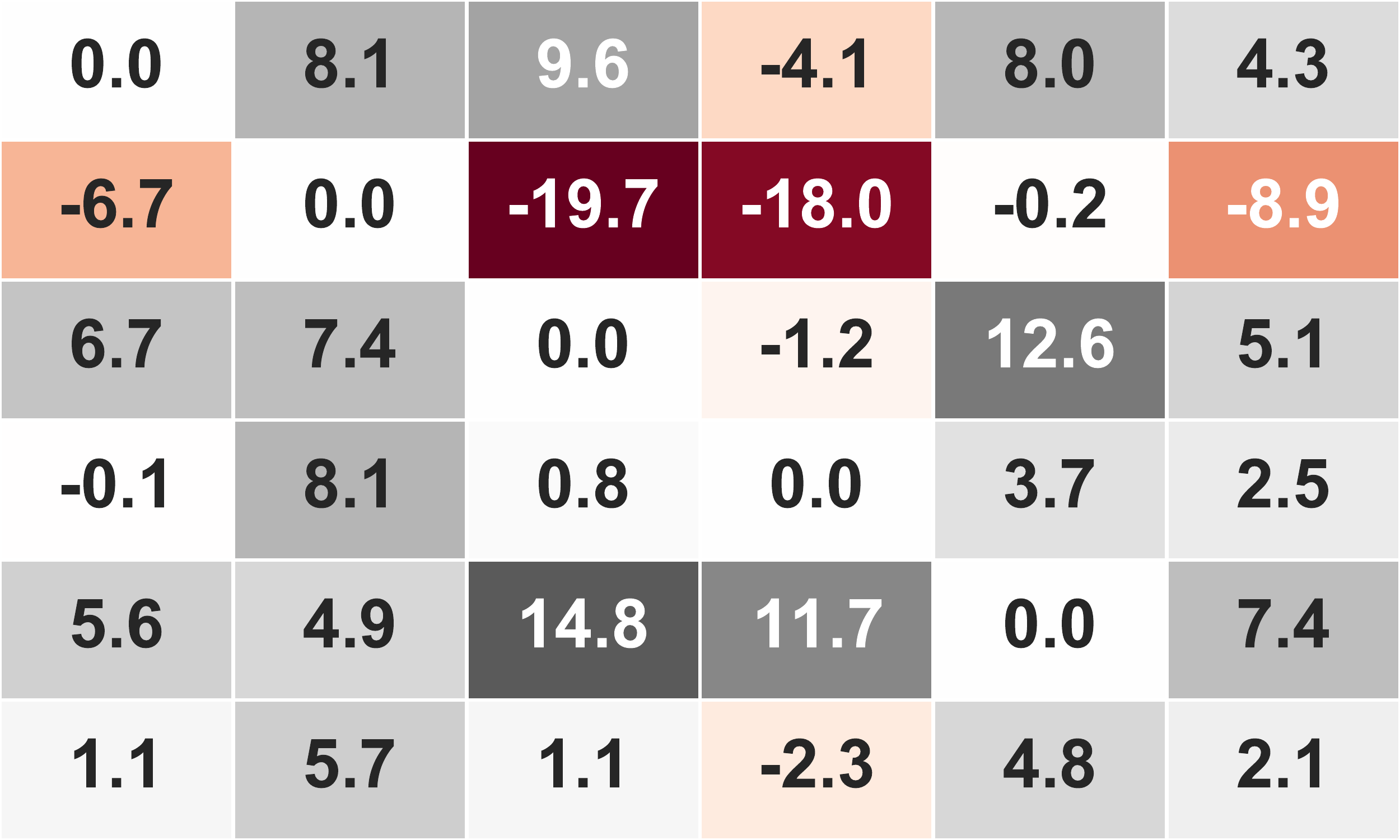}}}  &       & \multicolumn{6}{l}{\multirow{6}[2]{*}{\includegraphics[height=200pt,width=280pt]{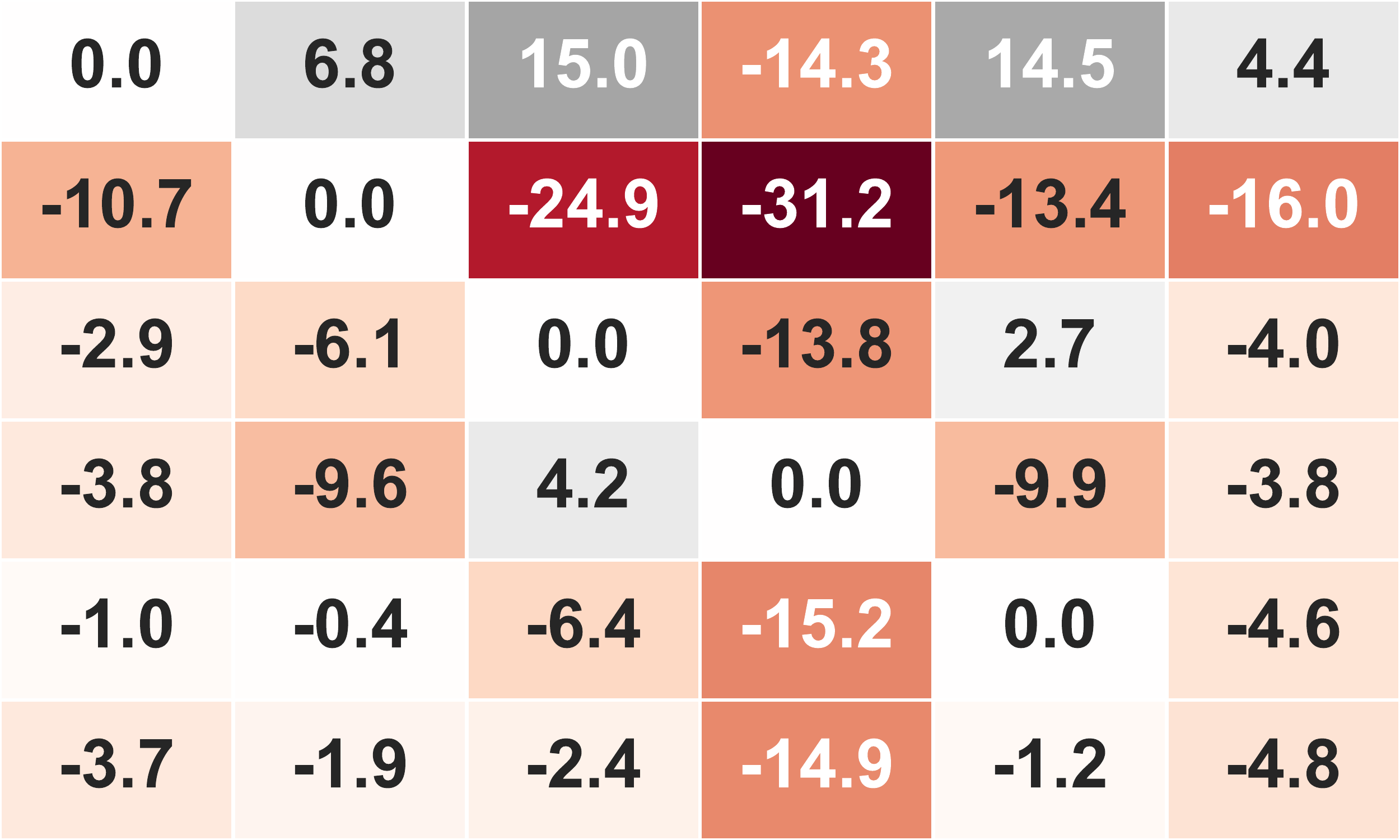} }}  &       & \multicolumn{6}{l}{\multirow{6}[2]{*}{\includegraphics[height=200pt,width=280pt]{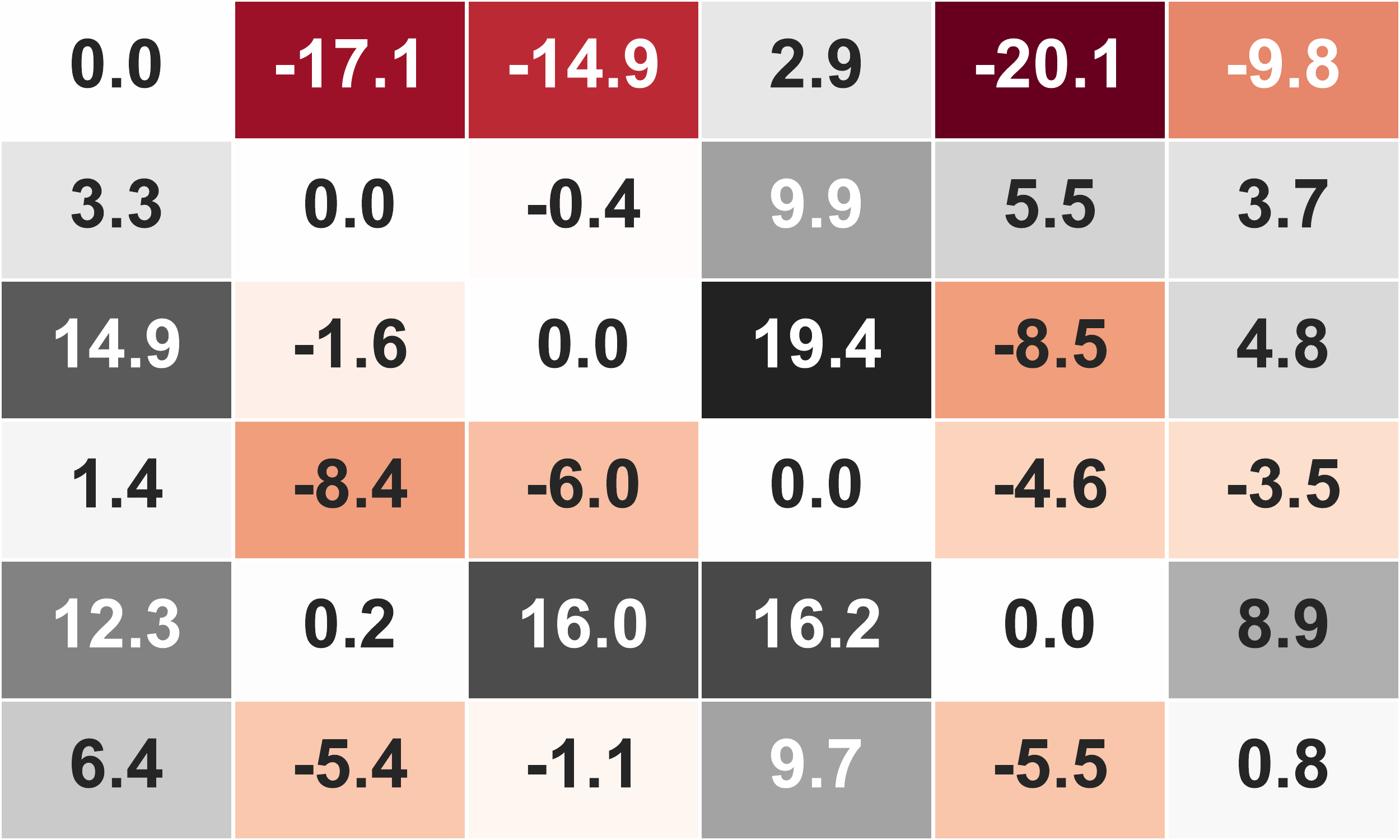} }}  &       & \multicolumn{6}{l}{\multirow{6}[2]{*}{\includegraphics[height=200pt,width=280pt]{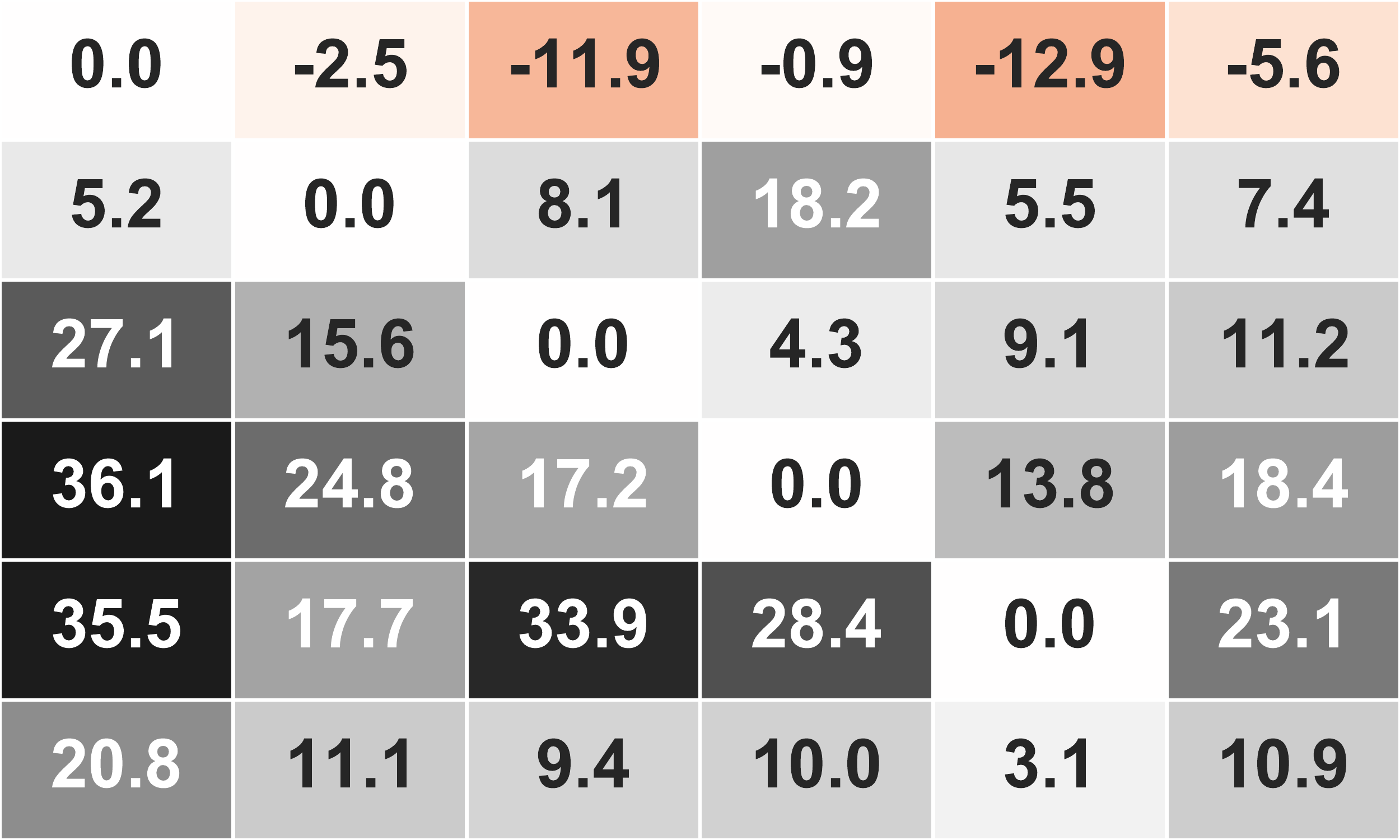} }}  &       & \multicolumn{6}{l}{\multirow{6}[2]{*}{\includegraphics[height=200pt,width=280pt]{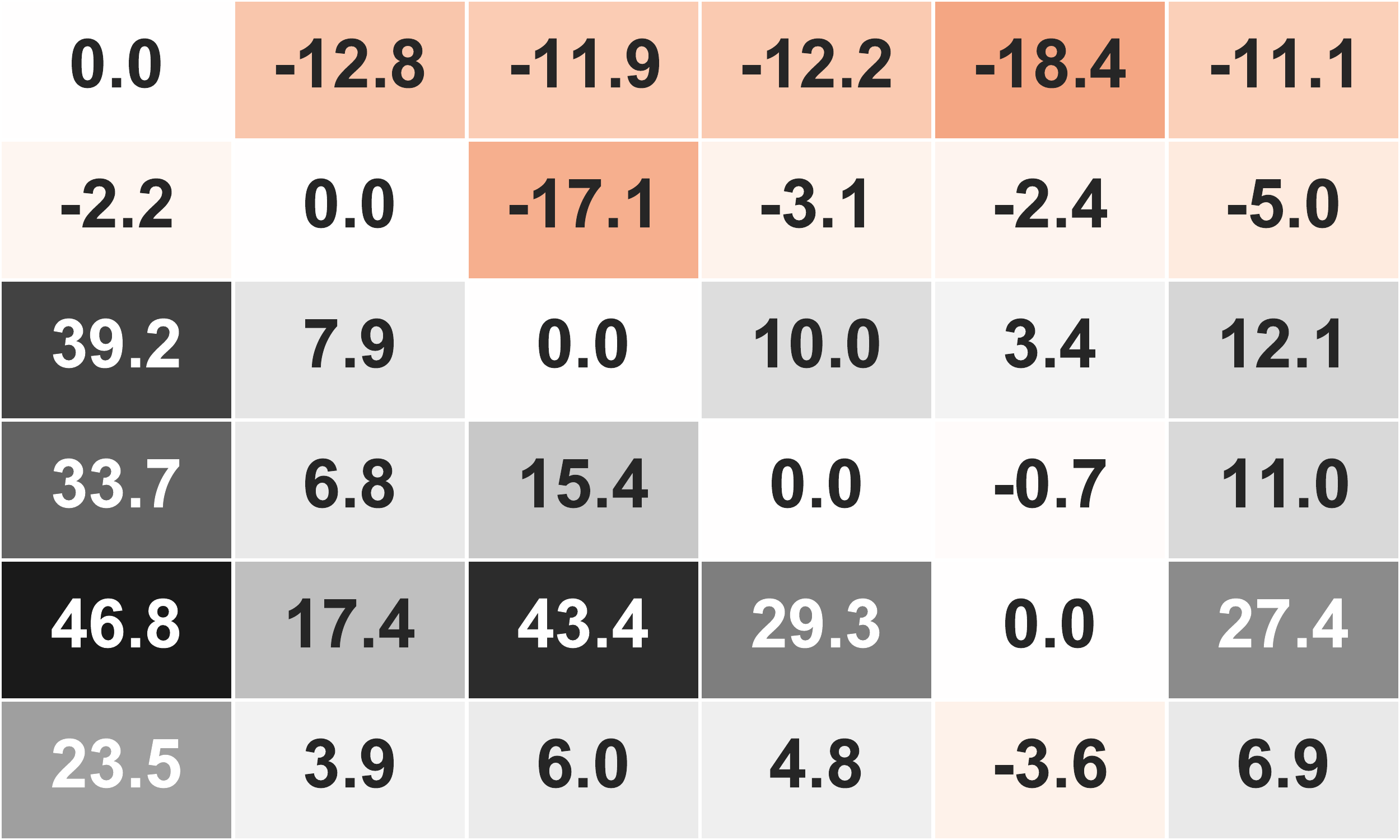} }}  &       & \multicolumn{6}{l}{\multirow{6}[2]{*}{\includegraphics[height=200pt,width=280pt]{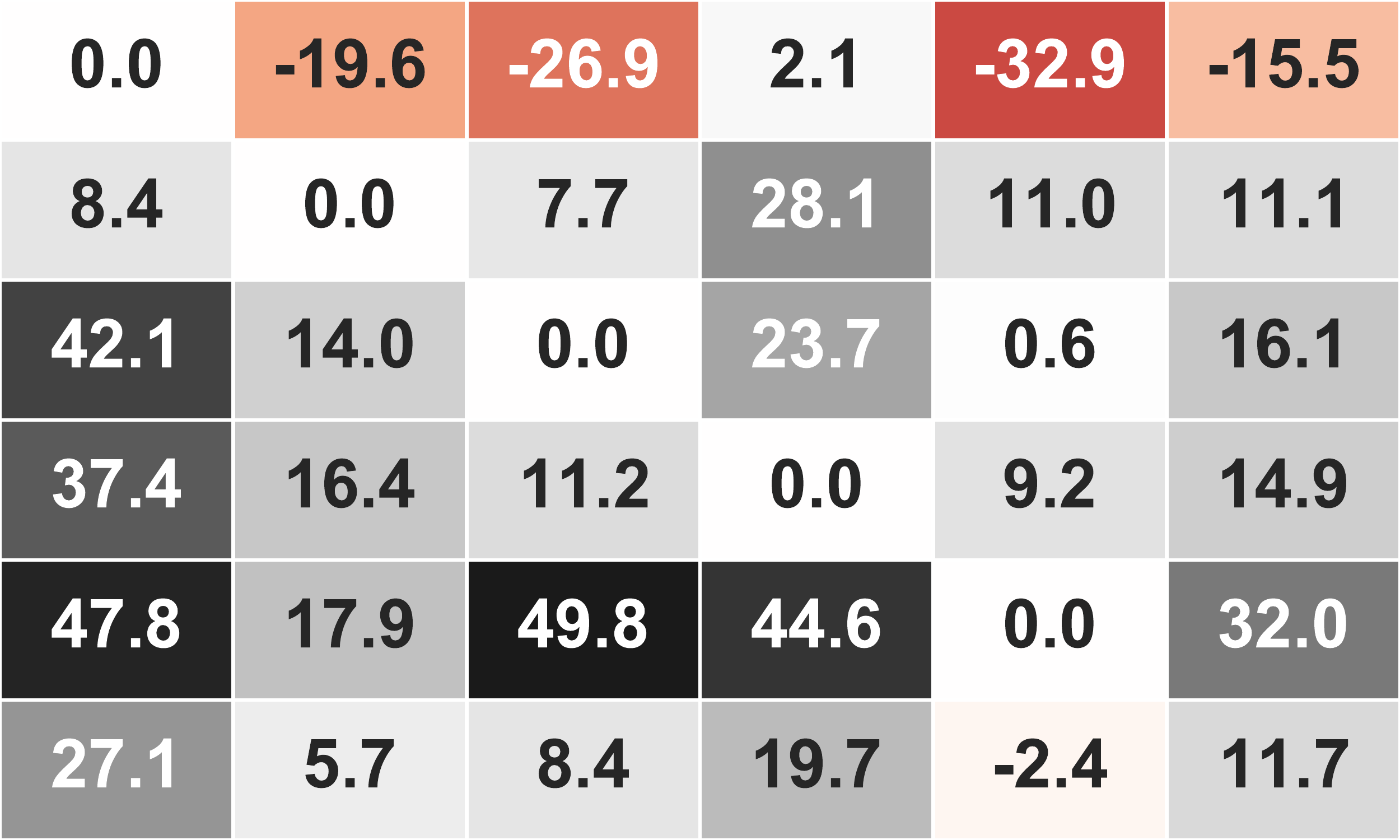} }}  &  \\
          &       & Xsum  &       & \multicolumn{6}{l}{}                          &       & \multicolumn{6}{l}{}                          &       & \multicolumn{6}{l}{}                          &       & \multicolumn{6}{l}{}                          &       & \multicolumn{6}{l}{}                          &       & \multicolumn{6}{l}{}                          &       & \multicolumn{6}{l}{}                          &  \\
          &       & Pubm. &       & \multicolumn{6}{l}{}                          &       & \multicolumn{6}{l}{}                          &       & \multicolumn{6}{l}{}                          &       & \multicolumn{6}{l}{}                          &       & \multicolumn{6}{l}{}                          &       & \multicolumn{6}{l}{}                          &       & \multicolumn{6}{l}{}                          &  \\
          &       & Patent b &       & \multicolumn{6}{l}{}                          &       & \multicolumn{6}{l}{}                          &       & \multicolumn{6}{l}{}                          &       & \multicolumn{6}{l}{}                          &       & \multicolumn{6}{l}{}                          &       & \multicolumn{6}{l}{}                          &       & \multicolumn{6}{l}{}                          &  \\
          &       & Red.  &       & \multicolumn{6}{l}{}                          &       & \multicolumn{6}{l}{}                          &       & \multicolumn{6}{l}{}                          &       & \multicolumn{6}{l}{}                          &       & \multicolumn{6}{l}{}                          &       & \multicolumn{6}{l}{}                          &       & \multicolumn{6}{l}{}                          &  \\
          &       & \textcolor[rgb]{ 1,  0,  0}{avg} &       & \multicolumn{6}{l}{}                          &       & \multicolumn{6}{l}{}                          &       & \multicolumn{6}{l}{}                          &       & \multicolumn{6}{l}{}                          &       & \multicolumn{6}{l}{}                          &       & \multicolumn{6}{l}{}                          &       & \multicolumn{6}{l}{}                          &  \\
    \midrule
    \multicolumn{3}{c}{analysis aspect} &       & \multicolumn{27}{c}{Architecture}                                                                                                                                                                                     &       & \multicolumn{21}{c}{Generation way} \\
\cmidrule{1-3}\cmidrule{5-31}\cmidrule{33-53}    \multicolumn{3}{c}{model type} &       & \multicolumn{27}{c}{EXT}                                                                                                                                                                                              &       & \multicolumn{6}{c}{LSTM}                      &       & \multicolumn{6}{c}{BERTSUM}                   &       & \multicolumn{6}{c}{Transformer}               &  \\
\cmidrule{1-3}\cmidrule{5-31}\cmidrule{33-38}\cmidrule{40-53}    \multicolumn{3}{c}{compare models} &       & \multicolumn{6}{c}{Trans$_{non}$ vs. LSTM$_{non}$}    &       & \multicolumn{6}{c}{Trans$_{auto}$ vs. Trans$_{non}$}   &       & \multicolumn{6}{c}{BERT$_{match}$ vs. BERT$_{non}$}    &       & \multicolumn{6}{c}{BERT$_{non}$ vs. Trans$_{non}$}      &       & \multicolumn{6}{c}{LSTM$_{non}$  vs.  L2L}         &       & \multicolumn{6}{c}{Trans$_{non}$ vs. T2T}          &       & \multicolumn{6}{c}{BERT$_{non}$ vs. BE2T}          &  \\
\cmidrule{1-10}\cmidrule{12-17}\cmidrule{19-24}\cmidrule{26-31}\cmidrule{33-38}\cmidrule{40-45}\cmidrule{47-52}    \multicolumn{3}{c}{\multirow{2}[2]{*}{holistic analysis}} &       & \multicolumn{6}{c}{stiff. : 28.02 vs. 28.51}  &       & \multicolumn{6}{c}{stiff. : 28.51 vs. 28.02}  &       & \multicolumn{6}{c}{stiff. : 32.27 vs. 28.98}  &       & \multicolumn{6}{c}{stiff. : 28.98 vs. 28.02}  &       & \multicolumn{6}{c}{stiff. : 28.51 vs. 18.03}  &       & \multicolumn{6}{c}{stiff. : 28.02 vs. 19.79}  &       & \multicolumn{6}{c}{stiff. : 28.98 vs. 23.49}  &  \\
    \multicolumn{3}{c}{}  &       & \multicolumn{6}{c}{stable. : 99.05 vs. 87.00} &       & \multicolumn{6}{c}{stable. : 88.71 vs. 99.05} &       & \multicolumn{6}{c}{stable. : 91.98 vs. 88.93} &       & \multicolumn{6}{c}{stable. : 88.93 vs. 99.05} &       & \multicolumn{6}{c}{stable. : 87.00 vs. 66.93} &       & \multicolumn{6}{c}{stable. : 99.05 vs. 62.12} &       & \multicolumn{6}{c}{stable. : 88.93 vs. 62.93} &  \\
\cmidrule{1-3}\cmidrule{5-10}\cmidrule{12-17}\cmidrule{19-24}\cmidrule{26-31}\cmidrule{33-38}\cmidrule{40-45}\cmidrule{47-52}    \multicolumn{3}{c}{fine-grain analysis} &       & \multicolumn{1}{c}{\rotatebox{90}{CNN.}} & \multicolumn{1}{c}{\rotatebox{90}{Xsum}} & \multicolumn{1}{c}{\rotatebox{90}{Pubm.}} & \multicolumn{1}{c}{\rotatebox{90}{Patent b}} & \multicolumn{1}{c}{\rotatebox{90}{Red.}} & \multicolumn{1}{c}{\textcolor[rgb]{ 1,  0,  0}{\rotatebox{90}{avg}}} &       & \multicolumn{1}{c}{\rotatebox{90}{CNN.}} & \multicolumn{1}{c}{\rotatebox{90}{Xsum}} & \multicolumn{1}{c}{\rotatebox{90}{Pubm.}} & \multicolumn{1}{c}{\rotatebox{90}{Patent b}} & \multicolumn{1}{c}{\rotatebox{90}{Red.}} & \multicolumn{1}{c}{\textcolor[rgb]{ 1,  0,  0}{\rotatebox{90}{avg}}} &       & \multicolumn{1}{c}{\rotatebox{90}{CNN.}} & \multicolumn{1}{c}{\rotatebox{90}{Xsum}} & \multicolumn{1}{c}{\rotatebox{90}{Pubm.}} & \multicolumn{1}{c}{\rotatebox{90}{Patent b}} & \multicolumn{1}{c}{\rotatebox{90}{Red.}} & \multicolumn{1}{c}{\textcolor[rgb]{ 1,  0,  0}{\rotatebox{90}{avg}}} &       & \multicolumn{1}{c}{\rotatebox{90}{CNN.}} & \multicolumn{1}{c}{\rotatebox{90}{Xsum}} & \multicolumn{1}{c}{\rotatebox{90}{Pubm.}} & \multicolumn{1}{c}{\rotatebox{90}{Patent b}} & \multicolumn{1}{c}{\rotatebox{90}{Red.}} & \multicolumn{1}{c}{\textcolor[rgb]{ 1,  0,  0}{\rotatebox{90}{avg}}} &       & \multicolumn{1}{c}{\rotatebox{90}{CNN.}} & \multicolumn{1}{c}{\rotatebox{90}{Xsum}} & \multicolumn{1}{c}{\rotatebox{90}{Pubm.}} & \multicolumn{1}{c}{\rotatebox{90}{Patent b}} & \multicolumn{1}{c}{\rotatebox{90}{Red.}} & \multicolumn{1}{c}{\textcolor[rgb]{ 1,  0,  0}{\rotatebox{90}{avg}}} &       & \multicolumn{1}{c}{\rotatebox{90}{CNN.}} & \multicolumn{1}{c}{\rotatebox{90}{Xsum}} & \multicolumn{1}{c}{\rotatebox{90}{Pubm.}} & \multicolumn{1}{c}{\rotatebox{90}{Patent b}} & \multicolumn{1}{c}{\rotatebox{90}{Red.}} & \multicolumn{1}{c}{\textcolor[rgb]{ 1,  0,  0}{\rotatebox{90}{avg}}} &       & \multicolumn{1}{c}{\rotatebox{90}{CNN.}} & \multicolumn{1}{c}{\rotatebox{90}{Xsum}} & \multicolumn{1}{c}{\rotatebox{90}{Pubm.}} & \multicolumn{1}{c}{\rotatebox{90}{Patent b}} & \multicolumn{1}{c}{\rotatebox{90}{Red.}} & \multicolumn{1}{c}{\textcolor[rgb]{ 1,  0,  0}{\rotatebox{90}{avg}}} &  \\
\cmidrule{1-10}\cmidrule{12-17}\cmidrule{19-24}\cmidrule{26-38}\cmidrule{40-45}\cmidrule{47-52}    \multirow{12}[4]{*}{\rotatebox{90}{ROUGE}} & \multirow{6}[2]{*}{\rotatebox{90}{origin}} & CNN.  &       & \multicolumn{6}{l}{\multirow{6}[2]{*}{\includegraphics[height=200pt,width=280pt]{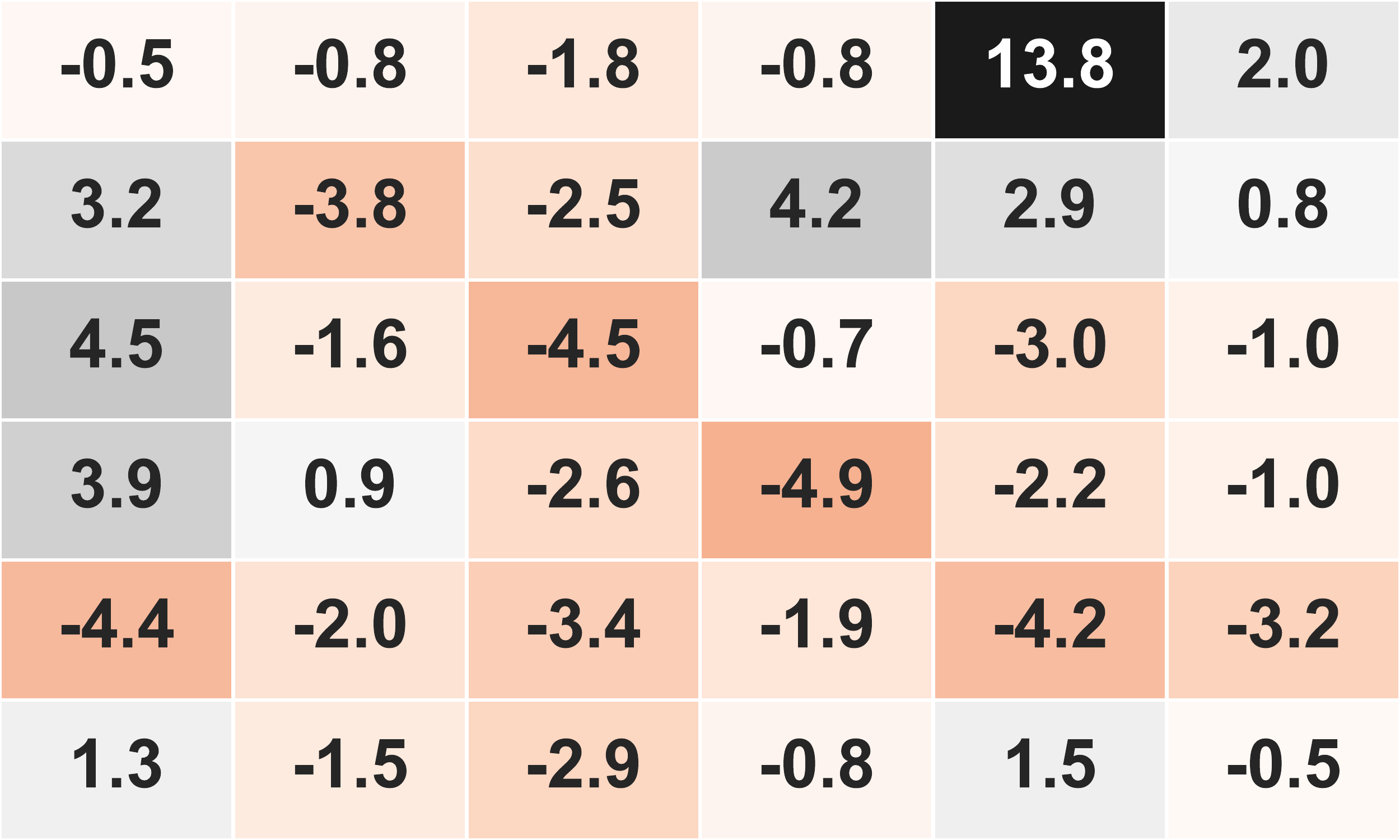} }}  &       & \multicolumn{6}{l}{\multirow{6}[2]{*}{\includegraphics[height=200pt,width=280pt]{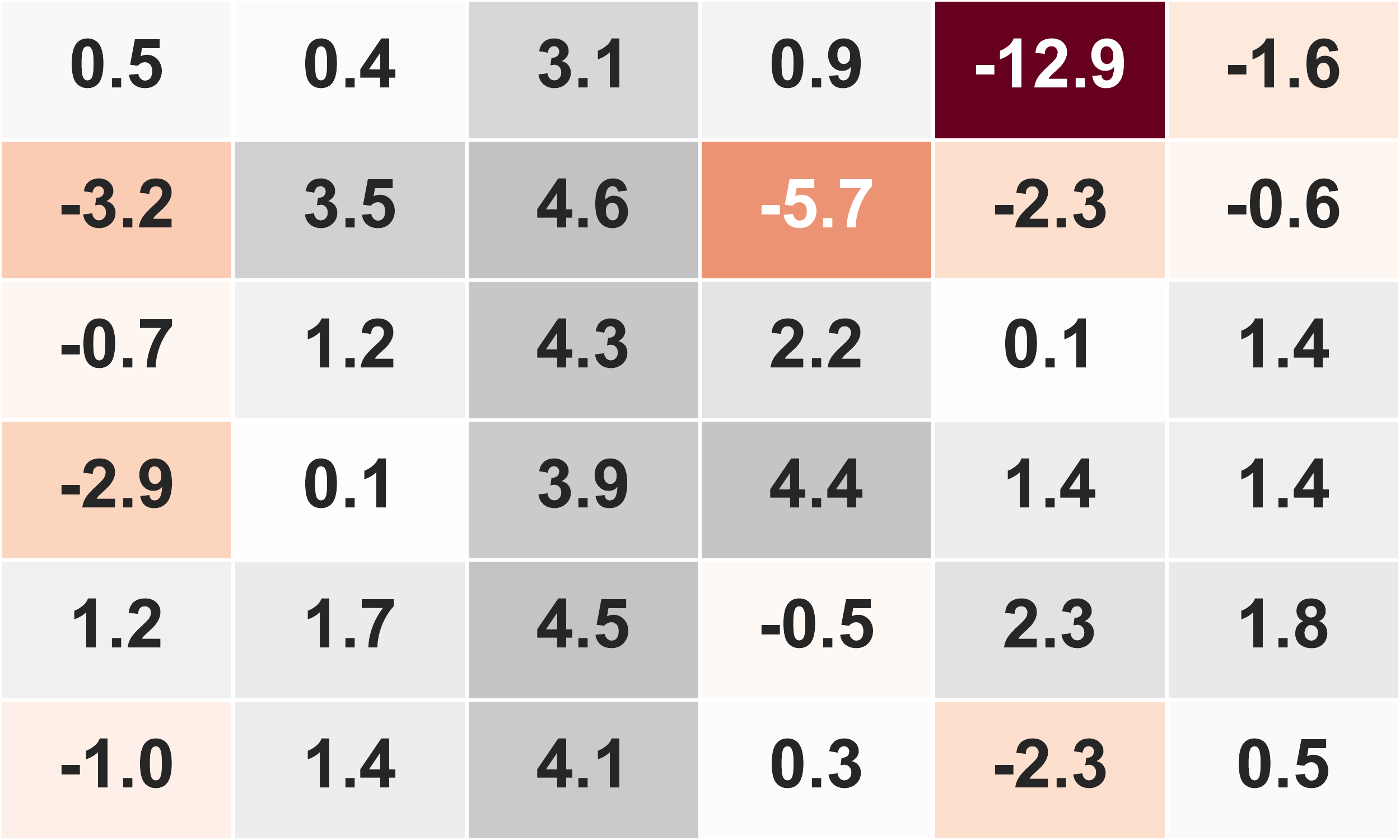} }}  &       & \multicolumn{6}{l}{\multirow{6}[2]{*}{\includegraphics[height=200pt,width=280pt]{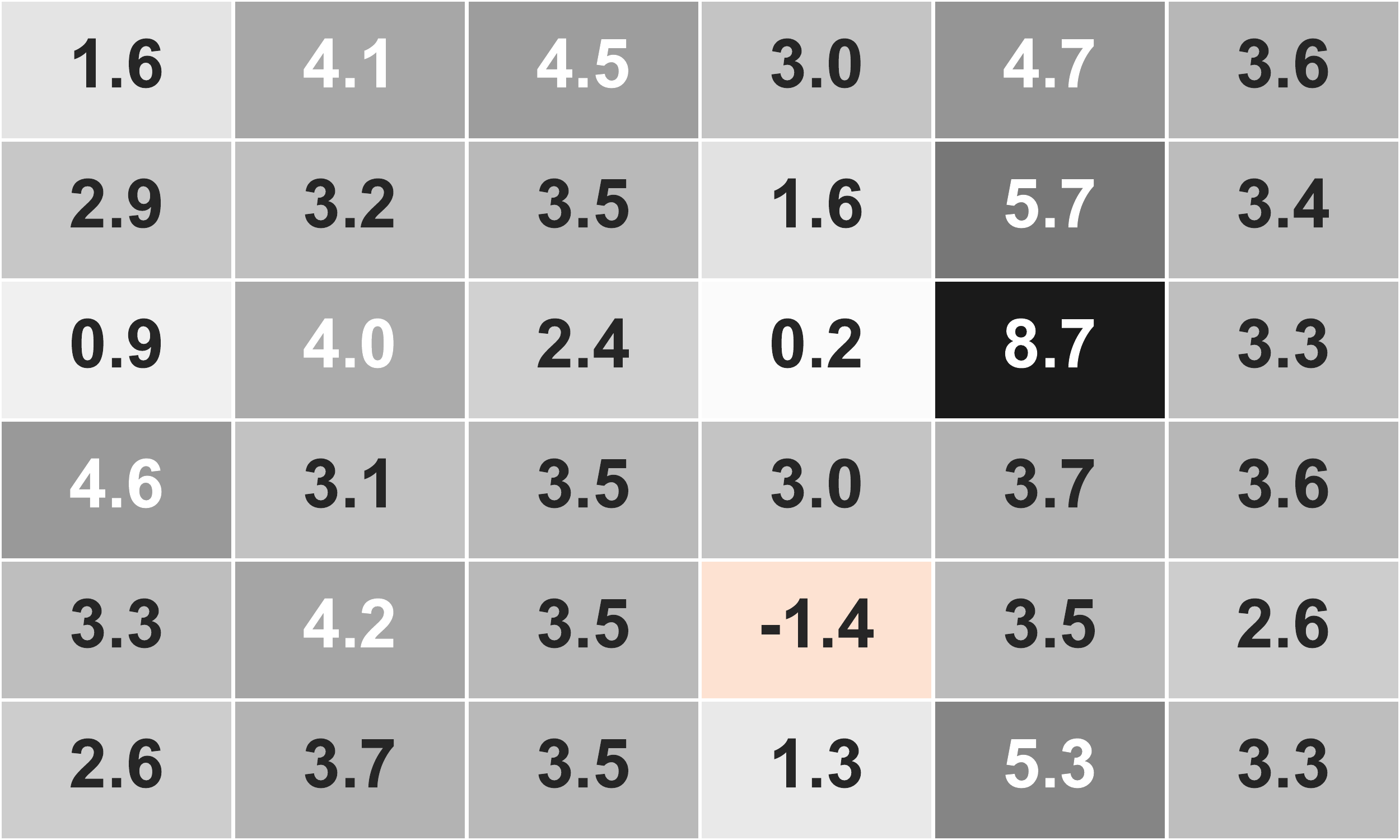} }}  &       & \multicolumn{6}{l}{\multirow{6}[2]{*}{\includegraphics[height=200pt,width=280pt]{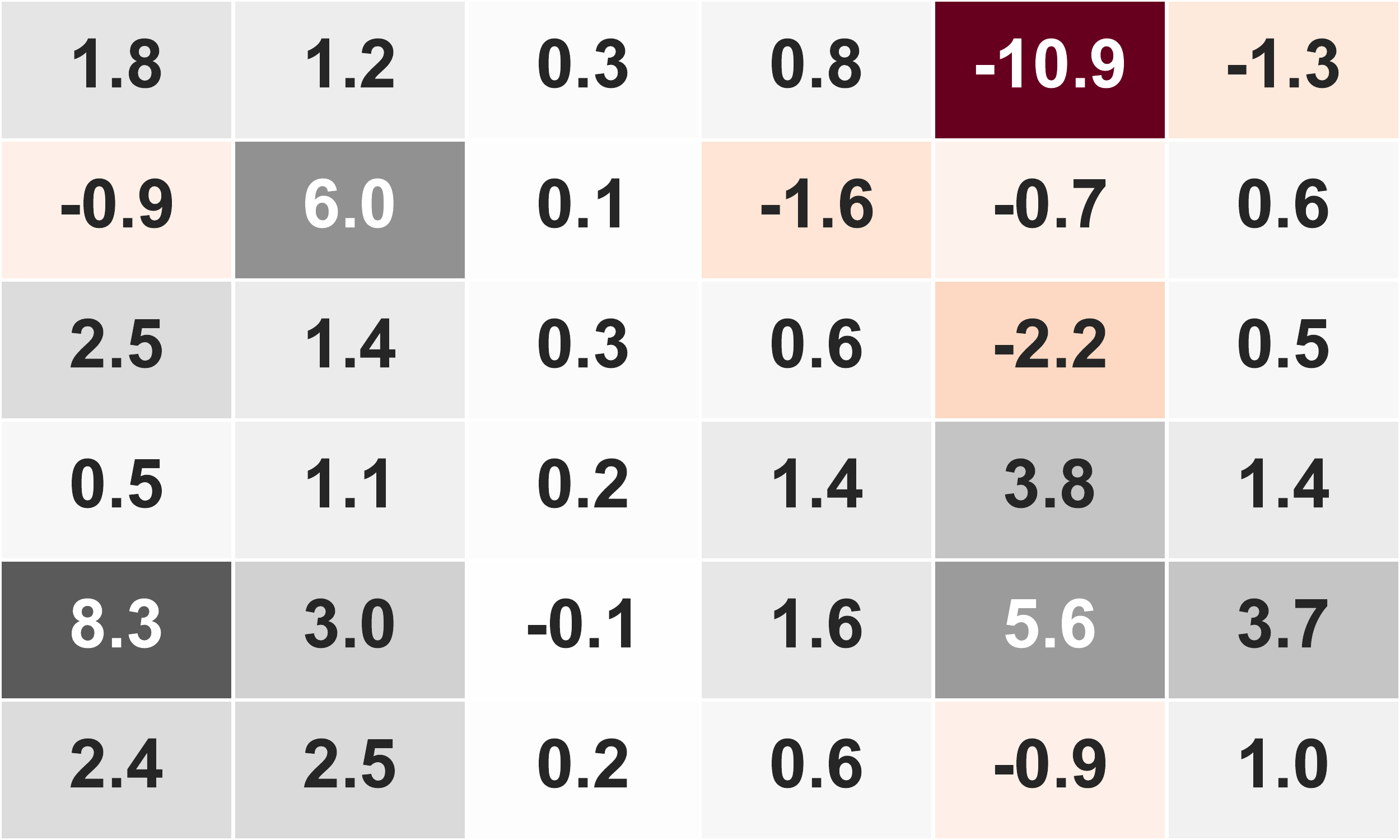} }}  &       & \multicolumn{6}{l}{\multirow{6}[2]{*}{\includegraphics[height=200pt,width=280pt]{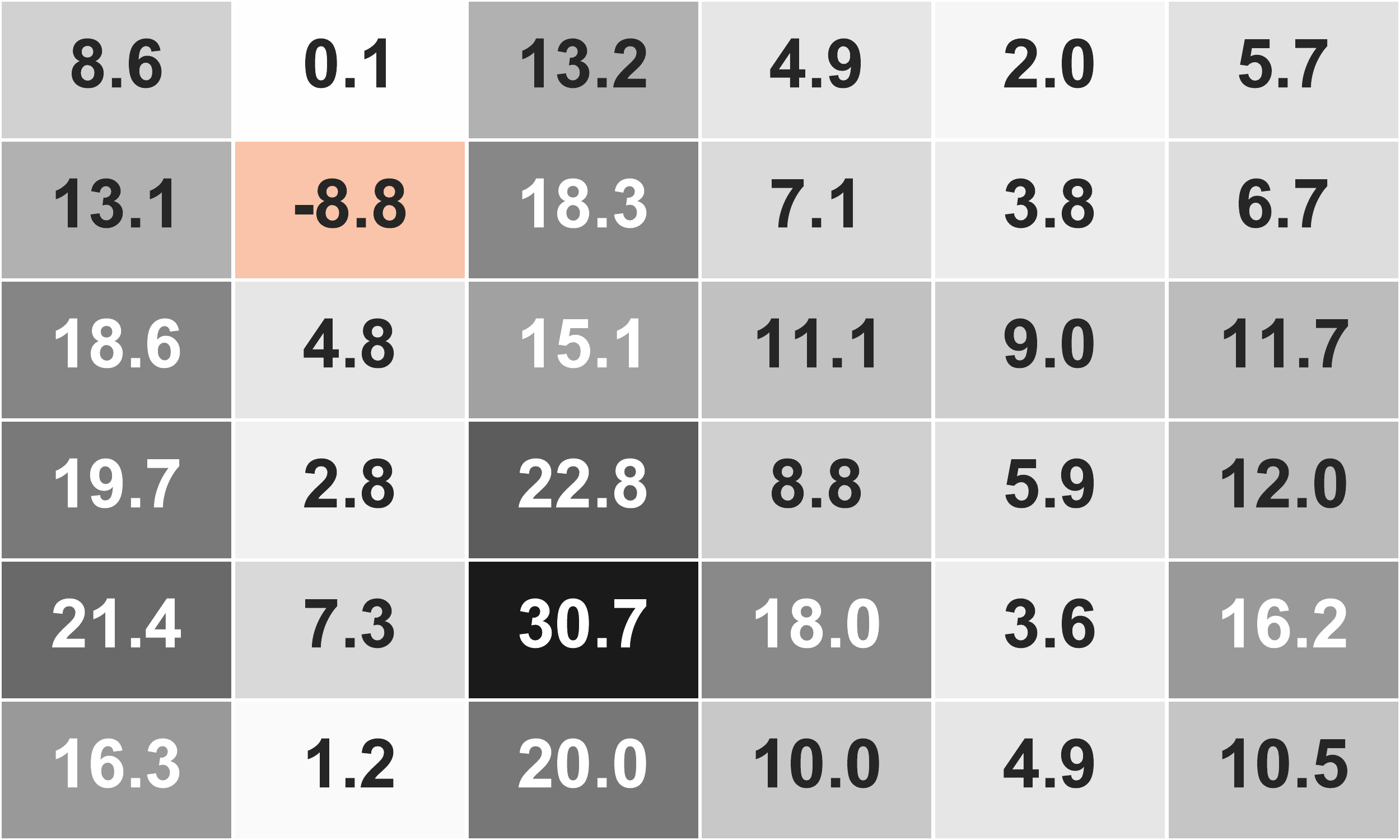} }}  &       & \multicolumn{6}{l}{\multirow{6}[2]{*}{\includegraphics[height=200pt,width=280pt]{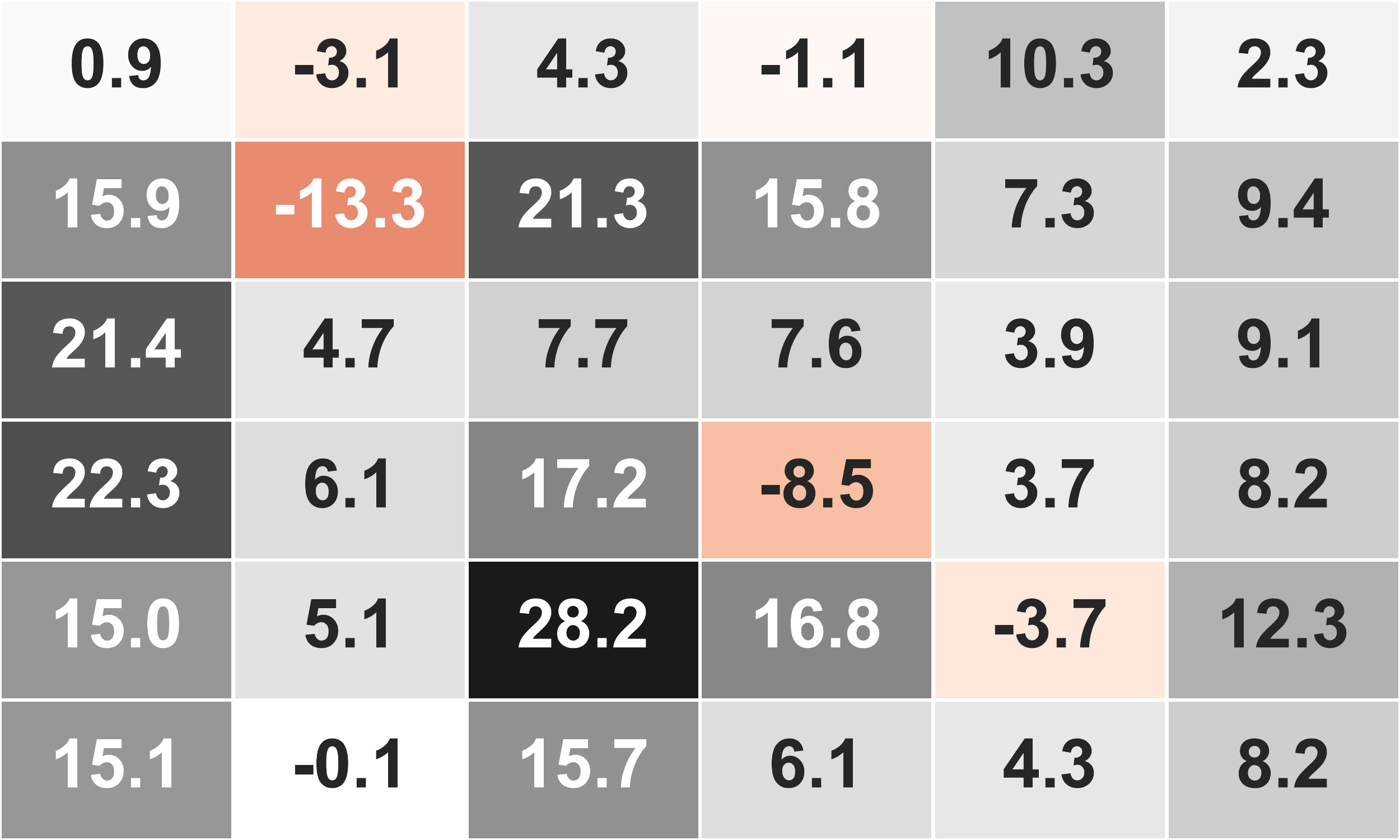} }}  &       & \multicolumn{6}{l}{\multirow{6}[2]{*}{\includegraphics[height=200pt,width=280pt]{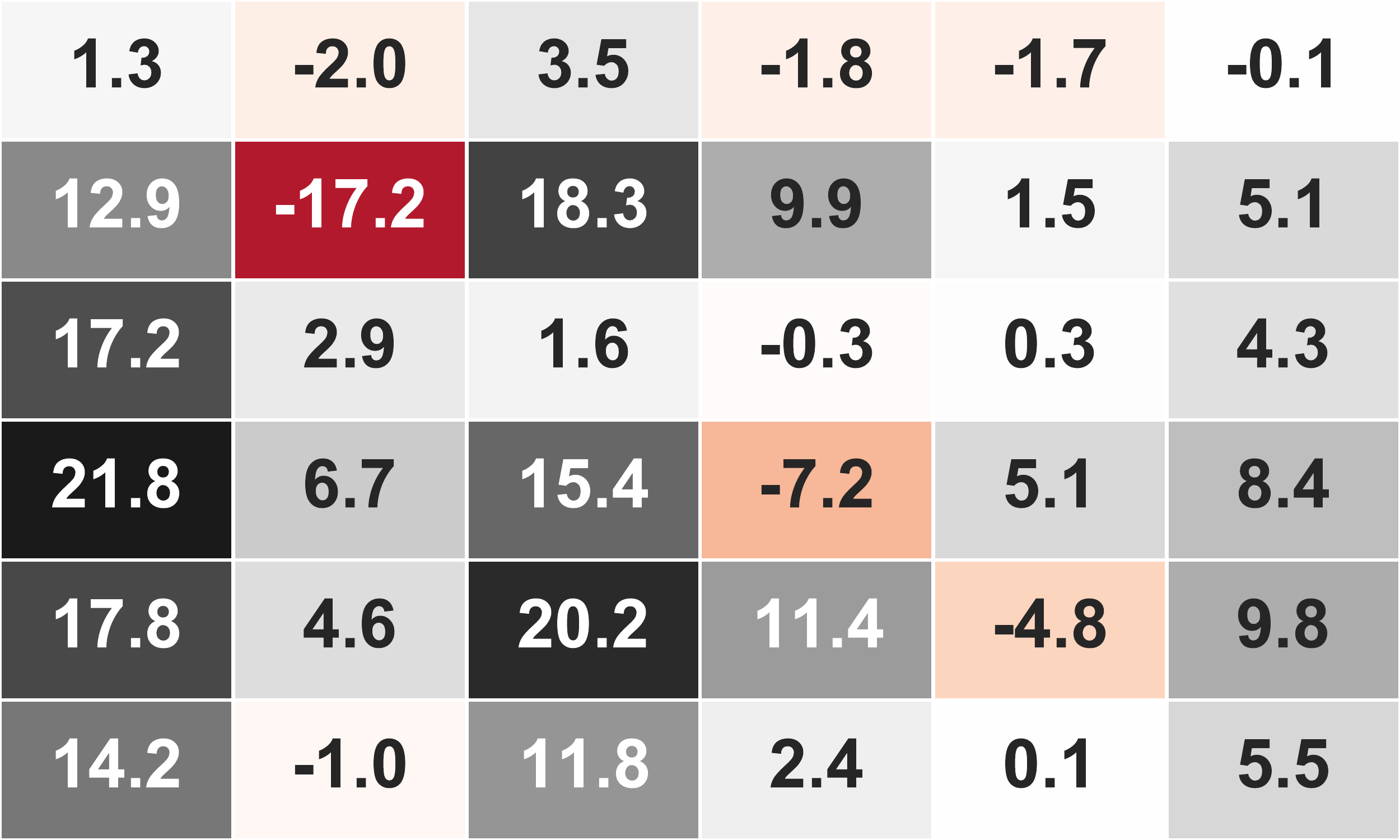} }}  &  \\
          &       & Xsum  &       & \multicolumn{6}{l}{}                          &       & \multicolumn{6}{l}{}                          &       & \multicolumn{6}{l}{}                          &       & \multicolumn{6}{l}{}                          &       & \multicolumn{6}{l}{}                          &       & \multicolumn{6}{l}{}                          &       & \multicolumn{6}{l}{}                          &  \\
          &       & Pubm. &       & \multicolumn{6}{l}{}                          &       & \multicolumn{6}{l}{}                          &       & \multicolumn{6}{l}{}                          &       & \multicolumn{6}{l}{}                          &       & \multicolumn{6}{l}{}                          &       & \multicolumn{6}{l}{}                          &       & \multicolumn{6}{l}{}                          &  \\
          &       & Patent b &       & \multicolumn{6}{l}{}                          &       & \multicolumn{6}{l}{}                          &       & \multicolumn{6}{l}{}                          &       & \multicolumn{6}{l}{}                          &       & \multicolumn{6}{l}{}                          &       & \multicolumn{6}{l}{}                          &       & \multicolumn{6}{l}{}                          &  \\
          &       & Red.  &       & \multicolumn{6}{l}{}                          &       & \multicolumn{6}{l}{}                          &       & \multicolumn{6}{l}{}                          &       & \multicolumn{6}{l}{}                          &       & \multicolumn{6}{l}{}                          &       & \multicolumn{6}{l}{}                          &       & \multicolumn{6}{l}{}                          &  \\
          &       & \textcolor[rgb]{ 1,  0,  0}{avg} &       & \multicolumn{6}{l}{}                          &       & \multicolumn{6}{l}{}                          &       & \multicolumn{6}{l}{}                          &       & \multicolumn{6}{l}{}                          &       & \multicolumn{6}{l}{}                          &       & \multicolumn{6}{l}{}                          &       & \multicolumn{6}{l}{}                          &  \\
\cmidrule{5-10}\cmidrule{12-17}\cmidrule{19-24}\cmidrule{26-38}\cmidrule{40-45}\cmidrule{47-52}          & \multirow{6}[2]{*}{\rotatebox{90}{normali.}} & CNN.  &       & \multicolumn{6}{l}{\multirow{6}[2]{*}{\includegraphics[height=200pt,width=280pt]{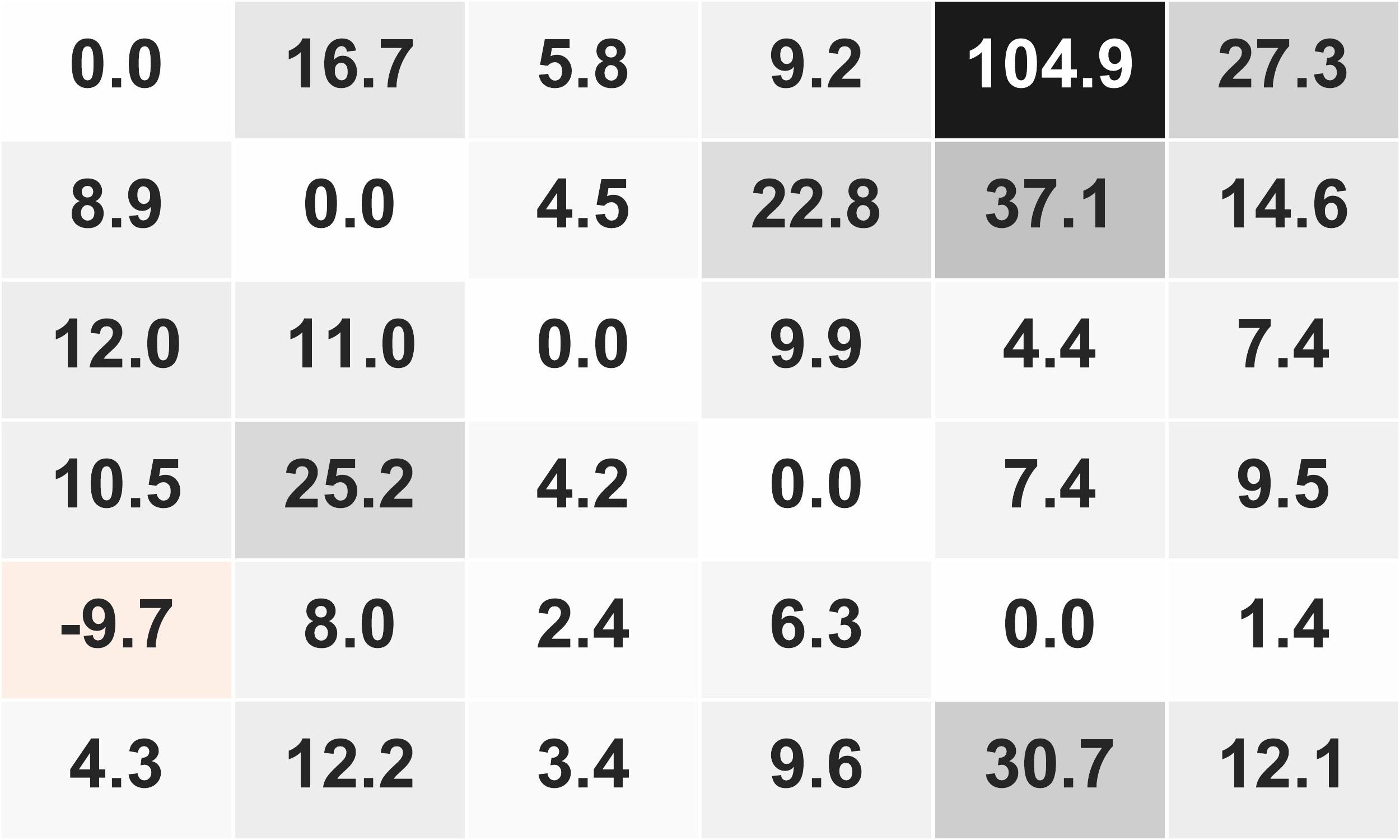} }}  &       & \multicolumn{6}{l}{\multirow{6}[2]{*}{\includegraphics[height=200pt,width=280pt]{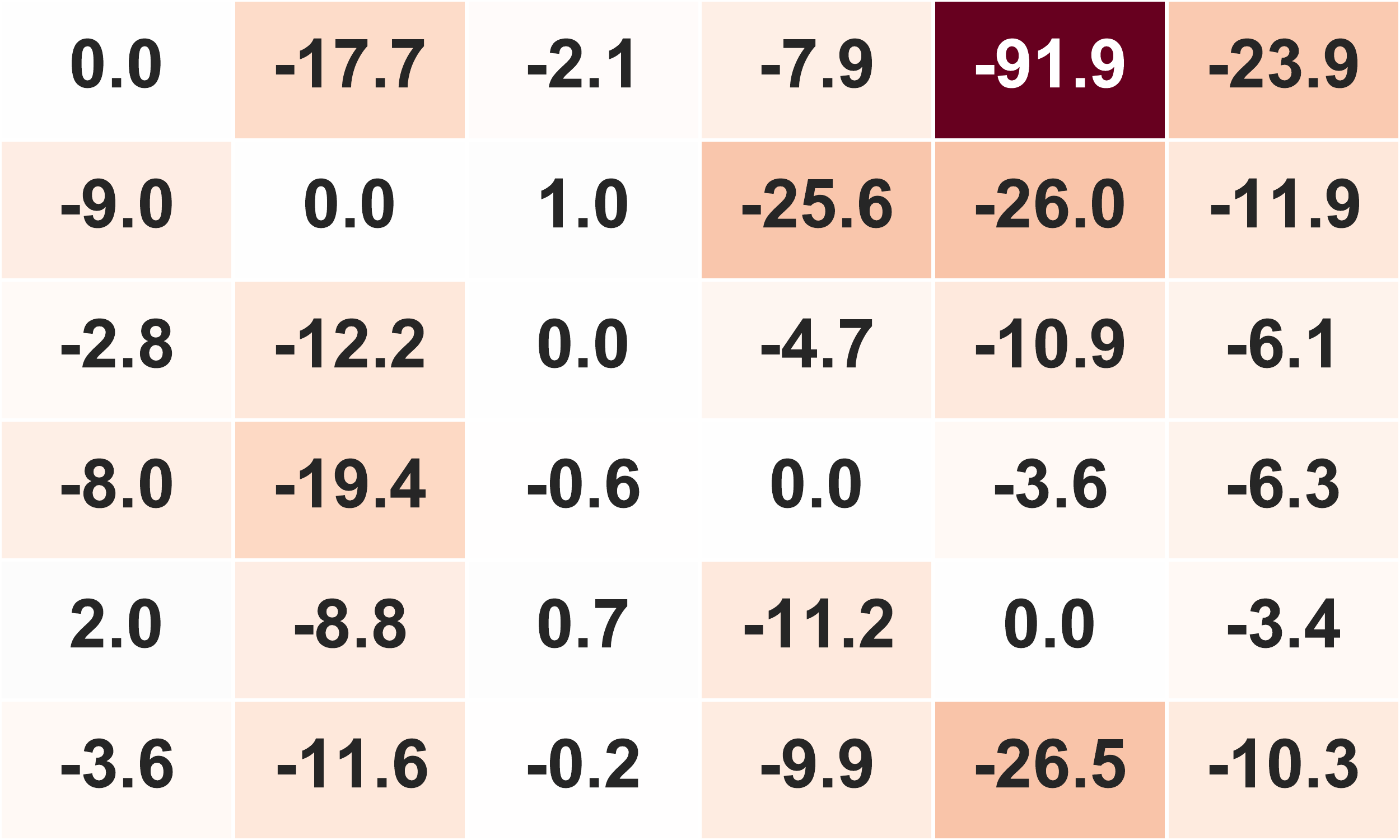} }}  &       & \multicolumn{6}{l}{\multirow{6}[2]{*}{\includegraphics[height=200pt,width=280pt]{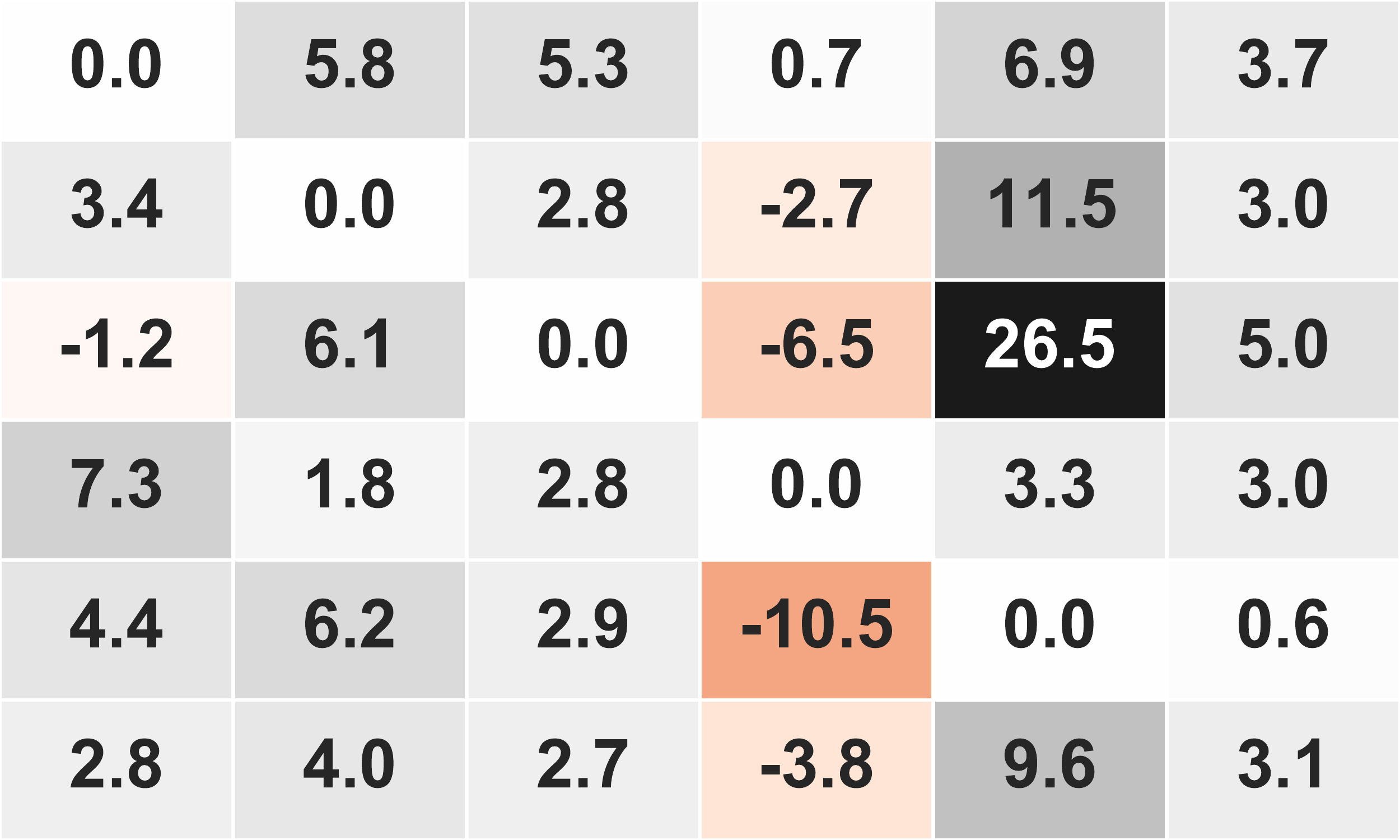} }}  &       & \multicolumn{6}{l}{\multirow{6}[2]{*}{\includegraphics[height=200pt,width=280pt]{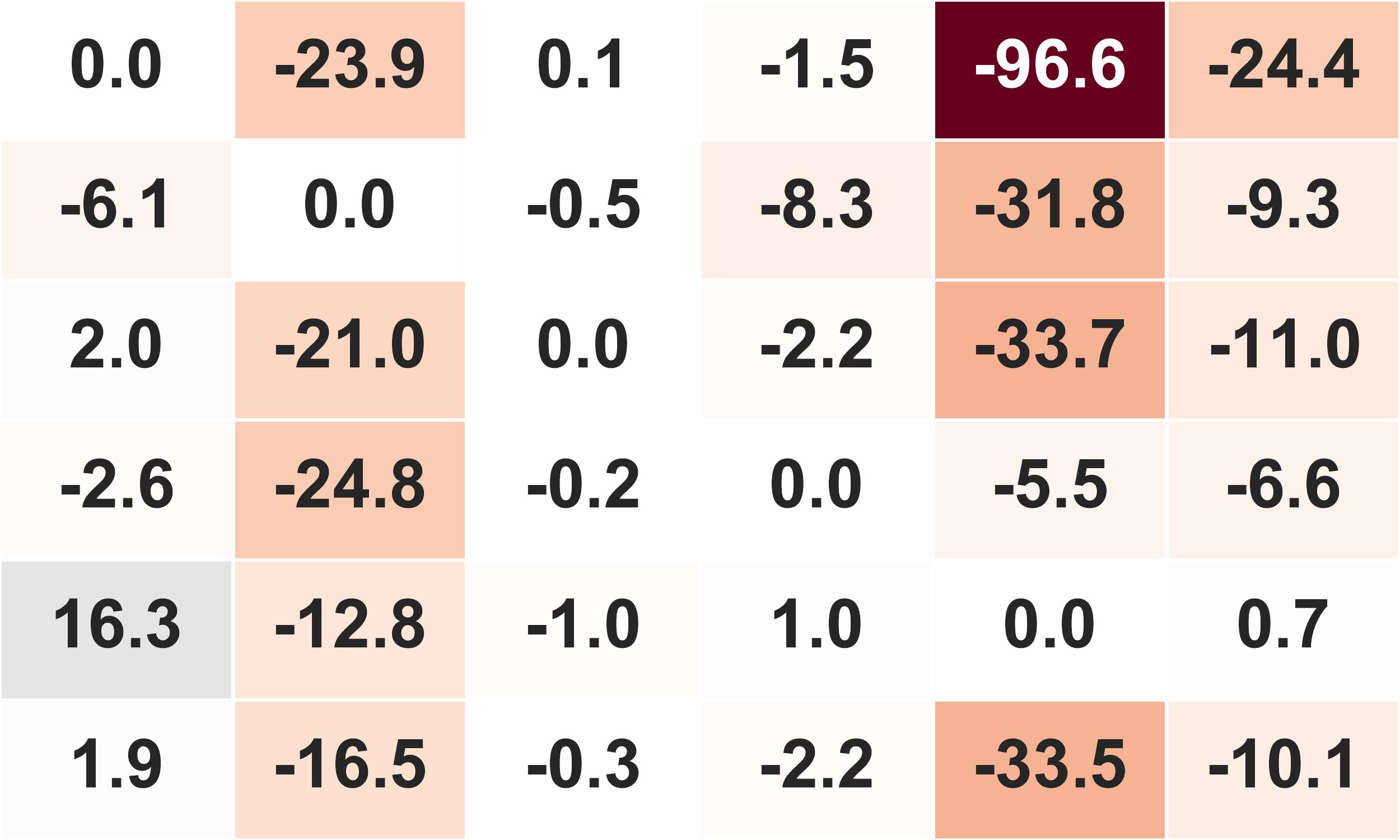} }}  &       & \multicolumn{6}{l}{\multirow{6}[2]{*}{\includegraphics[height=200pt,width=280pt]{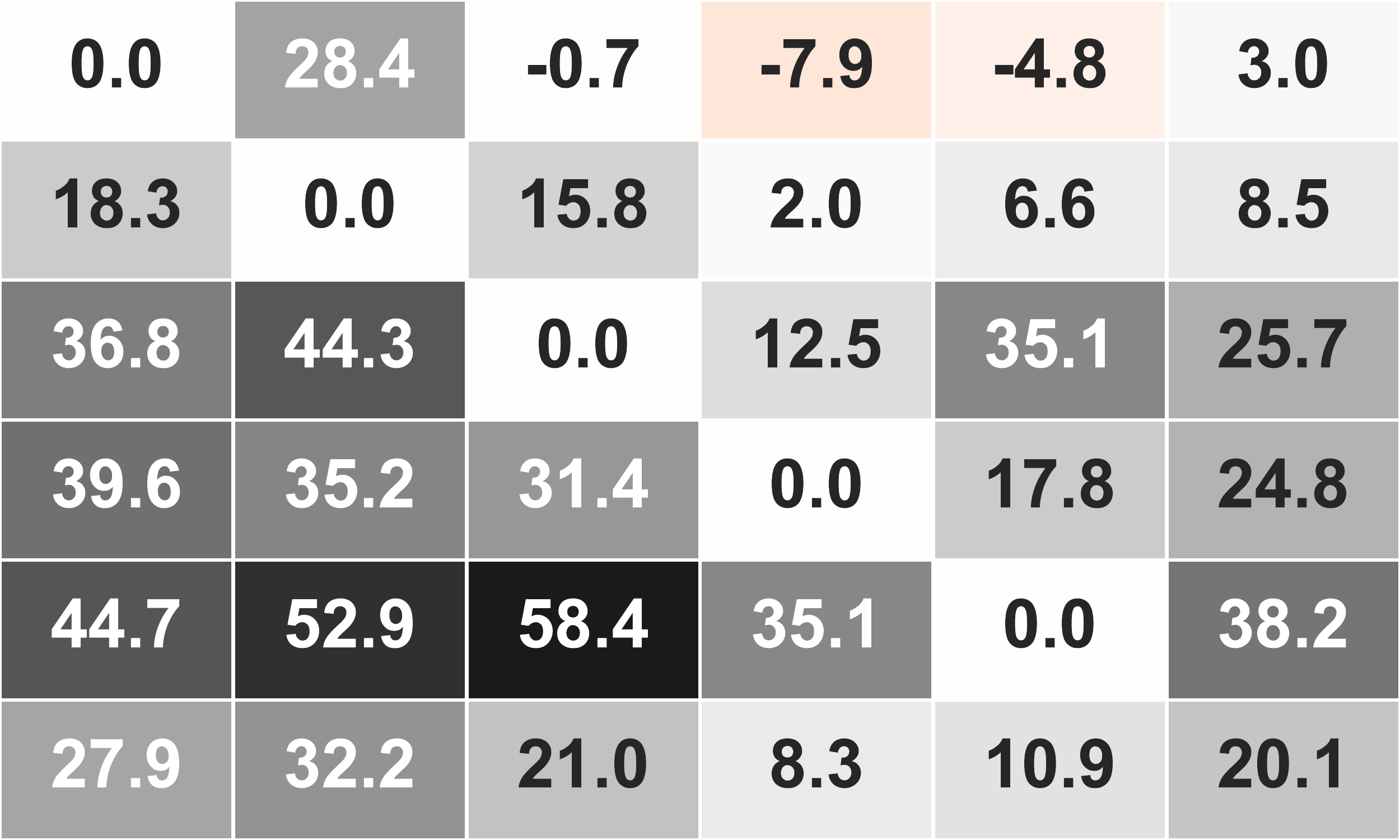} }}  &       & \multicolumn{6}{l}{\multirow{6}[2]{*}{\includegraphics[height=200pt,width=280pt]{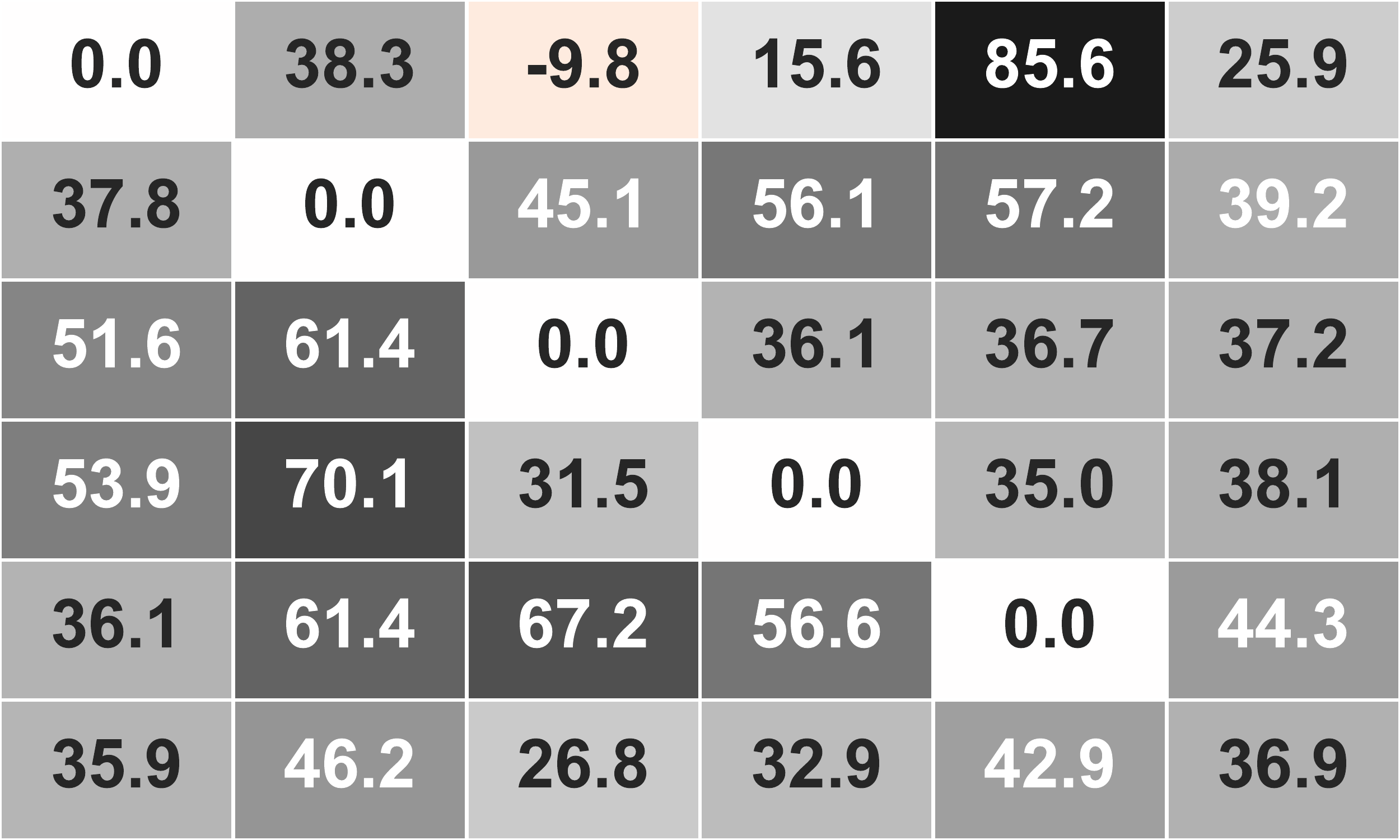} }}  &       & \multicolumn{6}{l}{\multirow{6}[2]{*}{\includegraphics[height=200pt,width=280pt]{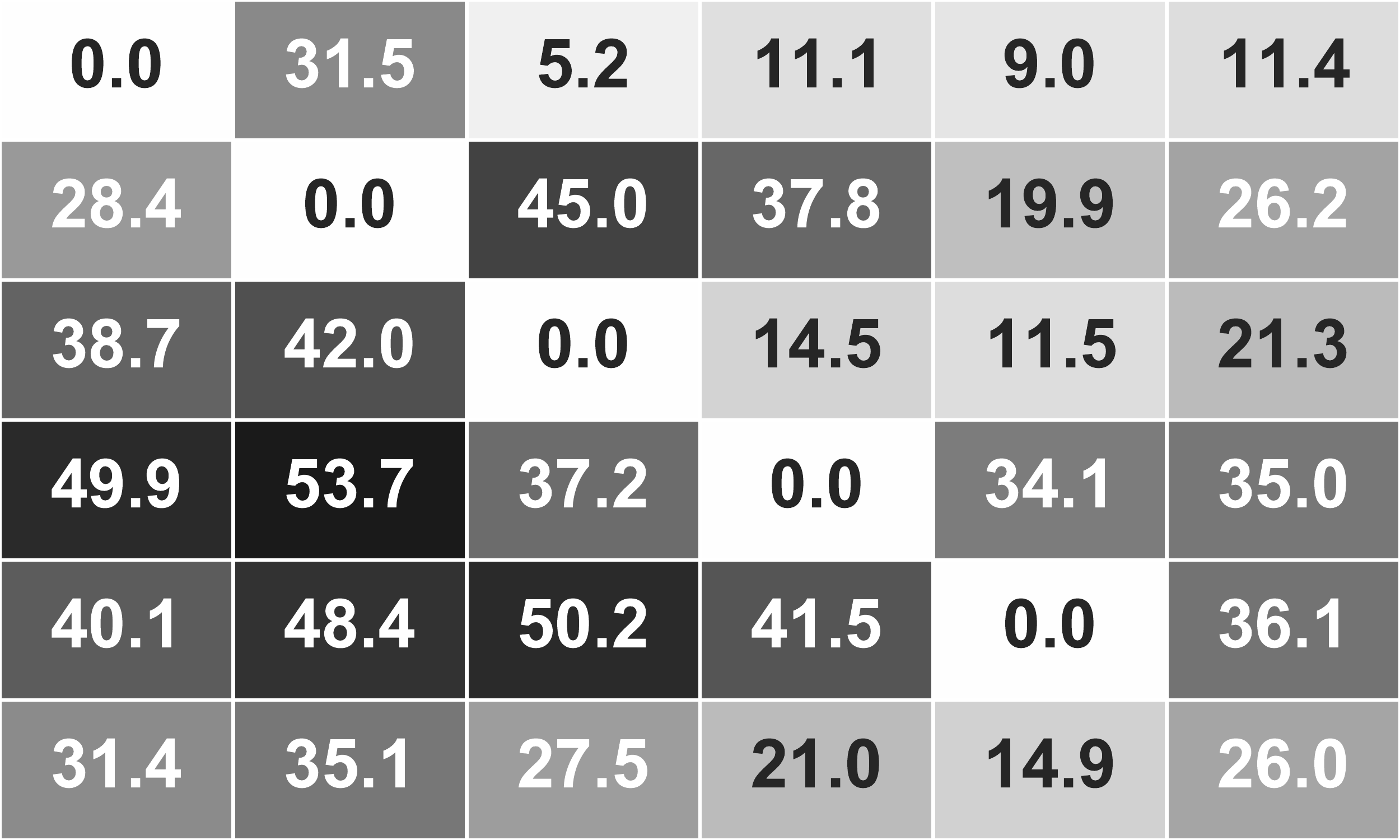}}}  &  \\
          &       & Xsum  &       & \multicolumn{6}{l}{}                          &       & \multicolumn{6}{l}{}                          &       & \multicolumn{6}{l}{}                          &       & \multicolumn{6}{l}{}                          &       & \multicolumn{6}{l}{}                          &       & \multicolumn{6}{l}{}                          &       & \multicolumn{6}{l}{}                          &  \\
          &       & Pubm. &       & \multicolumn{6}{l}{}                          &       & \multicolumn{6}{l}{}                          &       & \multicolumn{6}{l}{}                          &       & \multicolumn{6}{l}{}                          &       & \multicolumn{6}{l}{}                          &       & \multicolumn{6}{l}{}                          &       & \multicolumn{6}{l}{}                          &  \\
          &       & Patent b &       & \multicolumn{6}{l}{}                          &       & \multicolumn{6}{l}{}                          &       & \multicolumn{6}{l}{}                          &       & \multicolumn{6}{l}{}                          &       & \multicolumn{6}{l}{}                          &       & \multicolumn{6}{l}{}                          &       & \multicolumn{6}{l}{}                          &  \\
          &       & Red.  &       & \multicolumn{6}{l}{}                          &       & \multicolumn{6}{l}{}                          &       & \multicolumn{6}{l}{}                          &       & \multicolumn{6}{l}{}                          &       & \multicolumn{6}{l}{}                          &       & \multicolumn{6}{l}{}                          &       & \multicolumn{6}{l}{}                          &  \\
          &       & \textcolor[rgb]{ 1,  0,  0}{avg} &       & \multicolumn{6}{l}{}                          &       & \multicolumn{6}{l}{}                          &       & \multicolumn{6}{l}{}                          &       & \multicolumn{6}{l}{}                          &       & \multicolumn{6}{l}{}                          &       & \multicolumn{6}{l}{}                          &       & \multicolumn{6}{l}{}                          &  \\
    \bottomrule
    \end{tabular}}%
    \caption{The difference of ROUGE-1 F1 scores between different models pairs. Every column of the table represents the compared result of one pair of models. The line of holistic analysis displays the overall stiffness and stableness of compared models.
    The rest of the table is the fine-grained results, the first and third lines of which are the origin compared result ($\mathbf{U_A} - \mathbf{U_B}$ for models pairs $A$ and $B$) and the second and fourth lines are the normalized compared result ($\mathbf{\hat{U}_A} - \mathbf{\hat{U}_B}$ for models pairs $A$ and $B$). For all heatmap, `grey' represents positive, `red' represents negative and `white' represents approximately zero.}
  \label{tab:all fine grain R1}%
\end{table*}%

\end{document}